\documentclass{article}
\usepackage{footnote}
\makesavenoteenv{figure}
\usepackage{arxiv}
\usepackage[font=small]{caption}
\usepackage[utf8]{inputenc}
\usepackage[T1]{fontenc}
\usepackage{CJKutf8}
\usepackage{microtype}
\usepackage{amsmath}
\usepackage{amssymb,amsthm}
\usepackage{makecell}

\usepackage[dvipsnames]{xcolor}
\definecolor{kimiblue}{rgb}{0.09,0.5,0.99}
\usepackage[breaklinks,colorlinks,allcolors=kimiblue]{hyperref}
\usepackage{url}            
\usepackage{booktabs}       
\usepackage{amsfonts}       
\usepackage[normalem]{ulem}
\usepackage{xcolor}         
\usepackage{nicefrac}       
\usepackage{textgreek}
\usepackage{graphicx}
\usepackage{standalone} 
\usepackage{anyfontsize} 
\usepackage{tikz}
\usetikzlibrary{arrows.meta, positioning, calc, shapes.geometric, shapes.misc, fit, backgrounds}
\usepackage{pgfplots}
\usepackage{pgfplotstable}
\usepackage{listings}
\usepackage[utf8]{inputenc}
\usepackage{doi}

\usepackage[ruled,linesnumbered]{algorithm2e}
\IncMargin{1.5em}   

\usepackage{multirow}
\usepackage{multicol}
\usepackage{xcolor}
\usepackage{subcaption}
\usepackage[normalem]{ulem}
\usepackage{amsmath,amsfonts,bm}

\usepackage{booktabs}
\usepackage{tabularx}

\usepackage{enumitem}
\usepackage{subcaption}
\usepackage{adjustbox}
\usepackage{array}
\usepackage{makecell}
\usepackage{wrapfig}
\usepackage{ulem}
\usepackage[
    backend=biber,
    citestyle=numeric,
    bibstyle=numeric,
    mincitenames=1,
    maxcitenames=1,
    maxbibnames=3,
]{biblatex}
\newcommand{\citep}[1]{\parencite{#1}}
\newcommand{\kimilogo}{\raisebox{-0.15\height}{\includegraphics[width=0.032\textwidth]{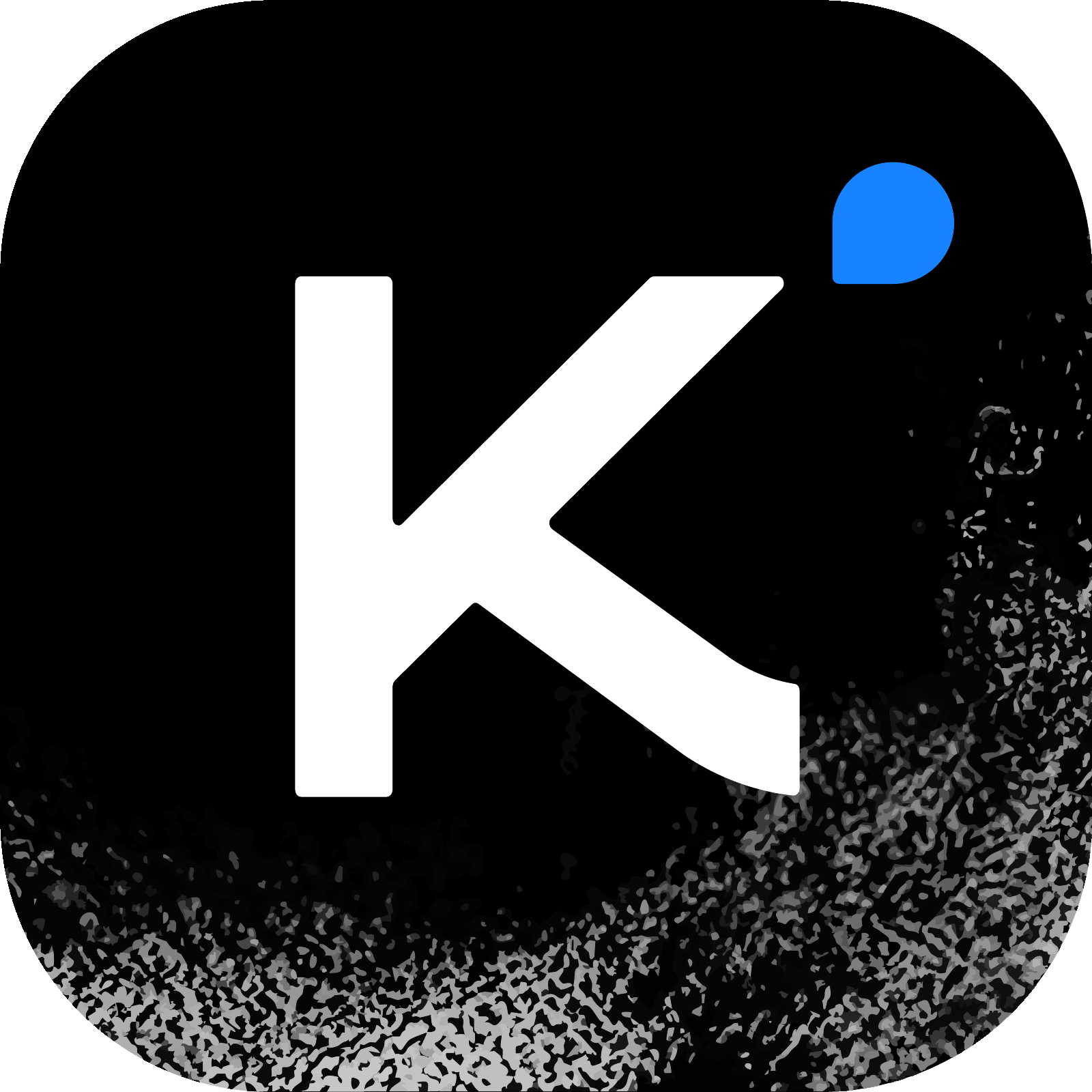}}}
\newcommand{\kimishort}{\raisebox{-0.12\height}{\includegraphics[width=0.02\textwidth]{figures/logo.pdf}}}
\setlist[itemize]{leftmargin=\dimexpr 12pt}
\setlist[enumerate]{leftmargin=\dimexpr 12pt}

\title{
\kimilogo \ Kimi K3: Open Frontier Intelligence}

\author{Kimi Team}
\date{}

\renewcommand{\headeright}{Technical Report}
\renewcommand{\undertitle}{Technical Report of Kimi K3}
\renewcommand{\shorttitle}{\kimishort\ Kimi K3: Open Frontier Intelligence}

\definecolor{darkblue}{rgb}{0.0, 0.0, 0.5}
\definecolor{darkgreen}{rgb}{0.0, 0.5, 0.0}
\definecolor{darkred}{rgb}{0.5, 0.0, 0.0}
\definecolor{darkpurple}{rgb}{0.5, 0.0, 0.5}
\definecolor{brickred}{HTML}{b92622}

\newcommand{\kimi}[1]{Kimi K#1}

\begin{document}
\maketitle

\vspace{-17pt}
\begin{abstract}
We introduce \kimi{3}, a 2.8T parameter Mixture-of-Experts model with 104 billion activated parameters, native vision capabilities, and a 1-million-token context window. 
\kimi{3} is built on Kimi Delta Attention~\citep{team2025kimi} and Attention Residuals~\citep{kimiteam2026attnres}, which improve information flow across sequence length and model depth. 
Together with Stable LatentMoE, which effectively activates 16 of 896 routed experts per token, and refined training and data recipes, these advances yield an approximately $2.5\times$ improvement in overall scaling efficiency over \kimi{2}~\citep{kimiteam2025kimik2openagentic}.
Post-training highlights reinforcement learning across general, agentic, and coding domains and multiple reasoning-effort levels, enabling compositional generalization and robust long-horizon execution. 
At 2.8T scale, Kimi K3 is supported by infrastructure advances in multiple areas: algorithm-system co-design for KDA, perfectly balanced expert-parallel training with efficient memory management, million-token agentic RL with persistent rollout and sandbox states, and deployment innovations.

Extensive evaluations show that \kimi{3} achieves frontier-level performance across long-horizon coding, agentic, knowledge, reasoning, and vision tasks. 
While its overall performance still trails the most powerful proprietary models, namely Claude Fable 5 and GPT-5.6 Sol, \kimi{3} consistently outperforms other open and proprietary models evaluated in our suite. 
We release the full \kimi{3} model weights to facilitate future research and accelerate the broader deployment and adoption of frontier intelligence.\footnote{\url{https://huggingface.co/moonshotai/Kimi-K3}}
\end{abstract}

\begin{figure}[htb]
    \centering
    \resizebox{0.95\textwidth}{!}{\input{figures/benchmark0727-fixed}}
    \caption{\kimi{3} main results.\protect}
    \label{fig:kimi-k3-results}
\end{figure}

\section{Introduction}
\label{sec:intro}

For much of the development of Large Language Models (LLMs), scaling meant investing more computation before deployment by training larger models on more data~\citep{kaplan2020scaling,hoffmann2022training}.
The rise of reasoning models has established test-time computation as a second axis of scaling: OpenAI's o-series scales reinforcement learning and test-time reasoning~\citep{openai2024o1,openai2025o3}; Anthropic's extended-thinking models allocate adaptive thinking budgets and interleave reasoning with tool use~\citep{anthropic2025extendedthinking,anthropic2025claude4}; DeepSeek-R1~\citep{deepseekai2025deepseekr1} and \kimi{1.5}~\citep{kimik15} show that large-scale reinforcement learning can elicit sophisticated reasoning behaviors from strong pre-trained models; and \kimi{2.5} Agent Swarm~\citep{kimik25} further extends test-time scaling from sequential reasoning to parallel agent coordination.
These advances have made test-time scaling a central focus of frontier research.
However, while the open-source model ecosystem has advanced rapidly on the second axis, it has progressed slowly on the first: many recent models remain within or slightly above the 1T-class parameter regime~\citep{glm5team2026glm5vibecodingagentic,deepseekv4,mimo2026v25pro,thinkingmachines2026inkling}.
As increasingly sophisticated reasoning and agentic reinforcement learning methods are applied to pre-trained foundations of similar scale, open-source progress risks converging while the gap to the strongest proprietary systems widens.
With \kimi{3}, we pursue both scaling axes together to the frontier: scaling the pre-trained foundation to unprecedented 3T-class parameters while scaling reinforcement learning, reasoning effort, and long-horizon interaction at 1M context length.

We introduce \kimi{3}, a native multimodal Mixture-of-Experts model with 2.8 trillion total parameters, 104 billion activated parameters, and a context window of up to one million tokens.
Its architecture scales information flow across sequence length, network depth, and model width.
Kimi Delta Attention (KDA)~\citep{team2025kimi} provides efficient long-sequence mixing, with periodically interleaved Gated MLA layers preserving global interaction. Attention Residuals (AttnRes)~\citep{kimiteam2026attnres} allows each layer to selectively attend to representations from all preceding layers. Stable LatentMoE expands the routed expert space to 896 experts, with 16 activated per token, while normalization, SiTU-GLU, and Quantile Balancing stabilize optimization at extreme sparsity. These architectural advances, combined with refined data and training recipes, yield an approximately $2.5\times$ improvement in overall scaling efficiency over \kimi{2}~\cite{kimiteam2025kimik2openagentic}.

We pair this pre-training foundation with post-training designed explicitly for 1M context test-time scaling.
\kimi{3} undergoes reinforcement learning across long-horizon coding, general agents, general reasoning and knowledge tasks, each spanning multiple reasoning-effort levels.
Training environments include verifiable search and professional knowledge work, software engineering and kernel optimization, multimodal reasoning with vision-in-the-loop tool use, persistent assistant workflows, web development, and autonomous execution tasks.
These environments train a general loop of reasoning, acting, observing, verifying, and adapting, often over hundreds or thousands of tool calls and millions of accumulated context tokens.
Domain- and effort-specialized policies are consolidated into a unified model through multi-teacher on-policy distillation~\citep{lu2025onpolicydistillation, xiao2026mimov2flash, deepseekv4}.

Realizing this regime requires infrastructure that scales with architecture complexity, model size, and trajectory length. 
For systems co-design for KDA, we develop fused kernels, KDA Context Parallelism, and state-aware prefix caching to make KDA efficient within devices, across devices, and across requests. 
For 2.8T-parameter MoE pre-training, MoonEP provides perfectly balanced expert execution with static computation shapes and zero-copy communication, while memory efficient training and multimodal encoder optimizations sustain utilization within bounded memory. 
For million-token agentic RL, our co-located system combines partial rollouts, external KV-cache retention, adaptive throttling and resumable microVM sandboxes to preserve long-lived model and environment state. 
Finally, specialized kernels, and cache- and budget-aware fleet scheduling translate these innovations into predictable production serving.

The resulting model establishes a new open frontier.
On benchmarks spanning long-horizon
coding, agentic, knowledge, reasoning, and vision tasks, \kimi{3} trails the strongest proprietary systems overall---Claude Fable 5 and GPT-5.6 Sol---and is consistently ahead of the other open and proprietary models evaluated in our suite, as shown in Fig.~\ref{fig:kimi-k3-results}.

Our contributions are summarized as follows:

\begin{itemize}[leftmargin=12pt, topsep=0pt]
    \item \textbf{Pre-training at the open frontier.}
          We train a 2.8T-parameter native multimodal MoE model with 104B activated parameters and a 1M-token context window.
          KDA, AttnRes, Stable LatentMoE, refined data and training recipes collectively improve overall scaling efficiency by approximately $2.5\times$ over \kimi{2}.

    \item \textbf{Reinforcement learning for multi-effort test-time scaling.}
          We conduct RL across general, agentic, and coding domains and multiple reasoning-effort levels, then consolidate the resulting capabilities into a unified model.

    \item \textbf{Infrastructure for multi-trillion-parameter, million-token intelligence.}
          We introduce KDA systems co-designs; MoonEP and memory-efficient infrastructure for 2.8T-parameter MoE pre-training; a co-located RL system with resumable sandboxes for million-token agentic trajectories; and more infrastructure innovations.

    \item \textbf{An open frontier model.}
          We release the full \kimi{3} model weights, making frontier intelligence available for research, deployment, and further innovation.

\end{itemize}

\section{Model Architecture}
\label{sec:arch}

\begin{figure}[t]
    \centering
    \resizebox{0.9\textwidth}{!}{\input{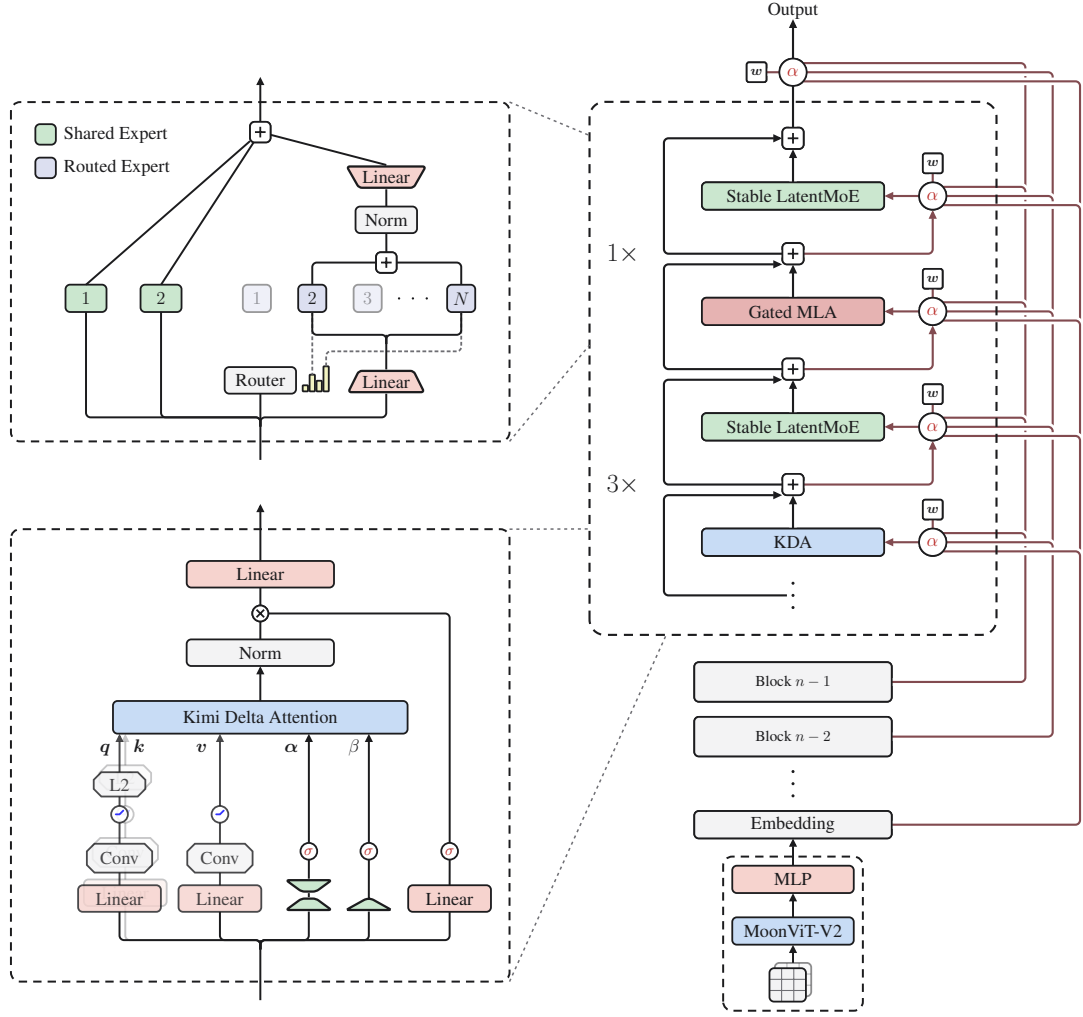}}
    \caption{The \kimi{3} architecture, organized around token, channel, and layer mixing, with a native vision pathway at the input.
        Each block contains three Kimi Delta Attention (KDA) layers followed by one Gated MLA layer, with each attention layer paired with a Stable LatentMoE feed-forward network.
        Attention Residuals (AttnRes) use learned pseudo-queries ($\bm{w}$) to derive attention weights ($\alpha$) over the embedding and preceding block outputs, enabling selective information flow across depth.
        \textbf{Top left}: the Stable LatentMoE module with shared and routed experts.
        \textbf{Bottom left}: the KDA module.
        \textbf{Bottom right}: the native vision pathway.}
    \label{fig:arch}
\end{figure}

The \kimi{3} architecture is designed to scale information flow along three complementary dimensions: sequence length, network depth, and model width.
Along the sequence dimension, Hybrid Attention combines three Kimi Delta Attention (KDA)~\citep{team2025kimi} layers with one Gated MLA layer in each block, providing an efficient mechanism for long-context token mixing while retaining selective high-capacity attention (\S\ref{sec:hybrid-attn}).
Along the depth dimension, Attention Residuals (AttnRes)~\citep{kimiteam2026attnres} enable each module to selectively retrieve representations from the embedding, the current block, and preceding blocks, extending information access beyond conventional sequential residual accumulation (\S\ref{sec:attnres}).
Along the width dimension, each attention layer is followed by a Stable LatentMoE layer that performs sparse channel mixing, effectively activating 16 of 896 routed experts for each token (\S\ref{sec:stable-latent-moe}).
For native vision, MoonViT-V2 encodes images and videos, and a lightweight projector maps the resulting visual features into the shared embedding space before backbone processing (\S\ref{sec:native-vision}).
Together with Per-Head Muon (\S\ref{sec:perhead_muon}), these components provide a unified architecture for scaling information flow across tokens, layers, and channels. Combined with refined training and data recipes, they yield an approximately $2.5\times$ improvement in overall scaling efficiency over \kimi{2}.
Figure~\ref{fig:arch} provides an overview of the architecture.

\subsection{Hybrid Attention}
\label{sec:hybrid-attn}

\kimi{3} uses a layerwise hybrid of linear and global attention, combining KDA~\citep{team2025kimi} with Gated MLA.
Each block contains 3 KDA layers followed by 1 Gated MLA layer, giving a $3{:}1$ mixing ratio.
This pattern is repeated throughout the backbone.
The two attention mechanisms are described separately below. An additional Gated MLA layer is placed at the end of the backbone, ensuring that the final layer always performs global attention.

\subsubsection{Kimi Delta Attention}
\label{sec:kda}

KDA extends the delta-rule recurrence~\citep{schlag-2021-deltanet,yang-2025-gdn} with a channel-wise forget gate~\citep{team2025kimi}.
Consider a sequence of hidden states $\bm{x}_t\in\mathbb{R}^d$, where $t$ indexes the token position and $d$ is the model hidden dimension.
For clarity, we first describe a single attention head, with query and key vectors $\bm{q}_t,\bm{k}_t\in\mathbb{R}^{d_k}$, value vector $\bm{v}_t\in\mathbb{R}^{d_v}$, and recurrent state $\mathbf{S}_t\in\mathbb{R}^{d_k\times d_v}$.
KDA applies channel-wise decay before the delta-rule update:
\begin{equation}
    \mathbf{S}_t
    = \left(\mathbf{I}-\beta_t\bm{k}_t\bm{k}_t^{\top}\right)
    \operatorname{Diag}(\bm{\alpha}_t)\mathbf{S}_{t-1}
    + \beta_t\bm{k}_t\bm{v}_t^{\top},
    \qquad
    \tilde{\bm{o}}_t = \mathbf{S}_t^{\top}\bm{q}_t.
    \label{eq:recurrent_KDA}
\end{equation}
Here, $\bm{\alpha}_t\in(0,1)^{d_k}$ is the channel-wise one-step retention factor, and $\beta_t\in(0,1)$ controls the delta-rule write strength.

Following Kimi Linear~\citep{team2025kimi}, KDA parameterizes the per-head quantities as
\begin{align}
    \bm{q}_t^h,\bm{k}_t^h
     & = \operatorname{L_2Norm}\!\left(\operatorname{Swish}\!\left(\operatorname{ShortConv}\!\left(\mathbf{W}_{q/k}^h\bm{x}_t\right)\right)\right)
    \in \mathbb{R}^{d_k}, \nonumber                                                                                                                \\
    \bm{v}_t^h
     & = \operatorname{Swish}\!\left(\operatorname{ShortConv}\!\left(\mathbf{W}_v^h\bm{x}_t\right)\right)
    \in \mathbb{R}^{d_v}, \nonumber                                                                                                                \\
    \beta_t^h
     & = \operatorname{Sigmoid}\!\left(\mathbf{W}_{\beta}^h\bm{x}_t\right)
    \in (0,1), \nonumber                                                                                                                           \\
    \bm{z}_t^h
     & = \mathbf{W}_{\alpha}^{\uparrow}\mathbf{W}_{\alpha}^{\downarrow}\bm{x}_t + \bm{b}_{\alpha}^h
    \in \mathbb{R}^{d_k}.
    \label{eq:kda-param}
\end{align}
The query, key, and value projections apply $\operatorname{ShortConv}$ followed by $\operatorname{Swish}$~\citep{yang-2025-gdn}, and the query and key are further normalized with $\operatorname{L_2Norm}$~\citep{yang-2024-parallelizing}.
The low-rank projection and head-specific bias $\bm{b}_{\alpha}^h\in\mathbb{R}^{d_k}$ produce a fine-grained decay logit $\bm{z}_t^h$ for each key channel.
The lower-bounded mapping from $\bm{z}_t^h$ to $\bm{\alpha}_t^h$ is introduced after the chunkwise formulation below.

\paragraph{Chunkwise parallel form}
Following Kimi Linear~\citep{team2025kimi}, KDA is recurrent across chunks and parallel within each chunk.
For a chunk size $C$, $\mathbf{X}_{[t]}$ stacks the token vectors in the $t$-th chunk for $\mathbf{X}\in\{\mathbf{Q},\mathbf{K},\mathbf{V},\mathbf{O},\mathbf{U},\mathbf{W}\}$.
The matrix $\mathbf{S}_{[t]}\in\mathbb{R}^{d_k\times d_v}$ denotes the recurrent state entering chunk $t$.
For positions $1\le i\le j\le C$, define the channel-wise cumulative decay
\begin{equation}
    \bm{\gamma}_{[t]}^{i\rightarrow j}
    := \prod_{r=i}^{j}\bm{\alpha}_{[t]}^r,
    \qquad
    \bm{\gamma}_{[t]}^r
    := \bm{\gamma}_{[t]}^{1\rightarrow r}.
    \label{eq:kda-cumulative-decay}
\end{equation}
As in Kimi Linear, $\bm{\Gamma}_{[t]}^{1\rightarrow C}\in\mathbb{R}^{C\times d_k}$ stacks $\bm{\gamma}_{[t]}^1,\ldots,\bm{\gamma}_{[t]}^C$ row-wise.
The UT transform produces $\mathbf{U}_{[t]}$ and $\mathbf{W}_{[t]}$, from which we define the pseudo-value term $\widetilde{\mathbf{V}}_{[t]}:=\mathbf{U}_{[t]}-\mathbf{W}_{[t]}\mathbf{S}_{[t]}$.
Given the incoming state $\mathbf{S}_{[t]}$, all outputs in chunk $t$ are computed in parallel as
\begin{align}
    \mathbf{A}_{[t]}
     & = \operatorname{Tril}\!\left[
                                  (\mathbf{Q}_{[t]}\odot \bm{\Gamma}_{[t]}^{1\rightarrow C})
                                  (\mathbf{K}_{[t]}/\bm{\Gamma}_{[t]}^{1\rightarrow C})^{\top}
                                  \right], \nonumber                        \\
    \mathbf{O}_{[t]}
     & = \underbrace{(\bm{\Gamma}_{[t]}^{1\rightarrow C}\odot\mathbf{Q}_{[t]})\mathbf{S}_{[t]}}_{\text{inter-chunk}}
    + \underbrace{\mathbf{A}_{[t]}\widetilde{\mathbf{V}}_{[t]}}_{\text{intra-chunk}}.
    \label{eq:kda-chunkwise}
\end{align}
For a matrix $\mathbf{M}$, $\operatorname{Tril}(\mathbf{M})$ sets all strictly upper-triangular entries to zero and retains the lower-triangular entries, including the diagonal.
This mask enforces causal interactions within the chunk, and the diagonal is retained because each output reads the state after the current-token update.
The first term in $\mathbf{O}_{[t]}$ carries information from preceding chunks, whereas the second term accounts for interactions within the current chunk.
We refer readers to Kimi Linear~\citep{team2025kimi} for the UT transform and the full derivation of the chunkwise form.

\paragraph{Lower-bounded decay}
Eq.~\ref{eq:kda-chunkwise} rescales the keys in each chunk by the reciprocal cumulative decay $1/\bm{\Gamma}_{[t]}^{1\rightarrow C}$.
Because $\bm{\Gamma}_{[t]}^{1\rightarrow C}$ is a product of retention factors in $(0,1)$, this reciprocal can grow without bound and overflow in finite precision~\citep{yang-etal-2024-gla,team2025kimi}.
Kimi Linear controls this numerical range by computing relative decay in log space and dividing each chunk into secondary $16$-token tiles~\citep{yang-etal-2024-gla,team2025kimi}.
The off-diagonal tiles can then be computed with dense matrix multiplications on Tensor Cores directly.
The diagonal tiles, in contrast, still require explicit position-pair computations, which remain the main intra-chunk bottleneck.

\begin{figure}[t]
    \centering
    \definecolor{kdafigink}{HTML}{1C1C1E}
\definecolor{kdafiggray}{HTML}{777777}
\definecolor{kdafiggrid}{HTML}{D6D6D8}
\definecolor{kdafigred}{HTML}{B92622}
\definecolor{kdafigblue}{HTML}{C8DCF5}
\definecolor{kdafigbluedark}{HTML}{4779B8}
\definecolor{kdafigorange}{HTML}{F2C6A0}
\definecolor{kdafigorangedark}{HTML}{B96A2D}

\begin{subfigure}[t]{0.47\textwidth}
    \centering
    \resizebox{\linewidth}{!}{%
        \begin{tikzpicture}[
                x=1pt,
                y=1pt,
                outer sep=0pt,
                labeltext/.style={font=\footnotesize, text=kdafigink},
                smalltext/.style={font=\scriptsize, text=kdafigink},
                axis/.style={draw=kdafigink, line width=0.55pt, -{Latex[length=4pt,width=3.5pt]}},
            ]
            \path[use as bounding box] (-18,0) rectangle (245,158);

            \node[smalltext, text=kdafiggray, anchor=north west] (linearlegend) at (65,150)
            {Kimi Linear: $g=-e^A\operatorname{Softplus}(z)$};
            \node[smalltext, text=kdafigred, anchor=north west] at ($(linearlegend.south west)+(0,-1pt)$)
            {Kimi K3: $g=g_{\min}\operatorname{Sigmoid}(e^A z)$};

            \draw[axis] (28,31) -- (28,144) node[above, labeltext] {$g$};
            \draw[axis] (28,31) -- (232,31) node[right, labeltext] {$z$};
            \draw[draw=kdafiggrid, line width=0.45pt, dashed] (28,113) -- (225,113);
            \draw[draw=kdafigred, line width=0.55pt, dashed] (28,53) -- (225,53);
            \node[smalltext, anchor=east] at (24,113) {$0$};
            \node[smalltext, text=kdafigred, anchor=east] at (24,53) {$g_{\min}=-5$};

            \draw[draw=kdafiggray, line width=1pt]
            plot[domain=-5.5:6, samples=120]
                ({126.5 + 16.4167*\x}, {113 - 12*ln(1 + exp(\x))});
            \node[smalltext, text=kdafiggray, anchor=north east] at (230,42) {$-\infty$};
            \draw[draw=kdafigred, line width=1.2pt]
            plot[domain=-5.5:6, samples=120]
                ({126.5 + 16.4167*\x}, {113 - 60/(1 + exp(-\x))});
            \node[smalltext, text=kdafiggray, anchor=north west] at (32,28) {$A=0$};
        \end{tikzpicture}%
    }
    \caption{Log-decay parameterization.}
    \label{fig:kda-lower-bound-mapping}
\end{subfigure}
\hfill
\begin{subfigure}[t]{0.47\textwidth}
    \centering
    \resizebox{\linewidth}{!}{%
        \begin{tikzpicture}[
                x=1pt,
                y=1pt,
                outer sep=0pt,
                labeltext/.style={font=\footnotesize, text=kdafigink},
                smalltext/.style={font=\scriptsize, text=kdafigink},
            ]
            \path[use as bounding box] (0,0) rectangle (226,150);

            \begin{scope}[xshift=-5pt]
                \path[fill=kdafigorange] (12,120) rectangle (32,100);
                \path[fill=kdafigblue] (12,100) rectangle (32,80);
                \path[fill=kdafigorange] (32,100) rectangle (52,80);
                \path[fill=kdafigblue] (12,80) rectangle (52,60);
                \path[fill=kdafigorange] (52,80) rectangle (72,60);
                \path[fill=kdafigblue] (12,60) rectangle (72,40);
                \path[fill=kdafigorange] (72,60) rectangle (92,40);
                \draw[draw=kdafigink, line width=0.7pt] (12,40) rectangle (92,120);
                \foreach \i in {1,...,3} {
                        \draw[draw=kdafigink, line width=0.4pt] ({12+20*\i},40) -- ({12+20*\i},120);
                        \draw[draw=kdafigink, line width=0.4pt] (12,{40+20*\i}) -- (92,{40+20*\i});
                    }
            \end{scope}
            \node[fit={(7,40) (87,120)}, inner sep=0pt] (lineargrid) {};

            \begin{scope}[xshift=5pt]
                \path[fill=kdafigblue] (134,120) rectangle (154,100);
                \path[fill=kdafigblue] (134,100) rectangle (174,80);
                \path[fill=kdafigblue] (134,80) rectangle (194,60);
                \path[fill=kdafigblue] (134,60) rectangle (214,40);
                \draw[draw=kdafigink, line width=0.7pt] (134,40) rectangle (214,120);
                \foreach \i in {1,...,3} {
                        \draw[draw=kdafigink, line width=0.4pt] ({134+20*\i},40) -- ({134+20*\i},120);
                        \draw[draw=kdafigink, line width=0.4pt] (134,{40+20*\i}) -- (214,{40+20*\i});
                    }
            \end{scope}
            \node[fit={(139,40) (219,120)}, inner sep=0pt] (k3grid) {};

            \node[labeltext, above=5pt of lineargrid] {Kimi Linear};
            \node[labeltext, above=5pt of k3grid] {Kimi K3};

            \coordinate (decaymid) at ($(lineargrid.east)!0.5!(k3grid.west)$);
            \draw[draw=kdafigred, line width=0.7pt, -{Latex[length=4pt,width=3pt]}]
            ($(lineargrid.east)+(8pt,0)$) -- ($(k3grid.west)+(-8pt,0)$);
            \node[font=\scriptsize, text=kdafigred, above=4pt of decaymid] {lower-bounded};

            \coordinate (legendcenter) at ($(lineargrid.south)!0.5!(k3grid.south)+(0,-10pt)$);
            \node[draw=kdafigink, fill=kdafigorange, line width=0.4pt, minimum size=8pt, inner sep=0pt, anchor=north]
            (pairlegend) at ($(legendcenter)+(-39pt,0)$) {};
            \node[smalltext, right=2pt of pairlegend] (pairlabel) {Position-pair Diagonal};
            \node[draw=kdafigink, fill=kdafigblue, line width=0.4pt, minimum size=8pt, inner sep=0pt, below=3pt of pairlegend]
            (tclegend) {};
            \node[smalltext, right=2pt of tclegend] (tclabel) {Tensor Core};
        \end{tikzpicture}%
    }
    \caption{Diagonal-tile computation.}
    \label{fig:kda-lower-bound-tiles}
\end{subfigure}
    \caption{Lower-bounded decay and its effect on chunkwise KDA computation.
        \textbf{(a)} Kimi Linear uses an unbounded negative-Softplus mapping, whereas \kimi{3} bounds the log-decay with a scaled sigmoid; the curves show $A=0$ and $g_{\min}=-5$.
        \textbf{(b)} Kimi Linear evaluates each diagonal tile with an explicit position-pair computation, while the bounded range in \kimi{3} allows all causal tiles to use dense Tensor Core matrix multiplications.}
    \label{fig:kda-lower-bound}
\end{figure}
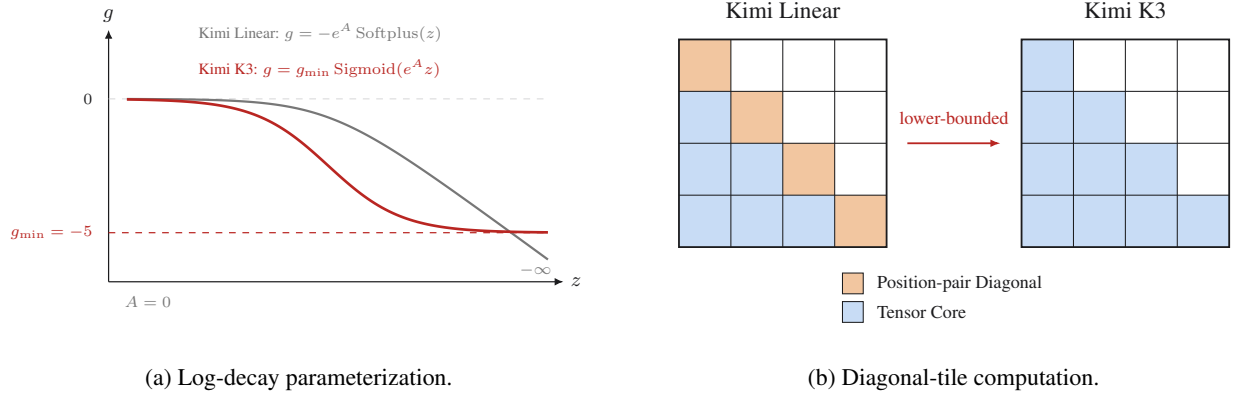

\kimi{3} addresses this bottleneck by changing the mapping from the decay logits $\bm{z}_t^h$ to the per-step log-decay $\bm{g}_t^h$.
Following GDN and Mamba-2, Kimi Linear uses the negative-Softplus mapping $\bm{g}_t^h=-e^{A_h}\operatorname{Softplus}(\bm{z}_t^h)\in(-\infty,0)^{d_k}$~\citep{yang-2025-gdn,mamba2,team2025kimi}.
\kimi{3} instead uses a scaled sigmoid to bound the log-decay from below:
\begin{align}
    \bm{g}_t^h
     & = g_{\min}\operatorname{Sigmoid}\!\left(e^{A_h}\bm{z}_t^h\right)
    \in (g_{\min},0)^{d_k}, \nonumber                                   \\
    \bm{\alpha}_t^h
     & = \exp(\bm{g}_t^h)
    \in \left(e^{g_{\min}},1\right)^{d_k},
    \label{eq:kda-forget-gate}
\end{align}
where $A_h$ is a learnable per-head log-scale and $g_{\min}=-5$ is fixed.
We initialize $A_h=0$, and each bias $\bm{b}_{\alpha}^h$ is initialized following~\citep{team2025kimi,mamba2,yang-2025-gdn}.
With $g_{\min}=-5$, every retention factor satisfies $\alpha_{t,j}^h>e^{-5}\approx6.7\times10^{-3}$, and the cumulative log-decay over a $16$-token tile lies in $(-80,0)$.
The corresponding reciprocal rescaling factor is therefore smaller than $e^{80}$ and remains within the BF16 dynamic range.
This finite range allows both diagonal and off-diagonal tiles to use dense Tensor Core matrix multiplications, eliminating the position-pair diagonal path.
This parameterization is closely related to the lower-bounded recurrence gates in prior work~\citep{qin2024hgrn2,de2024griffin,peng-2025-rwkv7}.
Fig.~\ref{fig:kda-lower-bound} illustrates the change in decay parameterization and its computational consequence.

\paragraph{Full-rank gate}
Finally, \kimi{3} changes KDA's output gate from the low-rank parameterization used by Kimi Linear~\citep{team2025kimi} to an input-dependent full-rank projection.
After applying head-wise RMSNorm~\citep{zhang2019root} to the recurrent output, KDA applies data-dependent output gating~\citep{qiu2025gated}:
\begin{equation}
    \bm{y}_t
    = \mathbf{W}_o\!\left[
        \operatorname{Sigmoid}\!\left(\mathbf{W}_g\bm{x}_t\right)
        \odot \operatorname{RMSNorm}(\tilde{\bm{o}}_t)
        \right].
    \label{eq:kda-output}
\end{equation}

\subsubsection{Gated MLA}
\label{sec:gated-mla}

Multi-head Latent Attention (MLA), introduced in DeepSeek-V2~\citep{deepseekv2}, compresses the key--value representation of each token into a low-dimensional latent vector $\bm{c}_t=\mathbf{W}_{c}\bm{x}_t$.
Instead of caching full head-specific keys and values, MLA caches $\bm{c}_t$ and reconstructs the content keys and values through learned up-projections during attention computation.
This factorization reduces the KV-cache footprint while retaining global token-to-token attention.
MLA was subsequently adopted by \kimi{2} and \kimi{2.5}~\citep{kimiteam2025kimik2openagentic,kimik25}, and \kimi{3} retains it in the periodic global-attention layers.

Unlike \kimi{2} and \kimi{2.5}, \kimi{3} follows the hybrid design of Kimi Linear~\citep{team2025kimi} and applies No Position Encoding (NoPE) to all MLA layers.
Consequently, no explicit positional encoding is applied to their queries or keys.
The intervening KDA layers provide position-sensitive and recency-aware sequence mixing, while the MLA layers provide unrestricted global content interaction.
This separation also avoids modifying positional-encoding parameters when extending the context length, such as retuning a RoPE frequency base or applying YaRN~\citep{peng2023yarn}.

In addition, \kimi{3} augments MLA with an input-dependent, channel-wise full-rank output gate.
Let $\tilde{\bm{o}}_t$ denote the ungated MLA output at position $t$; the gated output is
\begin{equation}
    \bm{y}_t
    = \mathbf{W}_o\!\left[
        \operatorname{Sigmoid}\!\left(\mathbf{W}_g\bm{x}_t\right)
        \odot \tilde{\bm{o}}_t
        \right].
    \label{eq:gated-mla-output}
\end{equation}
The gate projection $\mathbf{W}_g$ is full rank, matching the new parameterization used by KDA in \kimi{3}. This gate allows each token to modulate the channels read from global attention~\citep{qiu2025gated}.

To correct the biased rounding error identified by~\citep{qiu2025lowprecision} in flash attention, we keep the attention output in FP32 during training.
This choice doubles the on-chip footprint of the output tile; we therefore redesign the training kernel to overlap it with the KV staging buffers instead of the query tile, freeing shared memory for a deeper KV pipeline and higher training throughput.

\subsection{Attention Residuals}
\label{sec:attnres}

Standard residual connections~\citep{he2015resnet} compress all prior information into a single state $\bm{h}_l$ over depth --- a bottleneck reminiscent of RNNs over time. For sequence modeling, the Transformer replaced recurrence with attention~\citep{bahdanau2014attention,vaswani-2017-attention}, allowing each position to selectively access all previous positions with data-dependent weights. Attention Residuals (AttnRes)~\citep{kimiteam2026attnres} applies the same methodology to depth: each layer selectively retrieves representations from all preceding layers rather than accumulating them uniformly.

\paragraph{Full Attention Residuals} For each layer $l$, we define a layer-specific learnable pseudo-query $\bm{q}_l = \bm{w}_l \in \mathbb{R}^d$ and keys and values
\begin{equation}
    \bm{k}_{i} = \bm{v}_{i} = \begin{cases} \bm{h}_1 & i = 0 \\ f_i(\bm{h}_{i}) & 1 \leq i \leq l-1 \end{cases}
    \label{eq:attnres-qkv}
\end{equation}
where $f_i(\bm{h}_i)$ is the output of layer $i$ and $\bm{h}_1$ is the token embedding. The attention weights follow a softmax kernel $\phi(\bm{q}, \bm{k}) = \exp\left(\bm{q}^\top\operatorname{RMSNorm}(\bm{k})\right)$~\citep{katharopoulos-2020-transformers,zhang2019root}, where the $\operatorname{RMSNorm}$ prevents layers with large-magnitude outputs from dominating the weights:
\begin{equation}
    {\alpha_{i \to l}} = \frac{\phi\left(\bm{q}_{l}, \bm{k}_{i}\right)}{\sum_{j=0}^{l-1} \phi\left(\bm{q}_{l}, \bm{k}_{j}\right)},
    \qquad
    \bm{h}_{l} = \sum_{i=0}^{l-1} {\alpha_{i \to l}} \cdot \bm{v}_{i}.
    \label{eq:attnres-full}
\end{equation}
Since network depth is modest ($L < 100$), the $O(L^2 d)$ arithmetic of this \emph{full} form is affordable; the practical overhead is the $O(Ld)$ memory (and cross-stage communication under pipeline parallelism) for keeping all layer outputs alive.

\paragraph{Block Attention Residuals} To reduce this overhead, we partition the $L$ layers into $N$ blocks of $S = L/N$ layers each. Within block $n$ (layer indices $\mathcal{B}_n$), layer outputs are reduced to a single representation by summation, $\bm{b}_n = \sum_{j \in \mathcal{B}_n} f_j(\bm{h}_j)$, with $\bm{b}_n^i$ denoting the partial sum over the first $i$ layers of the block; we set $\bm{b}_0 = \bm{h}_1$ so the token embedding is always included as a source. Across blocks, full attention is applied over only the $N$ block-level representations: for the $i$-th layer in block $n$, the value matrix is
\begin{equation}
    \mathbf{V} = \begin{cases}
        [\bm{b}_0, \bm{b}_1, \ldots, \bm{b}_{n-1}]^\top                 & \text{if } i = 1 \text{ (first layer of block } n\text{)} \\
        [\bm{b}_0, \bm{b}_1, \ldots, \bm{b}_{n-1}, \bm{b}_n^{i-1}]^\top & \text{if } i \geq 2 \text{ (subsequent layers)}
    \end{cases}
    \label{eq:attnres-block}
\end{equation}
with keys and attention weights following Eq.~\ref{eq:attnres-qkv} and Eq.~\ref{eq:attnres-full}. The final output layer then aggregates all $N$ block representations. Under Block AttnRes, memory and communication overhead drop from $O(Ld)$ to $O(Nd)$, while this block structure also bounds the inference-time state, enabling the parallel inter-block results to be better merged with the sequential intra-block partial sums via online softmax~\citep{milakov2018online}, significantly reducing inference time cost.

Empirically, $N \approx 8$ recovers most of the benefit across model scales~\citep{kimiteam2026attnres}; for \kimi{3}, we partition its layers into 8 blocks with 12-layer size, giving a partial final block and 9 total blocks when counting the embedding layer.

\subsection{Stable LatentMoE}
\label{sec:stable-latent-moe}

Increasing both the expert pool and the number of active experts expands the space of expert specializations, but in a conventional MoE each selected expert receives the full $d$-dimensional token representation, so communication and expert-weight traffic grow with the routing multiplicity.
LatentMoE~\citep{elango2026latentmoe} makes this expansion affordable by separating the full model width from the routed-expert width: shared experts retain a full-width path for common transformations, whereas specialized routed experts operate in a compact latent space of width $\ell$. This enables \kimi{3} to scale channel mixing to 896 routed experts with 16 active experts per token, corresponding to a sparsity of 56.

This extreme sparsity amplifies two failure modes of the vanilla design.
First, the routed path composes $\mathbf{W}^{\downarrow}$, a gated multi-branch expert feed-forward network, and $\mathbf{W}^{\uparrow}$ into a chain of nearly four consecutive matrix multiplications.
This ill-conditioned structure, combined with the 2.8-trillion-parameter scale, produces exploding internal activations in the routed branch.
Second, balancing the load of nearly $10^3$ experts exceeds the regime in which existing auxiliary-loss-free bias updates remain well behaved.
Stable LatentMoE addresses these two failure modes with three components: an RMSNorm before the up-projection and Sigmoid Tanh Unit GLU (SiTU-GLU) to suppress activation explosion, and Quantile Balancing (QB) for load balancing.

As illustrated in Fig.~\ref{fig:arch}, the layer follows the shared- and routed-expert organization of DeepSeekMoE~\citep{dai2024deepseekmoe}.
For $\bm{x}\in\mathbb{R}^{d}$, the shared experts process $\bm{x}$ directly, while the routed path projects it to $\bm{z}=\mathbf{W}^{\downarrow}\bm{x}\in\mathbb{R}^{\ell}$, dispatches $\bm{z}$ to the selected experts, and maps their weighted aggregate back to $\mathbb{R}^{d}$ through $\mathbf{W}^{\uparrow}$:
\begin{align}
    \bm{u}
     & = \sum_{i\in\mathcal{T}_k(\bm{x})} p_i E_i^{\mathrm{routed}}(\mathbf{W}^{\downarrow}\bm{x}), \nonumber \\
    \bm{y}
     & = \sum_{j=1}^{N_s} E_j^{\mathrm{shared}}(\bm{x})
    + \mathbf{W}^{\uparrow}\operatorname{RMSNorm}(\bm{u}).
    \label{eq:latentmoe}
\end{align}
Here, $\bm{u}\in\mathbb{R}^{\ell}$ is the aggregated routed representation, $E_j^{\mathrm{shared}}:\mathbb{R}^{d}\!\rightarrow\!\mathbb{R}^{d}$ and $E_i^{\mathrm{routed}}:\mathbb{R}^{\ell}\!\rightarrow\!\mathbb{R}^{\ell}$ are the shared and routed expert feed-forward networks, and $p_i$ is the router weight defined by the Quantile Balancing rule below.
\kimi{3} fixes the number of full-width shared experts to $N_s=2$ in every layer.

\subsubsection{Normalized LatentMoE}
The original LatentMoE directly applies $\mathbf{W}^{\uparrow}$ to the aggregated routed representation $\bm{u}$, whose scale can vary with the selected experts and their routing weights.
As shown in Eq.~\ref{eq:latentmoe}, \kimi{3} instead inserts RMSNorm~\citep{zhang2019root} between expert aggregation and the up-projection.
This normalization reduces the sensitivity of the routed branch to scale variation before it is combined with the full-width shared branch.
Beyond stabilizing training, the additional RMSNorm consistently improves validation loss and downstream benchmarks.

\subsubsection{Sigmoid Tanh Unit GLU}
\label{sec:situ}
\begin{figure}[t]
    \centering
    \definecolor{situfigink}{HTML}{1C1C1E}
\definecolor{situfiggray}{HTML}{777777}
\definecolor{situfiglightgray}{HTML}{AAAAAA}
\definecolor{situfigred}{HTML}{B92622}
\definecolor{situfiggreen}{HTML}{3E7A55}

\begingroup
\newcommand{\siturowcell}[1]{\multirow{1}{*}{#1}}
\setlength{\tabcolsep}{2pt}
\renewcommand{\arraystretch}{1.2}
\begin{tabular*}{\linewidth}{@{\extracolsep{\fill}}m{0.11\textwidth}>{\centering\arraybackslash}m{0.26\textwidth}>{\centering\arraybackslash}m{0.11\textwidth}>{\centering\arraybackslash}m{0.47\textwidth}@{}}
    \toprule
    & Gate branch& Up branch& Curve \\
    \midrule
\end{tabular*}
\par\nointerlineskip
\renewcommand{\arraystretch}{3.0}
\footnotesize
\begin{tabular*}{\linewidth}{@{\extracolsep{\fill}}m{0.11\textwidth}>{\centering\arraybackslash}m{0.26\textwidth}>{\centering\arraybackslash}m{0.11\textwidth}>{\centering\arraybackslash}m{0.47\textwidth}@{}}
    \siturowcell{GLU~\citep{dauphin2017language}}
    & \siturowcell{$\sigma(x)$}
    & \siturowcell{$x$}
    & \multirow[c]{3}{0.47\textwidth}[-2pt]{%
        \makebox[0.47\textwidth][c]{%
            \begin{adjustbox}{width=0.37\textwidth}
                \begin{tikzpicture}[
                        x=1pt,
                        y=1pt,
                        outer sep=0pt,
                        curvelabel/.style={font=\fontsize{5}{6}\selectfont},
                        axis/.style={draw=situfigink, line width=0.4pt,
                                    -{Latex[length=3pt,width=2.2pt]}},
                    ]
                    \path[use as bounding box] (-10,-12) rectangle (148,84);
                    \coordinate (curveorigin) at (17.5,10);
                    \coordinate (curvexstart) at (3,10);   
                    \coordinate (curvexend) at (128,10);
                    \coordinate (curveyend) at (17.5,80);
                    \draw[axis] (curvexstart) -- (curvexend) node[right, curvelabel, text=situfigink] {$x$};
                    \draw[axis] (17.5,4) -- (curveyend) node[left, curvelabel, text=situfigink] {$f(x)$};
                    \foreach \value/\xpos in {-10/7,50/70,100/122.5} {
                            \draw[draw=situfigink, line width=0.3pt] (\xpos,8.5) -- (\xpos,11.5);
                            \node[curvelabel, text=situfigink, anchor=north]
                            at (\xpos,8) {$\value$};
                        }
                    \draw[draw=situfigink, line width=0.3pt] (17.5,8.5) -- (17.5,11.5);
                    \node[curvelabel, text=situfigink, anchor=north west] at (15.5,8) {$0$};
                    \draw[draw=situfigink, line width=0.3pt] (16,75) -- (19,75);
                    \node[curvelabel, text=situfigink, anchor=east] at (15,75) {$100$};

                    \coordinate (situboundleft) at (17.5,75);
                    \coordinate (situboundright) at (128,75);
                    \draw[draw=situfigred, line width=0.35pt, dashed]
                    (situboundleft) -- (situboundright);
                    \node[curvelabel, text=situfigred, anchor=south east]
                    (situboundlabel) at (140,76)
                    {$|f(x)|\leq\beta_1\beta_2=100$};

                    \begin{scope}
                        \clip (7,4) rectangle (128,79);

                        \draw[draw=situfiggreen, line width=0.5pt]
                        plot[domain=-10:0,samples=60]
                            ({17.5+1.05*\x},{10+0.65*\x*\x*exp(\x)/(1+exp(\x))})
                        plot[domain=0:11,samples=80]
                            ({17.5+1.05*\x},{10+0.65*\x*\x/(1+exp(-\x))});

                        \draw[draw=situfigred, line width=0.55pt]
                        plot[domain=-10:0,samples=60]
                            ({17.5+1.05*\x},
                            {10+65*(2*exp(\x/2)/(1+exp(\x/2))-1)
                                *(2*exp(2*\x/25)/(1+exp(2*\x/25))-1)
                                *exp(\x)/(1+exp(\x))})
                        plot[domain=0:100,samples=180]
                            ({17.5+1.05*\x},
                            {10+65*(2/(1+exp(-\x/2))-1)
                                *(2/(1+exp(-2*\x/25))-1)
                                /(1+exp(-\x))});

                        \draw[draw=situfiggray, line width=0.5pt, dashed]
                        plot[domain=-10:0,samples=60]
                            ({17.5+1.05*\x},{10+0.65*\x*exp(\x)/(1+exp(\x))})
                        plot[domain=0:104,samples=170]
                            ({17.5+1.05*\x},{10+0.65*\x/(1+exp(-\x))});
                    \end{scope}

                    \draw[draw=situfiggray, line width=0.5pt, densely dotted]
                    (11.2,9.4) rectangle (23.8,17.8);
                    \draw[draw=situfiggray, line width=0.5pt, densely dotted]
                    (23.8,17.8) -- (46,45);
                    \draw[draw=situfiggray, line width=0.5pt, densely dotted]
                    (23.8,9.4) -- (46,13);
                    \draw[draw=situfiggray, line width=0.4pt, fill=white]
                    (46,13) rectangle (94,45);
                    \begin{scope}
                        \clip (46.8,13.8) rectangle (93.2,44.2);
                        \draw[draw=situfiggreen, line width=0.5pt]
                        plot[domain=-6:0,samples=40]
                            ({70+4*\x},{15.46+2.46*\x*\x*exp(\x)/(1+exp(\x))})
                        plot[domain=0:6,samples=50]
                            ({70+4*\x},{15.46+2.46*\x*\x/(1+exp(-\x))});
                        \draw[draw=situfigred, line width=0.55pt]
                        plot[domain=-6:0,samples=40]
                            ({70+4*\x},
                            {15.46+246*(2*exp(\x/2)/(1+exp(\x/2))-1)
                                *(2*exp(2*\x/25)/(1+exp(2*\x/25))-1)
                                *exp(\x)/(1+exp(\x))})
                        plot[domain=0:6,samples=50]
                            ({70+4*\x},
                            {15.46+246*(2/(1+exp(-\x/2))-1)
                                *(2/(1+exp(-2*\x/25))-1)
                                /(1+exp(-\x))});
                        \draw[draw=situfiggray, line width=0.5pt, dashed]
                        plot[domain=-6:0,samples=40]
                            ({70+4*\x},{15.46+2.46*\x*exp(\x)/(1+exp(\x))})
                        plot[domain=0:6,samples=50]
                            ({70+4*\x},{15.46+2.46*\x/(1+exp(-\x))});
                        \draw[draw=situfiglightgray, line width=0.3pt] (46,15.46) -- (94,15.46);
                        \draw[draw=situfiglightgray, line width=0.3pt] (70,13) -- (70,45);
                    \end{scope}

                    \coordinate (swigluclip) at (26,68);
                    \node[curvelabel, text=situfiggreen, anchor=west]
                    (swiglucurvelabel) at (swigluclip) {SwiGLU};
                    \coordinate (situcurvelabelpoint) at (62,66);
                    \node[curvelabel, text=situfigred, anchor=north]
                    (situcurvelabel) at (situcurvelabelpoint) {SiTU-GLU};
                    \coordinate (glucurvelabelpoint) at (104,56);
                    \node[curvelabel, text=situfiggray, anchor=south west]
                    (glucurvelabel) at (glucurvelabelpoint) {GLU};
                \end{tikzpicture}
            \end{adjustbox}%
        }%
    } \\
    \cmidrule{1-3}
    \siturowcell{SwiGLU~\citep{shazeer2020glu}}
    & \siturowcell{$x\cdot\sigma(x)$}
    & \siturowcell{$x$}
    & \\
    \cmidrule{1-3}
    \siturowcell{SiTU-GLU}
    & \siturowcell{$\beta_1\tanh\!\left(\frac{x}{\beta_1}\right)\cdot\sigma(x)$}
    & \siturowcell{$\beta_2\tanh\!\left(\frac{x}{\beta_2}\right)$}
    & \\
    \bottomrule
\end{tabular*}
\endgroup
    \caption{Gate and up branches of GLU, SwiGLU, and SiTU-GLU, together with their scalar responses, where $\sigma$ denotes the sigmoid function.
        Both branches receive the scalar input $x$, and all curves share the domain $x\in[-10,100]$; the inset magnifies the near-origin region.
        SiTU-GLU, shown in red with $\beta_1=4$ and $\beta_2=25$, closely follows SwiGLU near the origin and approaches the bound $|f(x)|\leq\beta_1\beta_2=100$ for large positive inputs, whereas SwiGLU remains unbounded.}
    \label{fig:situglu}
\end{figure}
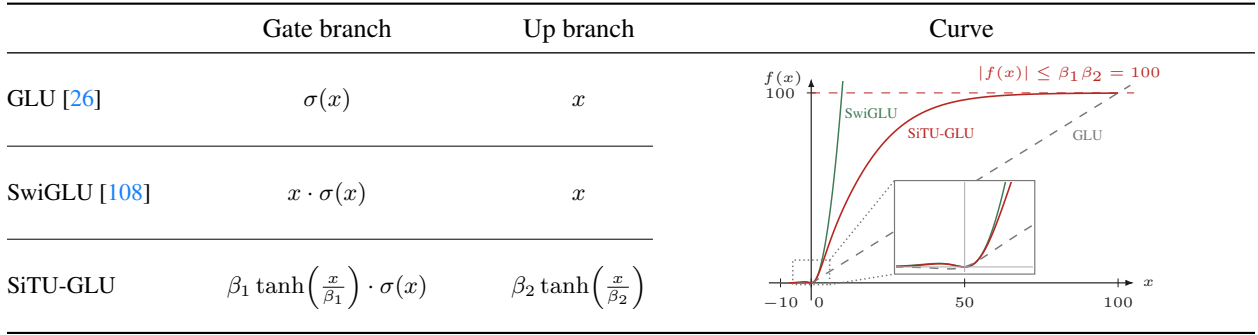
Gated Linear Units (GLUs) modulate a linear value branch with a sigmoid-activated gate, computing $\operatorname{Sigmoid}(\mathbf{W}_g\bm{x})\odot\mathbf{W}_u\bm{x}$~\citep{dauphin2017language}.
SwiGLU replaces the sigmoid gate with $\operatorname{Swish}(x)=x\operatorname{Sigmoid}(x)$ and yields strong empirical performance in Transformers~\citep{shazeer2020glu}.
SwiGLU has subsequently become a widely adopted FFN design in large language models, while a complete account of its empirical effectiveness remains open.

However, both multiplicative factors in SwiGLU are unbounded, so coincident large coordinates can produce activation outliers and increase overflow risk in low-precision arithmetic.
The sigmoid gate of the original GLU avoids unbounded gate growth, but it does not retain the approximately linear positive regime of Swish.
This motivates an activation that controls large-value growth while preserving the characteristic local and positive-side response of SwiGLU.
Other recent efforts have explored alternative parameterizations of this trade-off~\citep{jiang2026powlu}.

To satisfy these requirements, we propose Sigmoid Tanh Unit GLU (SiTU-GLU).
SiTU-GLU applies the smooth cap $\operatorname{softcap}(x,\beta)=\beta\tanh(x/\beta)$ to the linear factor of the Swish gate and independently to the up branch:
\begin{equation}
    \operatorname{SiTU\text{-}GLU}(\bm{x})
    = \left[\beta_1\tanh\!\left(\frac{\mathbf{W}_g\bm{x}}{\beta_1}\right)\odot\operatorname{Sigmoid}(\mathbf{W}_g\bm{x})\right]
    \odot \left[\beta_2\tanh\!\left(\frac{\mathbf{W}_u\bm{x}}{\beta_2}\right)\right],
    \label{eq:situglu}
\end{equation}
For \kimi{3}, we set the soft-cap hyperparameters to $\beta_1=4$ for the gate branch and $\beta_2=25$ for the up branch.
The scaled $\tanh$ is approximately linear near the origin and bounded at large magnitude, allowing SiTU-GLU to preserve the local response of SwiGLU while controlling both factors in the product.
Fig.~\ref{fig:situglu} compares the branch definitions and scalar responses of GLU, SwiGLU, and SiTU-GLU on a common slice.

\S~\ref{app:situglu} gives the local expansion, limiting case, formal output bound, and comparison with hard clamping.

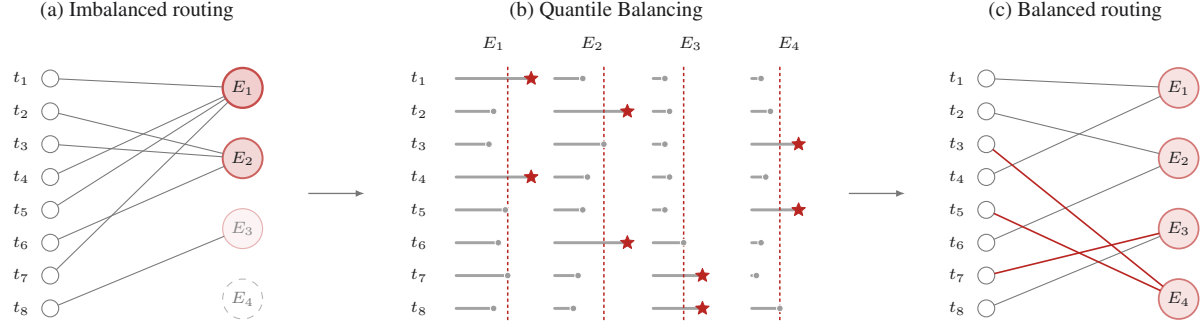
\begin{figure}[t]
    \centering
    \definecolor{qbfigink}{HTML}{1C1C1E}
\definecolor{qbfiggray}{HTML}{777777}
\definecolor{qbfiggrid}{HTML}{D6D6D8}
\definecolor{qbfigred}{HTML}{B92622}
\definecolor{qbfigredfill}{HTML}{F8E9E8}

\begin{adjustbox}{width=\textwidth,center}
        \begin{tikzpicture}[
                        x=1pt,
                        y=1pt,
                        outer sep=0pt,
                        paneltitle/.style={font=\small, text=qbfigink, align=center},
                        label/.style={font=\scriptsize, text=qbfigink},
                        small/.style={font=\scriptsize, text=qbfiggray},
                        token/.style={circle, draw=qbfiggray, fill=white, line width=0.45pt,
                                        minimum size=7.2pt, inner sep=0pt},
                        expert/.style={circle, draw=qbfigink, fill=white, line width=0.6pt,
                                        minimum size=17pt, inner sep=0pt, font=\scriptsize},
                        loadhigh/.style={expert, draw=qbfigred!82, fill=qbfigred!22,
                                        text=qbfigink, line width=1.0pt},
                        loadmedium/.style={expert, draw=qbfigred!72, fill=qbfigred!18,
                                        text=qbfigink, line width=0.85pt},
                        loadlow/.style={expert, draw=qbfigred!35, fill=qbfigred!5,
                                        text=qbfiggray!90, line width=0.5pt},
                        loadzero/.style={expert, draw=qbfiggray!55, fill=white,
                                        text=qbfiggray, line width=0.45pt, dashed},
                        balancedexpert/.style={expert, draw=qbfigred!62,
                                        fill=qbfigred!13, text=qbfigink, line width=0.75pt},
                        route/.style={draw=qbfiggray, line width=0.45pt},
                        reroute/.style={draw=qbfigred, line width=0.7pt},
                        scorebar/.style={draw=qbfiggray!68, line width=1.25pt, line cap=round},
                        adjustedpoint/.style={circle, fill=qbfiggray!72, draw=white, line width=0.25pt,
                                        minimum size=2.8pt, inner sep=0pt},
                        betacut/.style={draw=qbfigred, line width=0.65pt,
                                        dash pattern=on 1.4pt off 1.2pt},
                        qbchoice/.style={fill=qbfigred, draw=qbfigred,
                                        line width=0.1pt, line join=round},
                        flow/.style={draw=qbfiggray, line width=0.55pt,
                                                -{Latex[length=3.8pt,width=3pt]}},
                ]
                \path[use as bounding box] (8,55) rectangle (540,215);

                \node[paneltitle, anchor=north] (rawtitle) at (66,205)
                {(a) Imbalanced routing};
                \node[paneltitle, anchor=north] (quanttitle) at (266,205)
                {(b) Quantile Balancing};
                \node[paneltitle, anchor=north] (balancedtitle) at (466,205)
                {(c) Balanced routing};

                \node[token] (rawtoken1) at (29,169) {};
                \node[token, below=6.8pt of rawtoken1] (rawtoken2) {};
                \node[token, below=6.8pt of rawtoken2] (rawtoken3) {};
                \node[token, below=6.8pt of rawtoken3] (rawtoken4) {};
                \node[token, below=6.8pt of rawtoken4] (rawtoken5) {};
                \node[token, below=6.8pt of rawtoken5] (rawtoken6) {};
                \node[token, below=6.8pt of rawtoken6] (rawtoken7) {};
                \node[token, below=6.8pt of rawtoken7] (rawtoken8) {};
                \node[loadhigh] (rawexpert1) at ($(rawtoken1)+(82pt,-4pt)$) {$E_1$};
                \node[loadmedium, below=13pt of rawexpert1] (rawexpert2) {$E_2$};
                \node[loadlow, below=13pt of rawexpert2] (rawexpert3) {$E_3$};
                \node[loadzero, below=13pt of rawexpert3] (rawexpert4) {$E_4$};
                \foreach \i in {1,...,8} {
                                \node[small, text=qbfigink, anchor=east]
                                at ($(rawtoken\i.west)+(-2pt,0pt)$) {$t_{\i}$};
                        }

                \draw[route] (rawtoken1) -- (rawexpert1);
                \draw[route] (rawtoken4) -- (rawexpert1);
                \draw[route] (rawtoken5) -- (rawexpert1);
                \draw[route] (rawtoken7) -- (rawexpert1);
                \draw[route] (rawtoken2) -- (rawexpert2);
                \draw[route] (rawtoken3) -- (rawexpert2);
                \draw[route] (rawtoken6) -- (rawexpert2);
                \draw[route] (rawtoken8) -- (rawexpert3);

                \begin{scope}[xshift=12pt]
                        \node[small, text=qbfigink] (qbcol1) at (206,184) {$E_1$};
                        \node[small, text=qbfigink] (qbcol2) at (248,184) {$E_2$};
                        \node[small, text=qbfigink] (qbcol3) at (290,184) {$E_3$};
                        \node[small, text=qbfigink] (qbcol4) at (332,184) {$E_4$};
                        \foreach \y/\i in {169/1,155/2,141/3,127/4,113/5,99/6,85/7,71/8} {
                                        \node[small, text=qbfigink, anchor=east] at (181,\y) {$t_{\i}$};
                                }

                        \foreach \y/\endpoint in {169/222,155/206,141/204,127/222,
                                        113/211,99/208,85/212,71/206} {
                                        \draw[scorebar] (190,\y) -- (\endpoint,\y);
                                }
                        \foreach \y/\endpoint in {169/244,155/263,141/253,127/246,
                                        113/244,99/263,85/242,71/240} {
                                        \draw[scorebar] (232,\y) -- (\endpoint,\y);
                                }
                        \foreach \y/\endpoint in {169/279,155/281,141/279,127/281,
                                        113/279,99/287,85/295,71/295} {
                                        \draw[scorebar] (274,\y) -- (\endpoint,\y);
                                }
                        \foreach \y/\endpoint in {169/320,155/324,141/336,127/322,
                                        113/336,99/320,85/318,71/328} {
                                        \draw[scorebar] (316,\y) -- (\endpoint,\y);
                                }

                        \draw[betacut] (212,66) -- (212,174);
                        \draw[betacut] (253,66) -- (253,174);
                        \draw[betacut] (287,66) -- (287,174);
                        \draw[betacut] (328,66) -- (328,174);

                        \foreach \x/\y in {222/169,206/155,204/141,222/127,211/113,208/99,212/85,206/71,
                                        244/169,263/155,253/141,246/127,244/113,263/99,242/85,240/71,
                                        279/169,281/155,279/141,281/127,279/113,287/99,295/85,295/71,
                                        320/169,324/155,336/141,322/127,336/113,320/99,318/85,328/71} {
                                        \node[adjustedpoint] at (\x,\y) {};
                                }

                        \foreach \x/\y in {222/169,263/155,336/141,222/127,
                                        336/113,263/99,295/85,295/71} {
                                        \draw[qbchoice]
                                        ($(\x,\y)+(90:3.0pt)$) -- ($(\x,\y)+(126:1.35pt)$) --
                                        ($(\x,\y)+(162:3.0pt)$) -- ($(\x,\y)+(198:1.35pt)$) --
                                        ($(\x,\y)+(234:3.0pt)$) -- ($(\x,\y)+(270:1.35pt)$) --
                                        ($(\x,\y)+(306:3.0pt)$) -- ($(\x,\y)+(342:1.35pt)$) --
                                        ($(\x,\y)+(18:3.0pt)$) -- ($(\x,\y)+(54:1.35pt)$) -- cycle;
                                }
                \end{scope}

                \node[token] (newtoken1) at (428,169) {};
                \node[token, below=6.8pt of newtoken1] (newtoken2) {};
                \node[token, below=6.8pt of newtoken2] (newtoken3) {};
                \node[token, below=6.8pt of newtoken3] (newtoken4) {};
                \node[token, below=6.8pt of newtoken4] (newtoken5) {};
                \node[token, below=6.8pt of newtoken5] (newtoken6) {};
                \node[token, below=6.8pt of newtoken6] (newtoken7) {};
                \node[token, below=6.8pt of newtoken7] (newtoken8) {};
                \node[balancedexpert] (newexpert1) at ($(newtoken1)+(82pt,-4pt)$) {$E_1$};
                \node[balancedexpert, below=13pt of newexpert1] (newexpert2) {$E_2$};
                \node[balancedexpert, below=13pt of newexpert2] (newexpert3) {$E_3$};
                \node[balancedexpert, below=13pt of newexpert3] (newexpert4) {$E_4$};
                \foreach \i in {1,...,8} {
                                \node[small, text=qbfigink, anchor=east]
                                at ($(newtoken\i.west)+(-2pt,0pt)$) {$t_{\i}$};
                        }

                \draw[route] (newtoken1) -- (newexpert1);
                \draw[route] (newtoken4) -- (newexpert1);
                \draw[route] (newtoken2) -- (newexpert2);
                \draw[route] (newtoken6) -- (newexpert2);
                \draw[route] (newtoken8) -- (newexpert3);
                \draw[reroute] (newtoken3) -- (newexpert4);
                \draw[reroute] (newtoken5) -- (newexpert4);
                \draw[reroute] (newtoken7) -- (newexpert3);

                \coordinate (flowlevel) at ($(rawtoken4)!0.5!(rawtoken5)$);
                \coordinate (leftflowstart) at ($(rawexpert2.east |- flowlevel)+(19.5pt,0pt)$);
                \coordinate (rightflowstart) at ($(newtoken4.west)!0.5!(newtoken5.west)+(-55pt,0pt)$);
                \draw[flow] (leftflowstart) -- ++(24pt,0pt);
                \draw[flow] (rightflowstart) -- ++(24pt,0pt);
        \end{tikzpicture}
\end{adjustbox}
    \caption{Illustration of Quantile Balancing with $m=8$ tokens, $n=4$ routed experts, and $k=1$ selected expert per token.
        (a) Token-wise Top-$k$ routing (tokens on the left, experts on the right) produces loads $(4,3,1,0)$; darker circles indicate overheated experts, whereas faded and dashed circles indicate underutilized and dying experts, respectively.
        (b) Each gray bar is the margin of the currently biased score, $s_{i,j}+b_j^{(t)}-\alpha_i^{(t)}$, so the row-wise maxima reproduce the routing in (a).
        The dashed red line in each column is the bias adjustment $b_j^{(t)}-\widehat{b}_j^{(t+1)}$, placed at the $(q{+}1)$-th largest margin so that exactly $q=2$ margins exceed it.
        The marker \textcolor{brickred}{\ensuremath{\bigstar}} denotes the row-wise Top-$k$ choice after subtracting the column adjustments, i.e., the routing in (c).
        (c) The retained choices yield the balanced load $(2,2,2,2)$; red edges denote assignments changed by QB.}
    \label{fig:quantile-balancing}
\end{figure}

\subsubsection{Quantile Balancing}
\label{sec:qb}
Unlike auxiliary-loss-based routing~\citep{fedus2022switch}, \kimi{3} adopts auxiliary-loss-free routing~\citep{deepseekaiv3}.
Load balancing is implemented by adding an expert-specific bias $b_j$ to the router score used for Top-$k$ selection.
For token $\bm{x}_i$, the router computes $\bm{s}_i=\operatorname{Sigmoid}(\mathbf{W}_r\bm{x}_i)$ and applies
\begin{equation}
    \mathcal{T}_i
    = \operatorname{argtop}_{k}\!\left(\bm{s}_i+\bm{b}\right),
    \qquad
    p_{i,j}
    = \frac{s_{i,j}}{\sum_{r\in\mathcal{T}_i}s_{i,r}},
    \quad j\in\mathcal{T}_i.
    \label{eq:moe-routing}
\end{equation}
Because $\bm{b}$ is omitted from $p_{i,j}$, it regulates dispatch without altering the mixture weights or the gradient-based optimization of the router.
The original method updates $\bm{b}$ with the fixed-step rule $b_j^{(t+1)}=b_j^{(t)}+\gamma\operatorname{sign}(\bar{\ell}-\ell_j^{(t)})$~\citep{deepseekaiv3}, for which $\gamma$ trades off slow adaptation against load oscillation.
Maintaining balanced loads becomes more challenging as LatentMoE increases the routed expert pool to 896 per layer.
Imbalanced routing slows expert-parallel training and may leave some experts poorly trained~\citep{huang2026step35flash}.

To address this limitation, we introduce Quantile Balancing (QB), which sets each expert bias from the router-score quantile~\citep{su2026qb,sun2026expertthreshold} that matches its target load.
Consider a training batch of $m$ tokens routed to $n$ experts with Top-$k$ selection, so the target load is $q:=mk/n$ tokens per expert.
QB derives the next bias from a single forward pass.
Routing replaces the Top-$k$ selection with Top-$(k{+}1)$ on the biased score $\bm{s}_i+\bm{b}^{(t)}$: the first $k$ entries are the routes actually taken, while the $(k{+}1)$-th entry is the cutoff $\alpha_i^{(t)}$ that an expert must exceed to enter token $i$'s Top-$k$.
Taking the cutoff from Top-$(k{+}1)$ routing avoids a separate token-side quantile.
We then choose each expert bias so that expert $j$ receives its target load: with the cutoffs fixed, the token count routed to expert $j$ under a candidate bias $\widehat{b}_j^{(t+1)}$ is
\begin{equation*}
    \sum_{i=1}^{m}\mathbf{1}\!\left[s_{i,j}+\widehat{b}_j^{(t+1)}>\alpha_i^{(t)}\right],
\end{equation*}
which is monotonically decreasing in the threshold $-\widehat{b}_j^{(t+1)}$.
Assuming no ties, setting this count to $q$ makes $-\widehat{b}_j^{(t+1)}$ the $(q{+}1)$-th largest margin $s_{i,j}-\alpha_i^{(t)}$, so that exactly $q$ margins stay above the threshold.
Since $q/m=k/n$, this is the $(1-k/n)$-quantile of the margins across tokens, giving the QB update
\begin{align}
    \widehat{b}_j^{(t+1)}
     & \leftarrow -\operatorname{quantile}_{1-k/n}\!\left(\bm{s}_{:,j}-\bm{\alpha}^{(t)}\right), \nonumber \\
    \bm{b}^{(t+1)}
     & \leftarrow \widehat{\bm{b}}^{(t+1)}
    - \operatorname{mean}\!\left(\widehat{\bm{b}}^{(t+1)}\right)\mathbf{1}.
    \label{eq:qb-update}
\end{align}
The margins subtract the biased cutoff $\alpha_i^{(t)}$ from the raw score $s_{i,j}$, so the old bias enters the update only through the cutoffs, and the second line removes a common offset that leaves Top-$k$ selection unchanged.
For causality, the update takes effect only in the next step~\citep{deepseekaiv3}, i.e., a batch is never routed with a bias derived from itself.
Fig.~\ref{fig:quantile-balancing} illustrates the case $m=8$, $n=4$, and $k=1$, where each expert receives the target load $q=2$.
The final bias is frozen at inference.
The balanced-assignment derivation is given in \S~\ref{app:qb-derivation}.

\paragraph{Histogram estimation}
At scale, the quantile in Eq.~\ref{eq:qb-update} spans the full global batch, whose margins number in the millions and are spread across ranks and accumulation steps, so gathering them for an exact quantile is not viable at training time.
We instead read each expert's quantile from a histogram of its margins: a single all-reduce sums the per-rank bin counts, and the quantile is recovered from the pooled counts.
Because counts are additive, the histogram represents the pooled global batch regardless of how tokens are sharded, so the estimate reflects the whole-batch quantile up to the bin width, at a communication cost of only a few hundred bins per expert.
This histogram estimator is the method we use in practice; we give more detailed descriptions of it and its error bound in \S~\ref{app:qb-histogram}.

\subsection{Native Vision}
\label{sec:native-vision}

\kimi{3} is natively multimodal: text, images, and videos are processed by a single shared backbone within one context, with no post-hoc modality-alignment stage.
This design is the architectural foundation of the long-horizon, vision-in-the-loop behavior described in \S\ref{sec:intro}. Rendered outputs and the code that produced them live in the same token stream, the model can write code, inspect screenshots or video frames of the result, and iteratively refine visual artifacts---user interfaces, graphics, video---with no cross-model hand-off.

\begin{figure}[!th]
    \centering
    \includegraphics[width=0.9\linewidth]{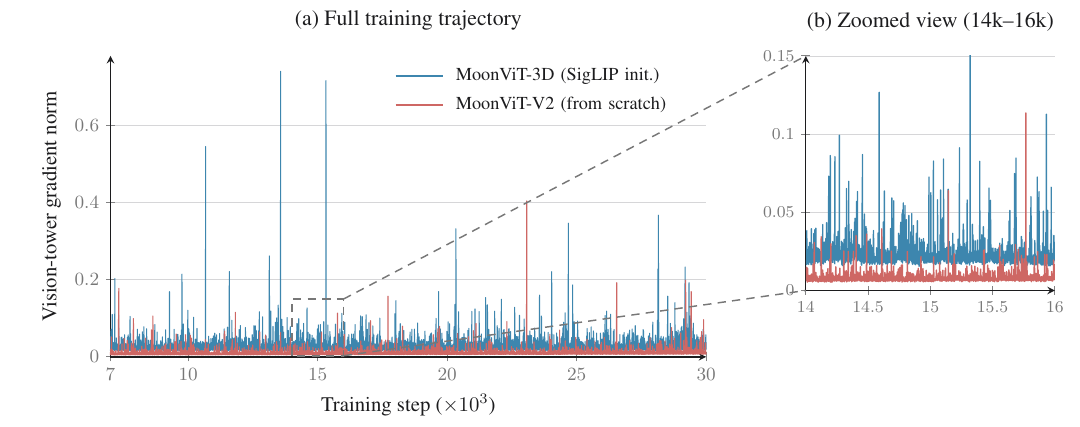}
    \caption{Vision-tower gradient norms in our pre-training ablations. Compared with the SigLIP-initialized MoonViT-3D, the from-scratch MoonViT-V2 maintains lower gradient norms with fewer spikes, indicating more stable optimization.}
    \label{fig:vitgradnorm}
\end{figure}

\paragraph{MoonViT-V2}
A key departure from \kimi{2.5} is that we train \kimi{3} vision encoder, \emph{MoonViT-V2, entirely from scratch with next-token prediction}.
Prior practice, including \kimi{2.5} itself, initializes the vision encoder from a contrastively pre-trained model such as SigLIP, under the premise that pre-trained visual knowledge gives the model a head start.
We depart from this practice primarily for training stability.
When a pre-trained encoder is attached to the LLM, joint optimization becomes unstable: the SigLIP-initialized MoonViT-3D shows persistently higher gradient norms with frequent spikes, while MoonViT-V2 remains stable throughout training (Fig.~\ref{fig:vitgradnorm}).
Training with next-token prediction also allows the encoder's representations to be shaped directly by the language-modeling objective, rather than by a contrastive loss that favors global semantics over fine-grained textual and structural cues.
Notably, we find MoonViT-V2 matches the SigLIP-initialized baseline across vision evaluations, indicating that contrastive pre-training is unnecessary as an initialization for multimodal language models at scale.

\paragraph{Architecture}
This training recipe builds on a vision pathway that follows the overall design of \kimi{2.5}~\citep{kimik25,team2025kimivl}: visual inputs are first encoded by MoonViT-V2 and then mapped by a lightweight MLP projector into the LLM.
MoonViT-V2 is a 27-layer vision transformer with roughly 0.4B parameters that adopts RMSNorm and removes all bias terms from its linear and attention projections, a design that further stabilizes the from-scratch optimization above.
Images and videos are processed with fully shared parameters, as in MoonViT-3D: attention is factorized into intra-frame spatial and inter-frame temporal passes, and temporal pooling further compresses tokens along the time dimension.
Before projection, a pixel-shuffle operation with $2\times2$ downsampling reduces the number of visual tokens by a factor of four, keeping inputs of up to $3584\times3584$ pixels affordable within the 1M-token context.

\subsection{Per-Head Muon}
\label{sec:perhead_muon}
Following \kimi{2}, \kimi{3} adopts Muon~\citep{jordan2024muon} as the optimizer for its matrix parameters. For attention projections, we further refine it into a per-head variant~\citep{su2025muon,glm5team2026glm5vibecodingagentic}: instead of applying Newton--Schulz orthogonalization to the full $Q$, $K$, and $V$ projection matrices, we partition their momentum matrices along the head dimension and orthogonalize each head's block separately.
The intuition is that full-matrix orthogonalization treats all heads as a single coupled block, so heads with larger gradient or momentum scales dominate the shared update direction, while smaller-scale heads receive insufficiently normalized updates; per-head orthogonalization equalizes the update scale across heads.
In practice, this design yields more balanced learning dynamics across heads and improves training stability at larger scales.
It also slightly reduces optimizer overhead, as Newton--Schulz iterations on tall per-head blocks are cheaper than on the full projection matrix.

\section{Pre-Training}

\subsection{Pre-Training Data}                        
\kimi{3} is pre-trained on a curated corpus spanning four primary text
domains---Web Text, Code, Mathematics, and Knowledge---together with a
large-scale vision corpus. The vision data covers captions, interleaved image--text
documents, OCR, perception, video, and visual coding data. Our
data pipelines build on those developed for \kimi{2}~\citep{kimiteam2025kimik2openagentic} and
refined in \kimi{2.5}~\citep{kimik25}.                                                                                                            
 \paragraph{Text data} Each domain is filtered by a combination of rule-based heuristics, classifier-based quality scoring, and deduplication, with domain-specific sampling rates determined by ablation studies on smaller models. Following the rephrasing recipe of \kimi{2}~\citep{kimiteam2025kimik2openagentic}, we rephrase knowledge and mathematics corpora with style and perspective-diverse prompting, chunk-wise autoregressive generation, and fidelity verification against the source documents.

 \paragraph{Vision data} The vision corpus follows the taxonomy of \kimi{2.5}~\citep{kimik25}, combining open-source collections with in-house pipelines for filtering, synthesis, and deduplication.
 During training, coordinate supervision is provided in both absolute and normalized ([0,1]) formats, enabling precise and resolution-robust localization.
 In addition to classical text-captioned images, we substantially scale up programmatic multimodal data, coupling code snippets with their rendered visuals across domain-specific formats including SVG, 3D assets, Webpage, Game, and CAD schematics.

\subsection{Scaling Law}
\begin{figure}[t]
    \centering
    \includegraphics[width=0.55\linewidth]{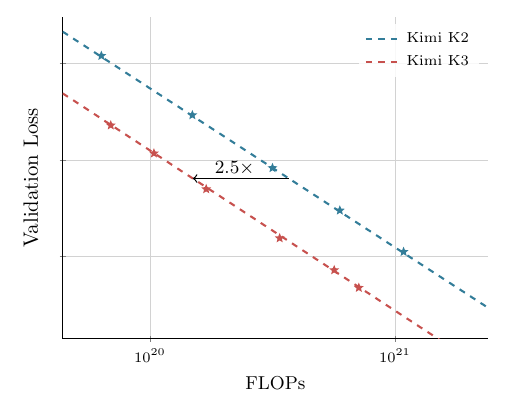}
    \caption{
        Fitted scaling-law curves for \kimi{2} and \kimi{3}.
        \kimi{3} achieves ~$2.5\times$ gain in scaling efficiency over \kimi{2}.
    }
    \label{fig:scaling-law}
\end{figure}
Taken together, the architectural, data, and training improvements described in the previous sections define our new model family. Since these changes also alter the optimal training regime, we conduct dedicated scaling-law studies to retune key hyperparameters, including the batch size, learning rate, tokens-per-parameter ratio (TPP) and the model shape. Evaluated on held-out OOD validation data, the scaling law curves in (Fig.~\ref{fig:scaling-law}) show that these improvements collectively deliver an approximately $2.5\times$ gain in overall scaling efficiency over \kimi{2}. Table~\ref{tab:k2-k3-comparison} provides a detailed architectural comparison between \kimi{2} and \kimi{3}, highlighting the structural changes that contribute to this improvement.

Our scaling-law study consistently favors cosine decay over Warmup Stable Decay (WSD) \citep{hu2024minicpm}, leading us to adopt cosine decay as the default learning rate schedule. We compare cosine decay and WSD under a fixed minimum learning rate. Although prior work has reported that WSD can match or even outperform cosine decay, we observe that the two schedules exhibit markedly different optimal hyperparameters. Even under the same model size and training-token budget, their optimal peak learning rates and batch sizes differ substantially. As a result, comparing the two schedules using a shared set of hyperparameters may unfairly favor one simply because those hyperparameters are better aligned with it. To ensure a fair comparison, we conduct an independent scaling-law search for each schedule. Under their respective optimal hyperparameter settings, cosine decay consistently achieves a lower final loss than WSD.

\begin{table}[t]
    \centering
    \caption{Architectural comparison between \kimi{2} and \kimi{3}.}
    \label{tab:k2-k3-comparison}
    \small
    \renewcommand{\arraystretch}{1.08}
    \setlength{\tabcolsep}{7pt}

    \begin{tabularx}{0.94\linewidth}{
        >{\raggedright\arraybackslash}X
        >{\centering\arraybackslash}p{0.16\linewidth}
        >{\centering\arraybackslash}p{0.22\linewidth}
        >{\centering\arraybackslash}p{0.10\linewidth}
    }
        \toprule
        & \textbf{\kimi{2}}
        & \textbf{\kimi{3}}
        & $\boldsymbol{\Delta}$ \\
        \midrule
        Architecture
            & MoE
            & MoE
            & -- \\
        $\#$Layers
            & 61
            & 93
            & $\uparrow 52\%$ \\
        Total Parameters
            & 1.04T
            & 2.78T
            & $\uparrow 167\%$ \\
        Activated Parameters
            & 32.6B
            & 104.2B
            & $\uparrow 220\%$ \\
        Hidden Dimension
            & 7,168
            & 7,168
            & $=$ \\
        Latent MoE Dimension
            & --
            & 3584 (0.5×)
            & -- \\
        MoE Hidden Dimension per Expert
            & 2,048
            & 3,072
            & $\uparrow 50\%$ \\
        Routed Experts
            & 384
            & 896
            & $\uparrow 133\%$ \\
        Experts Active per Token
            & 8
            & 16
            & $\uparrow 100\%$ \\
        Shared Experts
            & 1
            & 2
            & $\uparrow 100\%$ \\
        Attention Heads
            & 64
            & 96
            & $\uparrow 50\%$ \\
        Number of Dense Layers
            & 1
            & 1
            & $=$ \\
        Vocabulary Size
            & 160K
            & 160K
            & $=$ \\
        Training Context Length
            & 128K
            & 1M
            & $8\times$ \\
        Attention Mechanism
            & MLA
            & Hybrid KDA--MLA
            & -- \\
        Activation Function
            & SwiGLU
            & SiTU-GLU
            & -- \\
        Attention-Layer Composition
            & 61 MLA
            & 69 KDA + 24 MLA
            & -- \\
        Number of MTP Layers
            & 1 layer
            & 1 layer
            & $=$ \\
        Total Parameters of ViT 
            & - 
            & 401M
            & - \\
        $\#$ViT Layers
            & - 
            & 27 layers 
            & - \\
        Patch Size of ViT
            & - 
            & 14
            & - \\
        $\#$Attention Heads of ViT
            & - 
            & 12 
            & - \\
        \bottomrule
    \end{tabularx}
\end{table}

\subsection{Training Recipe}

\kimi{3} adopts a native multimodal training strategy in which language and vision are jointly optimized from the start of training, rather than grafting a vision encoder onto a pre-trained language model through a post-hoc alignment stage. Under this paradigm, visual and textual tokens are interleaved within a single next-token prediction objective, enabling the shared backbone to learn unified multimodal representations from the outset.

We optimize the model using the Per-Head Muon optimizer (\S~\ref{sec:perhead_muon}) together with the weight-clipping mechanism introduced in \kimi{2}, while adopting QB (\S~\ref{sec:qb}) for MoE load balancing. We use a cosine learning rate schedule with a $1\%$ linear warmup. Weight decay is set to $0.1$ throughout.

Our pre-training begins with a context length of $8\mathrm{k}$ tokens, which is later extended to $64\mathrm{k}$ tokens in a subsequent training phase.

\subsection{Long-Context Extension}

\paragraph{Positional encoding}
\kimi{3} uses no explicit positional embedding (NoPE), and instead encodes positional information implicitly through the recurrent gating and decay mechanism of KDA.
As a result, the model extrapolates directly to 1M-token contexts without any positional-encoding modification, such as RoPE rescaling or interpolation~\citep{peng2023yarn}.

\paragraph{Long-context data}
Long documents and videos from natural sources contain a substantial amount of low-quality content, including near-duplicates, binary blobs, truncated files, video clips, and invalid machine-generated logs.
We therefore process them through a dedicated cleaning pipeline that combines exact and fuzzy deduplication, supplemented by perceptual hashing over frames for video, together with heuristic and classifier-based quality filtering, and structural validation.
Because genuinely long and coherent documents and videos are scarce relative to short text, we upsample them so that the long-context distribution is not overwhelmed by short sequences during cooldown. 
Length alone, however, does not confer long-range capability.
To address this, we synthesize additional long-context data by carefully permuting and concatenating multimodal documents and sub-tasks, so that the embedded tasks can be solved only by attending to information scattered across the full 1M-token context.
This trains the attention mechanism at the intended scale and prevents it from degenerating into local patterns.

\paragraph{Progressive context extension}
\kimi{3} supports a context window of up to 1 million tokens.
We achieve this through extending the context window progressively as training proceeds, following a four-stage curriculum.
The window grows from 8K to 64K tokens during pre-training, and from 256K to 1M tokens during the cooldown phase.
Concentrating the costly long-sequence computation within a small fraction of the overall training budget keeps the curriculum economical while still allowing the model to adapt gradually to increasingly long-range dependencies.
The sequence-dimension partitioning that makes million-token training tractable for the KDA layers is described in \S\ref{sec:kda-cp}.

\section{Post-Training}
\label{sec:post-training}

\subsection{Method}

Our post-training pipeline follows a three-stage paradigm:
initializing baseline agent capabilities via supervised fine-tuning (SFT),
developing specialized domain experts at varying reasoning effort via Reinforcement Learning (RL),
and consolidating these domain-specific policies into a single model using Multi-Teacher On-Policy Distillation (MOPD).

\subsubsection{Supervised Fine-Tuning}
\label{sec:post-sft}
The SFT stage establishes a high-quality cold-start policy for the subsequent RL stage.
Building on the SFT pipeline of previous Kimi models \cite{kimiteam2025kimik2openagentic,kimik25}, we expand the SFT dataset for \kimi{3}, substantially broadening its coverage of complex agentic tasks.
Specifically, we synthesize data trajectories using domain-specialized models from the prior Kimi series, followed by multi-stage verification and human-in-the-loop annotation.
To represent these complex agentic trajectories consistently, we serialize all data with our XTML-based chat template (eXtensible Token Markup Language; see \S~\ref{sec:post-chat-template} for details).
Collectively, these steps yield a large-scale instruction dataset that endows \kimi{3} with adaptive reasoning, precise tool calling, and robust execution in long-horizon agentic scenarios.
In addition, we apply quantization-aware training (QAT) from the SFT stage onward, with MXFP4 weights and MXFP8 activations (\S~\ref{sec:mxfp4-qat}).

\subsubsection{Reinforcement Learning}
\label{sec:post-rl}

\begin{figure}[t]
    \centering
    \includegraphics[width=0.8\textwidth]{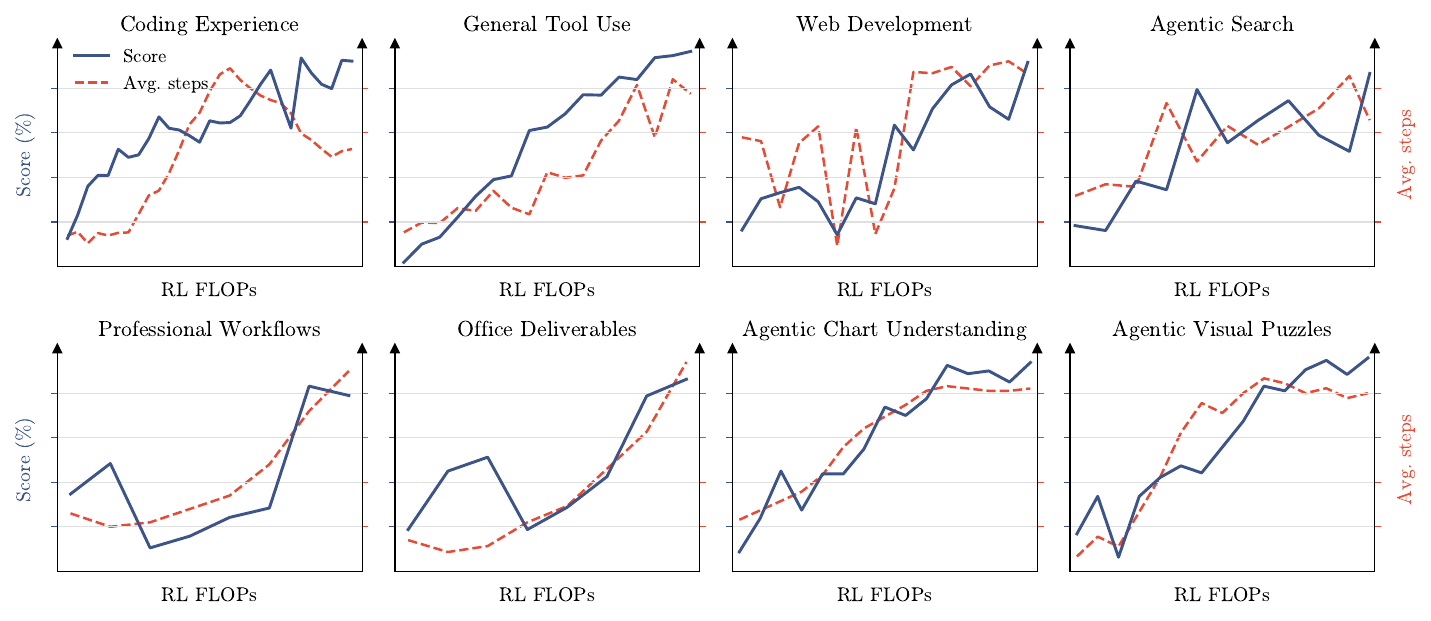}
    \caption{Scores and the average assistant steps across a variety of public and in-house evaluations during RL. By scaling RL FLOPs, tool-call steps scale up consistently, accompanied by a comprehensive improvement in the model's overall capability.}
    \label{fig:step_scaling_merged_curve}
\end{figure}

While SFT provides a solid cold-start foundation, RL is critical to unlocking higher-order reasoning and execution capabilities. 
Rather than training specialized RL models for individual tasks, we scale RL across three broad domains, each encompassing a wide spectrum of sub-tasks, and train a single expert for each domain at every reasoning effort level:
(i) \emph{general tasks}, spanning general experience, vision, reasoning, faithfulness, search capabilities, and knowledge work tasks; 
(ii) \emph{general agents}, spanning long-horizon assistant tasks, deep research, and paragraph-level writing; and 
(iii) \emph{coding agents}, spanning software engineering (SWE), coding experience, kernel tasks, and web development. 
As shown in Figure~\ref{fig:step_scaling_merged_curve}, 
scaling RL FLOPs consistently improves a variety of capabilities across knowledge, reasoning, vision, general agent, and coding. 
Crossing these three domain experts with three reasoning effort levels in $\{\text{low}, \text{high}, \text{max}\}$ yields a total of nine expert models.

\paragraph{Algorithm}

To mitigate the long-tail latency that intensifies in long-horizon tasks, we extend the \emph{partial rollout} scheme from our synchronous RL framework \citep{kimik15,kimik25}. 
During the rollout phase of each iteration, we sample $K$ completions for each of $N$ prompts, maintaining an active workload of $N \times K$ trajectories. Rather than waiting for all rollouts to terminate, the generation phase pauses as soon as a fraction $\lambda \in (0, 1)$ of trajectories completes (i.e., $\lambda NK$), allowing policy optimization to proceed without execution stragglers. Paused rollouts are enqueued and prioritized for resumption at the start of the next iteration, powered by our sandbox infrastructure (\S~\ref{sec:sandbox}). 
Once all $K$ responses for a prompt complete, they are immediately dispatched for policy optimization, which follows the algorithm in \kimi{2.5} \citep{kimik25}. 
Under our partial rollout scheme, an individual long-horizon trajectory naturally spans multiple iterations, introducing data staleness that threatens training stability. 
Our policy optimization algorithm inherently tolerates such an extreme off-policy regime through a per-token regularization.  By constraining policy updates within a localized neighborhood, this regularization enables the algorithm to robustly handle highly stale data and sustains training stability.

\paragraph{Reasoning Effort RL}

To fine-tune reasoning effort while maximizing token efficiency, we implement a per-problem budget control mechanism during RL \citep{kimik25}. 
We associate each problem $x$ with an initial token budget $b_0(x)$ estimated from the cold-start model, and override the task reward with $-1$ for trajectories whose total token budget $T(y)$ exceeds a scaled threshold $\tau \cdot b_0(x)$. 
For general tasks, $T(y)$ measures the number of thinking tokens, whereas for agentic tasks, $T(y)$ accounts for the cumulative output tokens, including both reasoning traces and tool-call arguments. 
Training follows a stage-wise curriculum over the budget multiplier $\tau$. We first train a \emph{max-budget} variant with a relatively large $\tau$, while still capping the maximum budget to suppress excessive overthinking. We then anneal $\tau$ to smaller values to obtain the \emph{high}- and \emph{low}-effort expert models. 
The adjustment of $\tau$ is configured per domain under human-in-the-loop guidance. Trajectories produced by the resulting experts at all reasoning levels are jointly collected for supervised fine-tuning and multi-teacher on-policy distillation.

\paragraph{Agentic Generative Reward Model}
For non-verifiable general tasks, we adopt an Agentic Generative Reward Model (GRM), retaining the tournament-style group reward with binary comparisons as in \kimi{2.5}~\cite{kimiteam2025kimik2openagentic,kimik25}. Beyond generic agentic capabilities for enhanced judgment, the agentic judge is required to follow a mandatory protocol: (1) read the outcome, product, or text output; (2) generate a rubric; (3) score each candidate against the rubric; and (4) record the rubric-assigned scores in a scorepad. To mitigate reward hacking toward increasingly verbose outputs, we apply a budget-based verbosity control analogous to the reasoning-effort control above: given an initial verbosity $\ell_0$ estimated from the cold-start model and a multiplier $\sigma$, a candidate whose output length exceeds $\sigma \cdot \ell_0$ automatically loses the binary comparison.

\subsubsection{Multi-Teacher On-Policy Distillation}

We adopt Multi-Teacher On-Policy Distillation (MOPD) to consolidate these domain-specialized capabilities across varying reasoning efforts into a unified model \citep{lu2025onpolicydistillation, xiao2026mimov2flash, deepseekv4}. 
During training, for a given domain $d$ and a sampled reasoning effort level $e \in \{\text{low}, \text{high}, \text{max}\}$, optimization is guided by the corresponding teacher model $\pi_{\text{teacher}}^{(d,e)}$ among the nine experts. 
Given an input query $x$ and the prefix response $y_{<t}$, the per-token OPD reward evaluated on $y_t$ between the teacher $\pi_{\text{teacher}}^{(d,e)}$ and the student $\pi_{\theta}$ is defined as:
\begin{align}
r^{d}_{\mathrm{opd}}(y_t \mid e, x, y_{<t}) = \mathrm{clip}\left( \mathrm{sg} \left(\log \frac{\pi_{\text{teacher}}^{(d,e)}(y_t \mid x, y_{<t})}{\pi_{\theta}(y_t \mid e, x, y_{<t})} \right) , -R_{\max}, R_{\max} \right)\, ,
\end{align}
where $\operatorname{sg}(\cdot)$ denotes the stop-gradient operator, and $R_{\max} > 0$ is a clipping threshold to constrain extreme advantage signals, thereby stabilizing RL training. 
This dense reward signal seamlessly integrates into our RL framework, naturally enabling infrastructure-level optimizations such as partial rollout training for long-horizon tasks. 
While we also experimented with more fine-grained top-$k$ distillation objectives, we observed no clear advantage in either convergence speed or final performance in our setting. 

\subsubsection{Deployment-Aware Post-Training}

\paragraph{MXFP4 Quantization-Aware Post-Training}
\label{sec:mxfp4-qat}

To reduce memory footprint and serving cost at deployment, we quantize the MoE expert weights --- which dominate the model's parameter memory --- to MXFP4~\citep{rouhani2023microscaling}, with activations computed in MXFP8, while all non-expert components (attention projections, latent MoE projections, shared experts, and MoE routers) remain in higher precision.
We perform quantization-aware training (QAT)~\citep{jacob2018quantization} throughout the entire post-training stage, covering both SFT and RL, so that the model adapts to quantization-induced precision loss.
During RL, rollout and training share the same quantization scheme --- eliminating the train--inference mismatch.

\paragraph{Draft Model Fine-Tuning}
\label{sec:post-eagle3}
Optimizing inference efficiency is crucial for serving complex, long-horizon agentic models.
\kimi{3} is pre-trained with a multi-token-prediction (MTP) layer that mirrors the structure of a backbone block.
As the draft model of EAGLE-3~\citep{li2025eagle3} comprises a single decoder layer whose structure matches the MTP layer, we fine-tune the pre-trained MTP layer into an EAGLE-3-style draft model, with the target model frozen and only the draft layer and its feature-fusion projection updated.
Following the training-time test protocol of EAGLE-3, the draft is unrolled for seven steps during training; beyond the first step, where the target-side features of the newest position are unavailable, the draft consumes its own outputs from earlier steps, mirroring the recurrent drafting procedure at inference.

The draft input fuses low-, mid-, and high-level features of the target model, taken from the outputs of the 1st, 4th, and final AttnRes blocks, respectively (\S~\ref{sec:attnres}).
These features are concatenated and projected to the hidden size by a bias-free matrix $\bm{W}_{\mathrm{E3}}$, initialized as $[\,\bm{0}\;\;\bm{0}\;\;\bm{I}\,]$ so that the fused representation coincides at initialization with the high-level feature $\bm{h}_{h}$ --- the input on which the MTP layer was pre-trained --- and gradually learns to incorporate the low- and mid-level features during fine-tuning.

The speedup of speculative decoding is governed by the per-token acceptance rate $\sum_{x \in \mathcal{V}} \min\!\left(p(x), q(x)\right)$ under lossless speculative sampling, where $p$ and $q$ denote the next-token distributions of the target and draft models.
Since minimizing the conventional KL-divergence surrogate does not guarantee maximizing this rate for a capacity-limited draft model, we directly optimize the likelihood-based LK loss~\citep{samarin2026lklosses}, the negative logarithm of the acceptance rate itself,
\begin{equation}
    \mathcal{L}_{\mathrm{LK}}
    = -\log \sum_{x \in \mathcal{V}}
    \min\!\left(p(x), q(x)\right),
\end{equation}
with $p$ and $q$ evaluated at temperature~1 and no auxiliary ground-truth cross-entropy term.
Draft fine-tuning follows the post-training QAT configuration (\S~\ref{sec:mxfp4-qat}), with MoE expert weights in MXFP4 and their input activations in MXFP8, while non-expert modules remain in higher precision.

\subsection{RL Task Synthesis and Agentic Environments}
\label{sec:post-envs}

The effectiveness of our RL framework relies heavily on rich, diverse, and robustly verifiable environments. To support scalable training across complex long-horizon tasks, we design a series of specialized white-box environments and task synthesis paradigms.

\subsubsection{Unified White-Box RL Environment}

Training with a single fixed agent harness can cause a model to overfit to a particular tool schema, system prompt, context management mechanism, or interaction protocol. To address this, we develop a unified white-box RL environment that represents an agent harness as a collection of configurable, composable modules, including tool interfaces, system prompts, context management strategies, skills, memories, subagents, and other components. Composing these modules through configuration, the environment can instantiate mainstream harnesses such as Kimi Code~\citep{kimicode}, Claude Code~\citep{claudecode}, Codex~\citep{openaicodex}, OpenClaw~\citep{openclaw}, and Hermes~\citep{hermesagent}, as well as entirely new ones. During RL training, we dynamically construct different harness configurations for different task groups,  exposing \kimi{3} to diverse combinations of these modules rather than the conventions of any single harness. The same abstraction also readily supports RL across various task domains, providing a scalable foundation for training more general-purpose agents.

\subsubsection{Knowledge-Graph-Guided Task Synthesis}
\label{sec:post-data}

\begin{figure}[htb]
    \centering
    \includegraphics[width=0.9\textwidth]{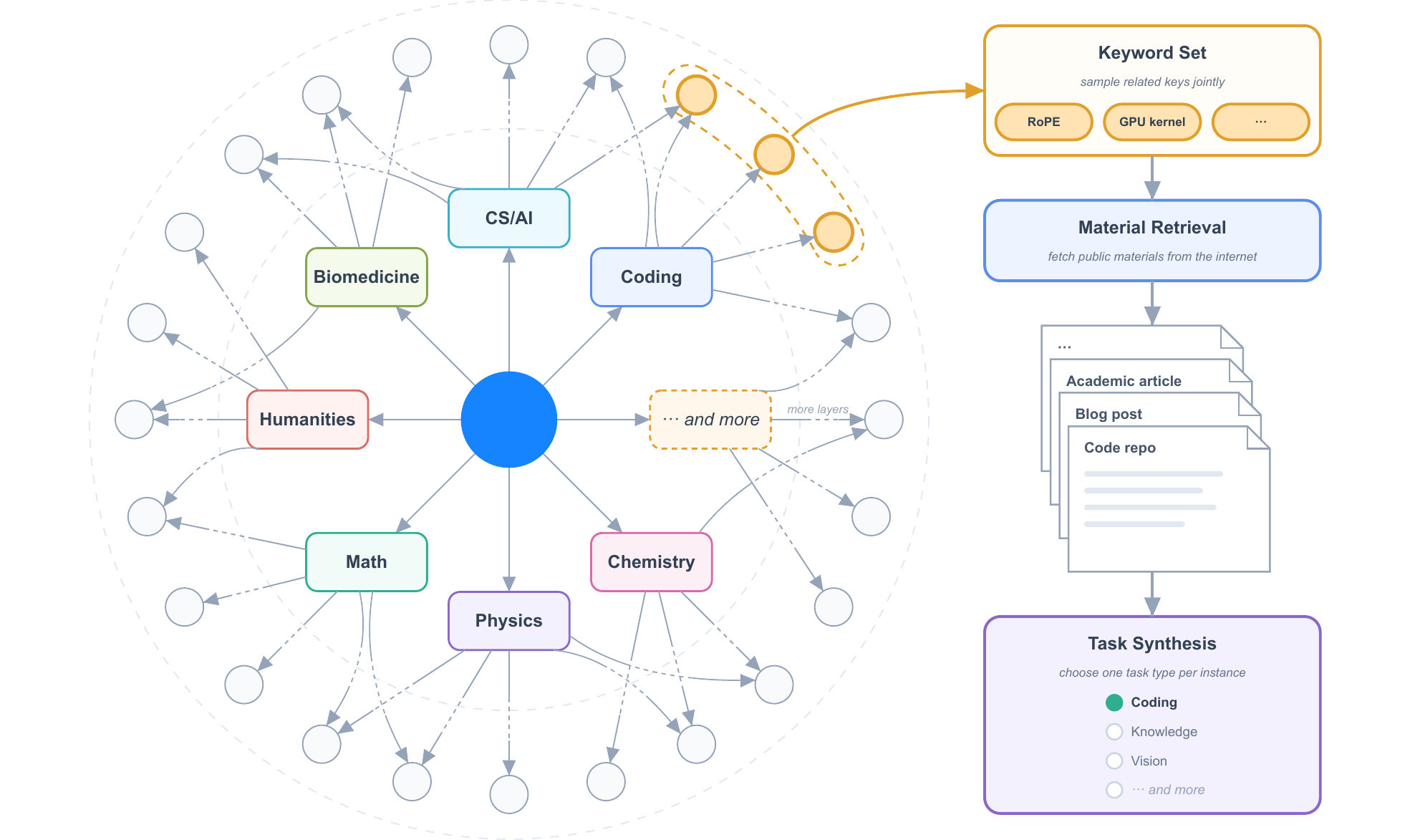}
    \caption{Overview of knowledge-graph-guided task synthesis. The hierarchically organized knowledge graph represents concepts at multiple levels, ranging from broad domains to fine-grained concepts. Related nodes are sampled to form a keyword set that guides the retrieval of publicly available source materials. For each synthesis instance, the system selects a task type and uses the retrieved materials to synthesize a corresponding task.}
    \label{fig:post-data-pipeline}
\end{figure}

\paragraph{Motivation and overview}
The quality and diversity of post-training tasks are largely determined by their source materials. Retrieval guided by fine-grained concepts surfaces specialized and underrepresented knowledge, while sampling across diverse concepts broadens domain coverage. To control both granularity and coverage at scale, we build a self-evolving, hierarchically organized knowledge graph that agents continuously expand through web-scale exploration across knowledge-intensive and coding domains. Figure~\ref{fig:post-data-pipeline} illustrates the task synthesis pipeline.

\paragraph{Agentic knowledge graph construction}
We construct the knowledge graph as a directed acyclic graph through recursive, agent-driven expansion. The expansion process begins with a predefined set of coarse-grained seed nodes. An agent instance is then assigned to each node and performs multiple web searches to investigate the corresponding concept. Before adding new nodes, the agent explores the existing graph to identify equivalent or related concepts, reuse existing nodes where appropriate, and minimize duplication. Edges are always directed from the coarser concept to the finer one, regardless of which endpoint the agent discovers first. Newly added nodes are subsequently assigned to agents for further exploration. A branch stops expanding when the assigned agent determines that the current concept is sufficiently atomic.

\paragraph{Material retrieval and task synthesis}
To target a desired distribution across domains and task types, the system samples nodes at varying levels of granularity, either individually or in related combinations. Keywords derived from the sampled nodes are combined with contextual information from their ancestors in the knowledge graph to formulate web queries. The retrieved real-world materials are assembled so that a synthesis agent produces training tasks of various task types.

\subsubsection{Verifiable Problems in Agentic Environments}
\label{sec:agentic-verifiable-problems}

We train \kimi{3} on verifiable problems in agentic environments; representative examples include multi-step complex information searching, where the model plans its research, gathers evidence from the web step by step, and produces a verifiable answer; the real day-to-day work of professionals, such as investment banking, data analysis, and legal practice, where the model decomposes a complex request, operates domain tools in a sandbox, and completes a deliverable over dozens to hundreds of steps; and multi-step verifiable visual reasoning over STEM problems, visual puzzles, and chart understanding. Each visual-reasoning trajectory is generated in an agent environment equipped with a Python interpreter in an isolated sandbox: the model iteratively writes and executes code to crop, zoom, or transform the input image, perform precise computation, or verify intermediate results, and receives the execution outputs --- including generated images --- as new observations over multiple interaction steps. As the model learns to perform more image operations and collect more observations, its performance on complex visual reasoning tasks steadily improves.

\subsubsection{Kernel Optimization Tasks}
\label{sec:post-envs-kernel}

To strengthen \kimi{3}'s GPU kernel optimization capabilities, we build a large-scale suite of kernel tasks ranging from single-operator kernels to fused mega-kernels, sourced from high-quality GitHub repositories such as Flash Linear Attention~\cite{yang2024fla}. The suite spans diverse GPU programming approaches, such as CUDA, Triton, CuTe DSL, Gluon, ThunderKittens~\cite{spector2025thunderkittens}, and TileLang~\cite{wang2025tilelang}, and covers widely used GPU architectures and numerical formats including BF16, FP8, and FP4. Rewards evaluate both correctness and performance: each kernel provides a PyTorch reference implementation, and solutions exceeding a predefined numerical error threshold receive zero reward. Performance is scored against an expert implementation, where matching it yields a reward of 0.5 and approaching the hardware roofline increases the reward toward 1. To ensure that rewards reflect genuine optimization, we develop a hacking-detection system that penalizes reward-hacking strategies such as CUDA graph replay, input caching, and precision reduction, and we continuously extend it with new safeguards as new hacking strategies are observed during \kimi{3}'s development.

\subsubsection{Personal Assistant Tasks}
\label{sec:week-task}
For long-horizon personal assistant tasks, we develop realistic mock implementations of widely used applications, such as Gmail, Notion, Slack, and Canvas. They preserve the core semantics of their real-world counterparts while enabling reproducible, large-scale interaction without external APIs or rate limits. Building on these mock applications, we design complex tasks inspired by real-world professional workflows in scenarios like human resources, legal services, and finance. In each task, the agent operates in a persistent, evolving environment over multiple simulated days and encounters dozens of interdependent events distributed across applications. A single rollout may involve up to thousands of tool calls and millions of context tokens. Each event carries its own evaluation criterion, assessed by deterministic rules or LLM-based evaluators. The initial workspace is constructed by agents that autonomously search the web for reference materials and transform them into a coherent, task-relevant environment. We also extend our RL framework to support such living environments, modeling complex event streams and the induced world-state transitions.

\subsubsection{Autonomous Execution Tasks}
\label{sec:post-envs-autonomous}

\begin{figure}[t]
    \centering
    \includegraphics[width=0.65\textwidth]{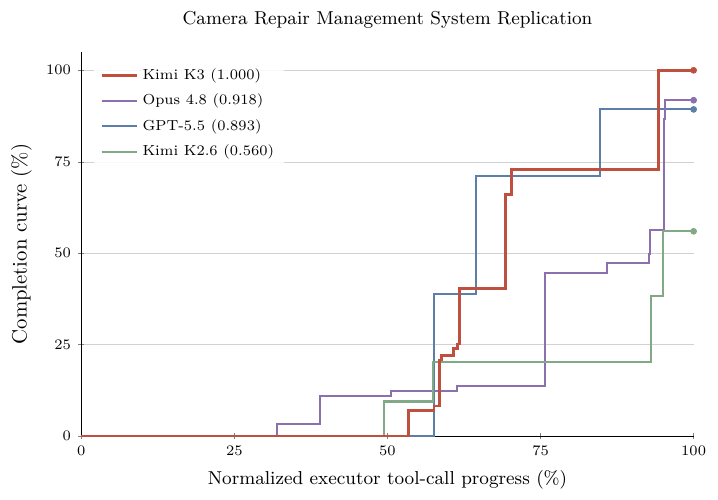}
    \caption{Completion curves on Camera Repair Management System, a black-box system replication task in which the agent reconstructs a hidden 3D-camera repair system as a web application through oracle queries. Completion denotes verifier-assessed task progress.}
    \label{fig:aet_blackbox_curve}
\end{figure}

We introduce Autonomous Execution Tasks (AET), an environment paradigm that trains long-horizon agent intelligence through verify-in-the-loop optimization. Each task specifies an initial state, a constrained goal, a tool-based action space, execution budgets, and an independent verifier. Agents see only the objective, context, constraints, and verification interfaces, without reference trajectories or predefined procedures, and must autonomously perform task decomposition, tool selection, planning, error recovery, and termination. Rewards are grounded in the verifier's evaluation of the final environment state rather than the agent's self-reported completion. We design multiple types of verifiers that support diverse environments, including black-box system replication (Figure~\ref{fig:aet_blackbox_curve}), quantitative factor discovery, and tax auditing. In each environment, agents iteratively submit solutions, receive verifier feedback, and refine their strategies, training a general loop of hypothesizing, acting, analyzing feedback, and adapting. Reward hacking is mitigated by isolating agents from verifiers, pairing public verifiers that offer diagnostic feedback with hidden verifiers that evaluate held-out scenarios, and applying penalty-based rewards under limited submission budgets.

\subsubsection{Web Development Tasks}
\label{sec:webdev-tasks-autonomous}

We construct a diverse suite of expert-curated web development tasks covering typical scenarios. Inputs range from one-line scene descriptions to multi-paragraph specifications; artifacts span websites, interactive games, 3D/WebGL scenes, data visualization, SVGs, and full-stack applications. Every task runs in a containerized sandbox and is rolled out under diverse agent scaffolds rather than a single fixed harness, to promote cross-scaffold generalization. Rewards consist of two components: deterministic checks and model judging by an internal reward model. Deterministic checks functionally test application behavior, and score structural and pixel-level similarity for tasks that replicate a reference. The reward is zeroed when a project fails to build, runs with errors, or fakes rather than implements the artifact. Model judging uses other models to perform source code inspection or to look at and interact with the output artifact.

\section{Infrastructure}

\kimi{3} combines three system challenges rarely encountered in a single model: hybrid KDA attention, 3T-class sparse multimodal training and inference, and million-token agentic workloads. 
Our infrastructure is co-designed with these challenges across the model lifecycle. 
At the architecture level, high-performance KDA kernels and Context Parallelism make the recurrent formulation efficient within and across devices, in both training and inference.
During pretraining, balanced expert execution, reduced memory footprint, and communication-overlapped scheduling sustain high utilization at scale.
During 1M-token agentic RL, hierarchical state management and resumable sandbox execution preserve long trajectories across iterations. 
Finally, state-aware KDA prefix caching, specialized inference kernels, and cache- and budget-aware scheduling translate these efficiencies into predictable production serving.

\subsection{Algorithm-System Co-Design for KDA}
\label{sec:kda-codesign}

KDA replaces the growing key--value cache of softmax attention with a fixed-size recurrent state $\mathbf{S}\in\mathbb{R}^{d_k\times d_v}$ (\S\ref{sec:kda}), whose serial update poses challenges in parallel execution, in exchange for a fixed-size state that is cheap to transfer and reuse.
The designs below address the first property and exploit the second at two levels of execution, with fused kernels within a device and KDA Context Parallelism across devices.

\subsubsection{KDA Kernels across Regimes}
The serial dependence of the KDA state is at odds with the GPU's preference for wide, uniform parallelism, and it manifests as a different bottleneck in each execution regime.
We design a dedicated kernel for each regime.

\paragraph{Chunkwise kernel for training and prefill}
The chunkwise form of KDA is parallel within each chunk but serial across chunks, since the recurrent state must propagate from chunk to chunk.
Executed naively, these two phases alternate, leaving the SMs idle during the serial propagation.
We therefore develop FlashKDA~\citep{flashkda2026}, a CUTLASS-based chunkwise kernel that overlaps intra-chunk computation with cross-chunk state propagation.
The kernel decomposes the work into token-parallel stages and a head-parallel recurrence, each scheduled and tuned independently, and substantially outperforms the Triton reference implementation.
FlashKDA serves both training and inference prefill and is auto-dispatched as a backend of flash-linear-attention~\citep{yang2024fla}.

\paragraph{Intra-device context parallelism for long-context prefill}
Tensor parallelism partitions heads across devices but never shortens the recurrence, so under pure TP deployment, prefilling an ultra-long sequence leaves most SMs idle when each rank holds only a few heads.
The key observation is that the state transition of each segment can be evaluated independently of the incoming state and composed exactly afterward.
An automatic SM-level context-parallel (CP) planner~\cite{yywang2025deltanetcp,yang2024fla} therefore partitions the sequence across the SMs of a single rank, evaluates the segment transitions in parallel, and merges them to recover each segment's exact initial state.
In contrast to the cross-device KCP of \S\ref{sec:kda-cp}, this parallelism is entirely intra-device and incurs no cross-device communication.

KDA decoding presents challenges distinct from those encountered during training and prefill. We discuss these challenges in detail in \S\ref{sec:kda-decoding}.

\subsubsection{KDA Context Parallelism}
\label{sec:kda-cp}

The communication overhead of context parallelism differs fundamentally between softmax and linear attention.
Softmax attention requires ranks to exchange key--value blocks whose size grows with the sequence length~\citep{liu2023ring}.
Linear attention instead carries the preceding context in a fixed-size recurrent state $\mathbf{S}\in\mathbb{R}^{d_k\times d_v}$.
Prior context-parallel methods exploit the additive recurrence of vanilla linear attention by computing, on each rank, the state that the local tokens generate from $\mathbf{S}=\mathbf{0}$ and summing these local states over the preceding ranks to recover the incoming state~\citep{sun2024lasp,sun2025lasp2}.

This direct summation, however, is insufficient for KDA.
Recall from Eq.~\ref{eq:recurrent_KDA} that KDA updates its state as $\mathbf{S}_t=\mathbf{M}_t\mathbf{S}_{t-1}+\beta_t\bm{k}_t\bm{v}_t^{\top}$, where $\mathbf{M}_t:=\left(\mathbf{I}-\beta_t\bm{k}_t\bm{k}_t^{\top}\right)\operatorname{Diag}(\bm{\alpha}_t)$.
KDA's delta rule applies the token-dependent matrix $\mathbf{M}_t$ to the incoming state before adding the current write.
Consequently, the effect of a local sequence segment depends on the state entering that segment and cannot be determined from the state computed with $\mathbf{S}=\mathbf{0}$ alone.

To preserve this dependence, we introduce KDA Context Parallelism (KCP), which decomposes the effect of each segment into two locally computable quantities, a cumulative transition acting on the incoming state and a state generated locally from zero.
Following the chunkwise notation of \S\ref{sec:kda}, we write $\mathbf{S}_{[i]}^{t}$ for the recurrent state within the segment of rank $i$ after $t$ local tokens, so that $\mathbf{S}_{[i]}^{T_i}$ denotes the state leaving rank $i$ and entering rank $i+1$.
We write $\widetilde{\mathbf{S}}_{[i]}^{t}$ for the state of the same recurrence started instead from $\mathbf{S}=\mathbf{0}$.
For an arbitrary state entering the $(i+1)$-th of $P$ context-parallel ranks, the state after $t$ local tokens is
\begin{equation}
    \begin{aligned}
        \mathbf{M}_{[i+1]}^{t \leftarrow 1}
        := \prod_{r \leftarrow 1}^{t}\mathbf{M}_r \in \mathbb{R}^{d_k\times d_k},
        \qquad   \mathbf{S}_{[i+1]}^{t}
         & =\widetilde{\mathbf{S}}_{[i+1]}^{t} + \mathbf{M}_{[i+1]}^{t \leftarrow 1}\mathbf{S}_{[i]}^{T_i}                                                                                                                                    \\
         & = \widetilde{\mathbf{S}}_{[i+1]}^{t} + \mathbf{M}_{[i+1]}^{t \leftarrow 1}\sum_{j=1}^{i}\Big(\prod_{l \leftarrow j+1}^{i}\mathbf{M}_{[l]}^{T_l \leftarrow 1}\Big)\widetilde{\mathbf{S}}_{[j]}^{T_j}\in \mathbb{R}^{d_k\times d_v}.
    \end{aligned}
    \label{eq:kcp-compose}
\end{equation}
\begin{center}
    \begin{adjustbox}{width=\linewidth}
        \definecolor{kcpfigink}{HTML}{1C1C1E}
\definecolor{kcpfiggray}{HTML}{6E6E73}
\definecolor{midnightblue}{HTML}{005C7F}
\definecolor{brickred}{HTML}{B92622}
\definecolor{kvcolor}{RGB}{241,140,74}
\definecolor{kcpgrayd}{HTML}{8A9199}

\tikzset{
        kcpcell/.style={draw=kcpfigink, line width=0.3pt, inner sep=0pt, outer sep=0pt},
        grid4/.style={kcpcell, minimum width=32pt, minimum height=32pt,
                        path picture={
                                        \foreach \r/\a/\b/\c/\d in {0/20/60/10/40,1/70/30/50/20,2/40/80/20/60,3/10/50/30/70} {
                                                        \fill[#1!\a] ([shift={(0,24-8*\r)}]path picture bounding box.south west) rectangle ([shift={(8,32-8*\r)}]path picture bounding box.south west);
                                                        \fill[#1!\b] ([shift={(8,24-8*\r)}]path picture bounding box.south west) rectangle ([shift={(16,32-8*\r)}]path picture bounding box.south west);
                                                        \fill[#1!\c] ([shift={(16,24-8*\r)}]path picture bounding box.south west) rectangle ([shift={(24,32-8*\r)}]path picture bounding box.south west);
                                                        \fill[#1!\d] ([shift={(24,24-8*\r)}]path picture bounding box.south west) rectangle ([shift={(32,32-8*\r)}]path picture bounding box.south west);
                                                }
                                }},
        grid42/.style={kcpcell, minimum width=16pt, minimum height=32pt,
                        path picture={
                                        \foreach \r/\a/\b in {0/20/60,1/70/30,2/40/80,3/10/50} {
                                                        \fill[#1!\a] ([shift={(0,24-8*\r)}]path picture bounding box.south west) rectangle ([shift={(8,32-8*\r)}]path picture bounding box.south west);
                                                        \fill[#1!\b] ([shift={(8,24-8*\r)}]path picture bounding box.south west) rectangle ([shift={(16,32-8*\r)}]path picture bounding box.south west);
                                                }
                                }},
        griddiag/.style={kcpcell, minimum width=32pt, minimum height=32pt,
                        path picture={
                                        \foreach \i/\s in {0/50,1/80,2/30,3/60} {
                                                        \fill[#1!\s] ([shift={(8*\i,24-8*\i)}]path picture bounding box.south west) rectangle ([shift={(8*\i+8,32-8*\i)}]path picture bounding box.south west);
                                                }
                                }},
        gridcol/.style={kcpcell, minimum width=8pt, minimum height=32pt,
                        path picture={
                                        \foreach \r/\s in {0/30,1/50,2/70,3/90} {
                                                        \fill[#1!\s] ([shift={(0,24-8*\r)}]path picture bounding box.south west) rectangle ([shift={(8,32-8*\r)}]path picture bounding box.south west);
                                                }
                                }},
        gridrow/.style={kcpcell, minimum width=32pt, minimum height=8pt,
                        path picture={
                                        \foreach \c/\s in {0/30,1/50,2/70,3/90} {
                                                        \fill[#1!\s] ([shift={(8*\c,0)}]path picture bounding box.south west) rectangle ([shift={(8*\c+8,8)}]path picture bounding box.south west);
                                                }
                                }},
        gridrowv/.style={kcpcell, minimum width=16pt, minimum height=8pt,
                        path picture={
                                        \foreach \c/\s in {0/50,1/90} {
                                                        \fill[#1!\s] ([shift={(8*\c,0)}]path picture bounding box.south west) rectangle ([shift={(8*\c+8,8)}]path picture bounding box.south west);
                                                }
                                }},
        grid42s/.style={kcpcell, minimum width=12pt, minimum height=24pt,
                        path picture={
                                        \foreach \r/\a/\b in {0/20/60,1/70/30,2/40/80,3/10/50} {
                                                        \fill[#1!\a] ([shift={(0,18-6*\r)}]path picture bounding box.south west) rectangle ([shift={(6,24-6*\r)}]path picture bounding box.south west);
                                                        \fill[#1!\b] ([shift={(6,18-6*\r)}]path picture bounding box.south west) rectangle ([shift={(12,24-6*\r)}]path picture bounding box.south west);
                                                }
                                }},
        grid4s/.style={kcpcell, minimum width=24pt, minimum height=24pt,
                        path picture={
                                        \foreach \r/\a/\b/\c/\d in {0/20/60/10/40,1/70/30/50/20,2/40/80/20/60,3/10/50/30/70} {
                                                        \fill[#1!\a] ([shift={(0,18-6*\r)}]path picture bounding box.south west) rectangle ([shift={(6,24-6*\r)}]path picture bounding box.south west);
                                                        \fill[#1!\b] ([shift={(6,18-6*\r)}]path picture bounding box.south west) rectangle ([shift={(12,24-6*\r)}]path picture bounding box.south west);
                                                        \fill[#1!\c] ([shift={(12,18-6*\r)}]path picture bounding box.south west) rectangle ([shift={(18,24-6*\r)}]path picture bounding box.south west);
                                                        \fill[#1!\d] ([shift={(18,18-6*\r)}]path picture bounding box.south west) rectangle ([shift={(24,24-6*\r)}]path picture bounding box.south west);
                                                }
                                }},
        griddiags/.style={kcpcell, minimum width=24pt, minimum height=24pt,
                        path picture={
                                        \foreach \i/\s in {0/50,1/80,2/30,3/60} {
                                                        \fill[#1!\s] ([shift={(6*\i,18-6*\i)}]path picture bounding box.south west) rectangle ([shift={(6*\i+6,24-6*\i)}]path picture bounding box.south west);
                                                }
                                }},
        gridcols/.style={kcpcell, minimum width=6pt, minimum height=24pt,
                        path picture={
                                        \foreach \r/\s in {0/30,1/50,2/70,3/90} {
                                                        \fill[#1!\s] ([shift={(0,18-6*\r)}]path picture bounding box.south west) rectangle ([shift={(6,24-6*\r)}]path picture bounding box.south west);
                                                }
                                }},
        gridrows/.style={kcpcell, minimum width=24pt, minimum height=6pt,
                        path picture={
                                        \foreach \c/\s in {0/30,1/50,2/70,3/90} {
                                                        \fill[#1!\s] ([shift={(6*\c,0)}]path picture bounding box.south west) rectangle ([shift={(6*\c+6,6)}]path picture bounding box.south west);
                                                }
                                }},
}

\begin{tikzpicture}[
                x=1pt,
                y=1pt,
                outer sep=0pt,
                op/.style={font=\footnotesize, text=kcpfigink, inner sep=1pt},
                note/.style={font=\scriptsize, text=kcpfiggray, inner sep=1pt},
        ]
        \path[use as bounding box] (-10,52) rectangle (406,108);

        \node[grid4=brickred] (Mt) at (18,88) {};
        \node[op, right=3pt of Mt] (eq1) {$=$};
        \node[op, right=1pt of eq1] (prod) {$\prod_r$};
        \node[op, right=0pt of prod] (blp) {$\Biggl($};
        \node[op, right=0pt of blp] (lp) {$\Bigl($};
        \node[griddiags=black, right=1pt of lp] (I) {};
        \node[op, right=0pt of I] (minus) {$-$};
        \node[gridcols=midnightblue, right=1pt of minus] (k) {};
        \node[op, right=0pt of k] (times) {$\times$};
        \node[gridrows=midnightblue, anchor=west] at ([shift={(3pt,-3pt)}]k.north east) (kT) {};
        \node[op, right=28pt of k] (rp) {$\Bigr)$};
        \node[griddiag=brickred, right=0pt of rp] (alpha) {};
        \node[op, right=2pt of alpha] (brp) {$\Biggr)$};

        \node[grid42=kvcolor, right=18pt of brp] (s3) {};
        \node[op, right=4pt of s3] (eq3) {$=$};
        \node[grid42=midnightblue, right=4pt of eq3] (t3) {};
        \node[op, right=4pt of t3] (p3) {$+$};
        \node[grid4=brickred, right=4pt of p3] (m3) {};
        \node[op, right=3pt of m3, yshift=-1pt] (sumj) {$\sum\nolimits_j$};
        \node[grid4=brickred,opacity=0.4] at ($(sumj.east)+(23,3)$) (m2b) {};
        \node[grid4=brickred, right=4pt of sumj] (m3b) {};
        \node[grid42=midnightblue, right=7pt of m3b] (s1in3) {};

        \coordinate (sepmid) at ($(alpha.east)!0.5!(s3.west)$);
        \draw[draw=kcpfiggray, line width=0.4pt, dash pattern=on 3pt off 3pt]
        ([shift={(0,3)}]sepmid |- Mt.north) -- ([shift={(0,-3)}]sepmid |- Mt.south);

        \node[note, below=12pt of Mt, anchor=base] {$\mathbf{M}_{[i+1]}^{t \leftarrow 1}$};
        \node[note, anchor=base] at ($(I |- alpha.south)!0.5!(kT |- alpha.south)+(0,-12)$) {$\mathbf{I}-\beta_r\bm{k}_r\bm{k}_r^{\top}$};
        \node[note, below=12pt of alpha, anchor=base] {$\operatorname{Diag}(\bm{\alpha}_r)$};
        \node[note, below=12pt of s3, anchor=base] {$\mathbf{S}_{[i+1]}^{t}$};
        \node[note, below=12pt of t3, anchor=base] {$\widetilde{\mathbf{S}}_{[i+1]}^{t}$};
        \node[note, below=12pt of m3, anchor=base] {$\mathbf{M}_{[i+1]}^{t \leftarrow 1}$};
        \node[note, below=12pt of m3b, anchor=base] {$\prod_l\mathbf{M}_{[l]}^{T_l \leftarrow 1}$};
        \node[note, below=12pt of s1in3, anchor=base] {$\widetilde{\mathbf{S}}_{[j]}^{T_j}$};
\end{tikzpicture}

    \end{adjustbox}
\end{center}
where $\mathbf{M}_{[i+1]}^{t \leftarrow 1}$ denotes the cumulative transition of the first $t$ local tokens.
The first term contains the state generated by the local tokens, whereas the second term propagates the context from preceding ranks through the local KDA updates.
At $t=T_{i+1}$, both quantities $\mathbf{M}_{[i+1]}^{T_{i+1} \leftarrow 1}$ and $\widetilde{\mathbf{S}}_{[i+1]}^{T_{i+1}}$ can be computed using only the local tokens, before $\mathbf{S}_{[i]}^{T_i}$ is available, and are the fragments each rank exchanges with the others.

The summation in Eq.~\ref{eq:kcp-compose} shows that every state is composed purely from locally computed fragments.
These rank-level updates compose associatively, so the incoming state of each rank can be recovered by a prefix scan~\citep{martin-2018-parallelizing}.
Each rank first computes $\mathbf{M}_{[i]}^{T_i \leftarrow 1}$ and $\widetilde{\mathbf{S}}_{[i]}^{T_i}$ locally, then exchanges both tensors with one \texttt{all-gather}~\citep{yang2024fla}.\footnote{
    The construction builds on DeltaNet context parallelism~\citep{yywang2025deltanetcp}.
    The KDA implementation is available in \href{https://github.com/fla-org/flash-linear-attention/pull/691}{FLA PR~\#691}.
}
After the \texttt{all-gather}, rank $i+1$ reconstructs $\mathbf{S}_{[i]}^{T_i}$ by processing preceding fragments of the same document in order, starting from $\mathbf{S}=\mathbf{0}$ and applying $\mathbf{S}\leftarrow\mathbf{M}_{[j]}^{T_j \leftarrow 1}\mathbf{S}+\widetilde{\mathbf{S}}_{[j]}^{T_j}$ at each fragment.
Therefore, KCP requires only a fixed-size \texttt{all-gather} for recurrent-state synchronization and achieves linear compute scaling.

\subsection{Infra for 3T-class Pre-Training}
\label{sec:infra-training}

\kimi{3} pre-training combines Pipeline Parallelism (PP) with virtual stages (VP)~\citep{huang2019gpipe, narayanan2021efficient}, Expert Parallelism (EP)~\citep{lepikhin2020gshard}, ZeRO-1 Data Parallelism~\citep{rajbhandari2020zero}, Pipeline ZeRO-2 gradient sharding~\citep{glm5team2026glm5vibecodingagentic}, and Context Parallelism (CP, \S\ref{sec:kda-cp})~\citep{jacobs2023deepspeedulyssesoptimizationsenabling}.
The MoE layers employ shared experts replicated across EP ranks, and the all-to-all communication for expert dispatch and combine is overlapped with computation to hide its latency.

\begin{figure}
    \centering
    \includegraphics[width=1\linewidth]{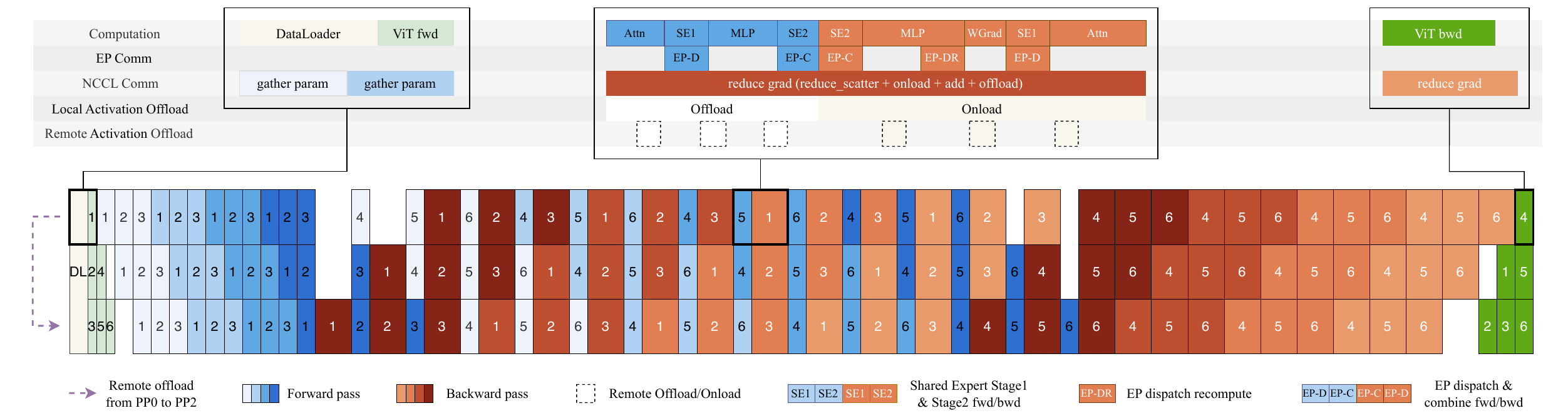}
    \caption{Computation, communication and offloading overlapped in different PP phases.}
    \label{fig:k3pp}
\end{figure}

Natively multimodal pre-training at the 3T-class poses three critical problems: (i) token loads are imbalanced across EP ranks; (ii) activations, gradients, and optimizer states exceed the memory budget; and (iii) the vision encoder's highly variable computation is exposed on the critical path.
The following subsections address these problems in turn: perfectly balanced expert-parallel MoE training (\S\ref{sec:moe_moonep}), memory-efficient training (\S\ref{sec:memory-efficient-training}), and multimodal encoder optimization (\S\ref{sec:multimodal-encoder-optimization}).
Fig.~\ref{fig:k3pp} illustrates the resulting execution schedule.

\subsubsection{Perfectly Balanced Expert-Parallel MoE Training}
\label{sec:moe_moonep}
In conventional EP schemes, token loads are imbalanced across ranks.
The resulting computational imbalance degrades training throughput, and the dynamically varying shapes of routed-expert activations cause substantial memory fragmentation.
We therefore propose MoonEP\footnote{\url{https://github.com/MoonshotAI/MoonEP}}, an EP scheme that achieves perfect load balance with dynamic redundant experts.
MoonEP preserves the overall computation flow of conventional schemes such as DeepEP~\citep{deepep2025} and additionally introduces online planning and migration of redundant experts.
In the forward pass, we plan the redundant experts from the router outputs of the current micro-batch and layer and prefetch them before the routed-expert computation.
In the backward pass, we stage their gradients in a local reduce buffer and, once the computation completes, reduce them back to the gradient buffers of their home ranks.

\paragraph{Perfect balance with bounded redundant experts}
MoonEP requires every rank to receive exactly $S \times K$ tokens, where $S$ is the sequence length and $K$ is the number of experts selected per token, so that all ranks perform identical amounts of computation.
The key question is how many redundant experts suffice to guarantee such a balance.
Let $E$ be the number of experts and $R$ the EP size.
We prove that a balanced plan always exists with at most $E/R$ redundant experts per rank and that this bound is essentially tight (\S~\ref{app:moonep-proof}).
Reserving $E/R$ redundant-expert slots per rank therefore guarantees that planning always admits a feasible solution, so training is never interrupted.
In contrast, prior work such as ECHO~\citep{yan2026scalabletrainingmixtureofexpertsmodels} and UltraEP~\citep{wei2026ultraepunleashmoetraining} presets the number of redundant experts or imposes a per-rank token cap.
Training is then forced to stop whenever no feasible plan exists within the cap, and the cap itself requires manual tuning while still leaving residual imbalance.

\paragraph{Online planning}
Computing the exact optimum at every training step is prohibitively expensive.
We therefore compute exact solutions offline with integer linear programming (ILP) for representative cases as references and design a GPU planning kernel that is near-optimal, incurs negligible overhead, and always respects the $E/R$ upper bound.

\paragraph{Zero-copy communication}
Perfect balance also simplifies the communication path.
We implement a fused permute/unpermute operator in which the planning kernel precomputes the destination of every token, so tokens are sent directly to their expert-grouped positions on remote ranks, and views of the communication buffer are returned directly to the computation, eliminating intermediate copies.
Under worst-case imbalance, supporting the same copy-free data path in DeepEP requires a communication buffer of size $S \times K \times R$, whereas MoonEP requires only a fixed $S \times K$ buffer owing to the perfect balance.

\paragraph{Sync-free execution with static shapes}
In conventional MoE implementations, the per-expert token counts vary across steps and layers, and the host must synchronize with the device at every layer to obtain the actual computation shapes before launching the expert computation, stalling the pipeline between layers.
With perfect balance, every rank receives exactly $S \times K$ tokens and the computation shapes of all layers are statically known. This eliminates the per-layer MoE host synchronization and alleviates the host-side kernel-launch overhead.

\paragraph{Expert-GEMM scheduling and overlap}
Even with the aggregate load perfectly balanced across ranks, the per-expert token counts within each rank remain skewed, and a fixed-order, workload-oblivious schedule turns this skew into an imbalanced makespan across SM workers.
We therefore schedule the routed-expert GEMM with a workload-aware scheduler that adapts its parameters to the current token distribution before launch and keeps them fixed during execution.
A lightweight heuristic selects these parameters using an analytical cost model of hardware metrics, with key coefficients calibrated through offline autotuning.
For the shared experts, we dispatch their GEMMs to a separate stream so that they overlap with other kernels.

\subsubsection{Memory-Efficient Training}
\label{sec:memory-efficient-training}

\paragraph{Unified activation manager}
We design a unified storage abstraction for activations, in which every tensor saved for the backward pass is associated with a pluggable storage backend.
Recomputation, quantization, and offload/remote-offload are merely storage policies under this abstraction and can be freely composed at tensor granularity; policies are declared via lightweight annotations on tensors, fully decoupled from the model code.
Recomputation is performed at function granularity, which supports cross-layer recomputation.
In our implementation, all GPU memory is allocated on the main compute stream and managed within a single memory pool, avoiding multi-stream fragmentation and host-bound overhead; activations are prefetched back at layer granularity and overlapped with computation, introducing negligible extra overhead.
In \kimi{3}, most activations use block-wise FP8 quantization~\citep{kimiteam2025kimik2openagentic, deepseekaiv3} combined with offload/remote-offload, and element-wise operators are configured with recomputation.

\paragraph{Memory-efficient MoE}
In the native MoE implementation, the gradient computation of permuted probs depends on the forward output \texttt{output}.
Inspired by SonicMoE~\citep{guo2025sonicmoeacceleratingmoeio}, we rewrite this gradient through a mathematical transformation into a form that depends only on the intermediate activation \texttt{act\_output} and the upstream gradient \texttt{doutput}, eliminating the backward dependency on \texttt{output} at the cost of an additional lightweight element-wise computation.
Furthermore, in the forward pass of the group GEMM, we save only the input of the dispatch operation; during the backward pass, the input of the group GEMM is recovered by recomputing dispatch.
As shown in Fig.~\ref{fig:k3pp}, the communication introduced by this recomputation can be overlapped with part of the group-GEMM backward computation, eliminating this portion of activation storage at a negligible cost.

\paragraph{Memory-efficient Attention residual}
For the attention residual, we design a companion optimization based on Block AttnRes.
The block representation is generated once at the boundary layer and shared by all subsequent layers, residing directly on the GPU.
The AttnRes computation is entirely wrapped with checkpointing, so the activation saved for the backward pass at each layer is identical to that of the standard residual architecture.
For pipeline parallelism, we adopt cache-based pipeline communication~\citep{kimiteam2026attnres}, in which only newly generated blocks are incrementally transferred between stages and released as soon as the micro-batch finishes, reaching the theoretical lower bound on memory footprint.

\paragraph{Balancing activations across PP ranks}
Under interleaved 1F1B pipeline parallelism, activations are unevenly distributed across PP ranks due to pipeline warmup, and the number of resident activations decreases as the PP rank increases.
To avoid out-of-memory (OOM) errors, we remotely offload activations to the memory of other PP ranks using the Mooncake Transfer Engine~\citep{qin2024mooncake}, achieving balanced activation memory across PP ranks.

\paragraph{Pipeline ZeRO-2 gradient sharding and offloading}
\label{para:pipeline-zero2}
Beyond activations, we use Pipeline ZeRO-2 gradient sharding~\citep{glm5team2026glm5vibecodingagentic} to shard gradients across data-parallel (DP) ranks.
Furthermore, we store the sharded gradients in CPU memory to reduce peak GPU memory usage, while keeping the double grad buffer on the GPU.
After gradients are reduced across DP ranks into the double grad buffer, they are accumulated into the CPU shards.

\paragraph{P2P-based Muon orthogonalization}
The distributed optimizer shards parameters evenly across DP ranks, whereas the Newton--Schulz orthogonalization in Muon requires the full parameter matrix, necessitating a communication step to gather complete parameters before each update.
The naive approach performs an all-gather over the entire parameter buffer on every rank~\citep{liu-2025-moonlight}, which incurs a substantial memory footprint on top of making communication the primary bottleneck at scale.
Instead, each rank retrieves only the shards of its locally owned parameters via peer-to-peer (P2P) communication with the corresponding owner ranks, eliminating the full-parameter buffer and reducing both memory usage and communication volume.
Communication and computation are further pipelined at the granularity of model-chunk buffers, hiding the communication overhead.

\subsubsection{Multimodal Encoder Optimization}
\label{sec:multimodal-encoder-optimization}

\paragraph{Dynamic CP in multimodal encoder}
In long-context multimodal training, large images and long videos substantially increase the computation time of the vision encoder and cause significant load imbalance across devices.
To address this, we extend context parallelism to such large samples.
A single large image is partitioned along the patch dimension across multiple devices, and attention is computed by gathering key--value pairs (gather-KV) across CP ranks.
In addition, we divide each CP group into several sub-CP groups and distribute multiple large images across them in a load-balanced manner, preventing the communication fraction from growing with scale.
This reduces both the encoder latency of large visual samples and the cross-device load imbalance, allowing the remaining encoder computation to be hidden in pipeline bubbles.

\paragraph{Encoder computation in PP bubbles}
In \kimi{2.5}, we introduced the Decoupled Encoder Process (DEP)~\citep{kimik25}, which splits ViT and text training into separate stages and balances vision forward and backward passes across PP stages.
We observe that, under the interleaved 1F1B pipeline schedule, the text forward passes of the first PP micro-batches are all scheduled at the very beginning, while the text backward passes of the last PP micro-batches finish only at the very end.
We therefore further decompose the ViT computation~\citep{feng2025optimusacceleratinglargescalemultimodal}.
The ViT forward passes of the first PP micro-batches are executed synchronously upfront, the remaining forward passes are scheduled into pipeline bubbles, and the backward passes are handled analogously.
As a result, most of the ViT computation is hidden within pipeline bubbles, largely eliminating the effective overhead of the vision encoder.

\subsection{Infra for 1M Agentic RL}

Scaling agentic RL for a model as large as \kimi{3} to million-token contexts under a bounded compute budget makes resource efficiency a first-order goal. We therefore develop long-context RL infrastructure for efficient training and rollout, together with high-performance, resumable sandboxes for long-horizon environment interaction.

\subsubsection{Long-context RL infrastructure}
We adopt co-located RL training \cite{kimiteam2025kimik2openagentic} to keep each 1M-context \kimi{3} RL experiment within a few hundred GPUs, and use partial rollouts~\cite{kimik15} to reduce tail latency from ultra-long trajectories. This design improves hardware utilization, but long-context rollouts introduce extra DRAM demand for KV-cache retention, which competes with training-side states. Further, achieving high efficiency for both prefill and decoding requires careful prefix management and request scheduling.

\paragraph{External KV cache pool}
At 1M-context multi-step rollout, a prefix KV-cache miss is extremely expensive. Partial rollout exacerbates this issue at the beginning of each iteration, due to many unfinished long prefill requests from the previous iteration arriving at the same time. Speculative decoding further accelerates request turnover within relatively fixed tool-call intervals, increasing prefix-block churn. These issues can trigger preemption and lower the cache hit rate, which is critical for long-context RL.

We therefore decouple prefix retention from GPU residency with a write-back design.
Active decoding blocks remain in GPU KV cache, while reusable idle prefixes are written back to an \textit{external KV cache pool} in CPU DRAM only when it is evicted from GPU, and is prefetched back before the next reuse.
KDA states are offloaded and prefetched together with the corresponding MLA KV cache blocks, keeping their lifecycles aligned.
Compared with a write-through strategy, this policy incurs CPU DRAM usage and transfer bandwidth only for prefixes that leave the active decode path, avoiding redundant CPU copies of blocks that are still resident and active on GPU.

To provide sufficient DRAM for the external pool, we offload training states (model weights and optimizer states) to NVMe after a training iteration finishes. After a rollout iteration, the pool is released to avoid contention with training workloads.

\paragraph{Rollout auto-throttling scheduler}
In multi-step rollout, contexts grow progressively as the trajectory advances, making fixed concurrency based on the full-trajectory average length both hard to estimate and overly conservative early on.
Conversely, setting concurrency too high creates KV cache pressure in later stages and can trigger preemption.
We therefore design an auto-throttling mechanism at the LLM request scheduling layer, using runtime signals such as active request count, queued request count, and KV cache utilization to dynamically control how many requests are sent to the inference engine.
This keeps early rollout well utilized while reducing concurrency as KV cache pressure rises, avoiding both under-saturation and overload without manual tuning.

\paragraph{Gradient-buffer reuse for non-policy model forwarding}
RL loss computation often requires forward-only non-policy models, such as reference models, whose weights are too large to keep resident on GPU.
We keep these weights in CPU memory and materialize them only when needed, backing their parameter tensors with the policy model's FP32 gradient-buffer storage.
This reuses existing GPU memory without extra allocation or fragmentation, and remains safe because the buffers are overwritten when real gradients are later computed.

With ZeRO-2 gradient sharding and offloading (\S~\ref{para:pipeline-zero2}), each GPU retains gradient buffers for only two VPP chunks in \kimi{3} RL training.
We stream reference weights into these slots chunk by chunk: one slot is used for the current forward computation while the other prefetches the next chunk, hiding copy overhead without increasing GPU memory.

\subsubsection{Sandbox Infrastructure}
\label{sec:sandbox}

We employ multiple sandbox runtimes to support the diverse requirements of \kimi{3} post-training and evaluation, including a traditional container-based runtime, a GPU sandbox runtime, and, most notably, a new microVM-based sandbox runtime called AgentENV.

AgentENV\footnote{AgentENV is open-sourced at \url{https://github.com/kvcache-ai/AgentENV}}, developed in collaboration with our partners, is a sandbox system specifically designed for agentic AI workloads.
It is built around three core design goals:

\begin{itemize}[leftmargin=12pt]

    \item \textbf{High-fidelity isolated sandbox runtime} As agents become more capable and tasks more difficult, they tend to explore more aggressively and may even attempt reward hacking.
          On the one hand, this poses unique security challenges: in our early experiments with traditional container-based sandbox runtimes, we observed several kernel panics and deadlocks caused by unintended agent operations.
          On the other hand, we want to permit as much exploration as possible so as not to constrain agent capability, and complex tasks require a sandbox close to a real-world environment --- for example, agents should be able to mount disks, run containers, or even launch virtual machines at will.
          By running isolated microVMs with Firecracker~\citep{agache2020firecracker}, AgentENV provides a level of isolation and fidelity that container-based runtimes cannot match.

    \item \textbf{Flexible sandbox life-cycles for agentic RL} At the low level, AgentENV supports incremental checkpointing and resuming of sandbox states, where only memory pages dirtied since the last checkpoint are saved during checkpointing, achieving checkpoint and resume latencies as low as 133\,ms and 49\,ms, respectively.
          On top of this, AgentENV provides three high-level operations that help improve agentic RL efficiency.
          \textbf{(a) Pause and Resume}: a paused sandbox consumes no memory or CPU resources; a sandbox can therefore be paused while the agent is waiting for the model's inference result, which can account for as much as 98\% of the sandbox lifetime.
          \textbf{(b) Fork}: fork creates a new sandbox from the exact state of the original one while keeping the original running, which is useful for reward judging without side effects.
          \textbf{(c) Snapshot}: snapshots of a sandbox can be saved at regular intervals for error recovery.

    \item \textbf{High efficiency and high density} In our workloads, tens of thousands of sandboxes, each with a unique set of images, may need to be created within seconds.
          We adopt OverlayBD~\citep{li2020dadi} as the image format, together with a custom ublk driver implementation, storage-layer sharing, and P2P transport, achieving sub-second launch latency at large scale.
          We further reduce memory usage with copy-on-write memory and page-cache optimizations, achieving a memory overcommit ratio of up to 6.5$\times$ in real workloads.

\end{itemize}

Throughout \kimi{3}'s training and evaluation, a total of 51{,}219{,}741 sandboxes across 1{,}505{,}678 images were created.

\subsection{Inference and Online Serving}
\label{sec:inference}

Serving \kimi{3} exposes the same challenges from the production side: the hybrid KDA--MLA architecture maintains two fundamentally different caches that must be managed jointly at million-token contexts, its new modules and highly sparse experts demand kernels tailored to each, and production traffic mixes requests whose per-request cost spans three orders of magnitude.
The designs below address these challenges at three levels.
At the engine level, a KDA-aware prefix cache packs the fixed-size recurrent state into the same paged pool as the MLA KV cache and keeps long prefixes reusable across requests.
At the device level, dedicated kernels for KDA decoding, Block AttnRes, and the sparse latent MoE minimize per-token latency and memory traffic.
At the fleet level, cache-aware affinity scheduling and budget-based admission control translate these efficiencies into predictable serving.

\subsubsection{KDA-Aware Prefix Cache Management}

The hybrid architecture in \kimi{3} complicates prefix caching: the KDA recurrent state and the MLA KV cache differ fundamentally in size and lifetime, yet a cached prefix is reusable only when both can be restored together at the same boundary.
We therefore design a KDA-aware prefix cache that manages the two cache types jointly---from a unified paged layout to fine-grained prefix reuse and consistency under concurrent scheduling---keeping million-token prefixes cheap to retain and reusable across requests.

\paragraph{Unified cache layout for hybrid KDA--MLA attention}
Each \kimi{3} block consists of three KDA layers and one Gated MLA layer, whose caches differ fundamentally.
The MLA KV cache grows with sequence length and is paged per token, whereas the KDA recurrent state is fixed in size with a single copy per request.
Maintaining a separate manager for each would duplicate the allocation, eviction, and transfer logic.
We therefore pack KDA states into the same paged block pool as MLA KV, unifying pages to the same byte size so that both page types share one implementation of allocation, reference counting, and eviction.
Within a page, the states of all heads are stored contiguously head by head, so that each head's byte stream is self-contained and serves as the minimal unit of cross-node transfer.
Under prefill/decode disaggregation, when prefill and decode nodes adopt different TP degrees, re-layout is performed on the transfer path with zero GPU-side reshuffling.
This asymmetry proved useful during development: any type-confused access yields garbage rather than plausible data --- a zero-overhead sanity check on the pooled layout.

\paragraph{KDA prefix cache optimization}

Block-hash-based prefix caching reuses the KV cache at the granularity
of one physical block: only complete blocks are hashed, so only
block-aligned prefixes are reusable.

This coupling breaks down in \kimi{3}. Block-hash matching requires
one block size shared by all layers, and a prefix hit is reusable only
if the KDA state at the hit boundary has been persisted. A KDA layer
maintains a single large recurrent state per sequence rather than
per-token entries, so state snapshots are affordable only at sparse
boundaries; the shared block size is therefore forced to 1024--6144
tokens---and, since hashing is tied to the storage block, the hash
granularity as well, although MLA's per-token entries alone would
tolerate much finer blocks. At such a coarse granularity caching is
nearly useless: requests shorter than one block can never be reused,
and chunked prefill exports no cacheable prefix until it crosses a
full block boundary.

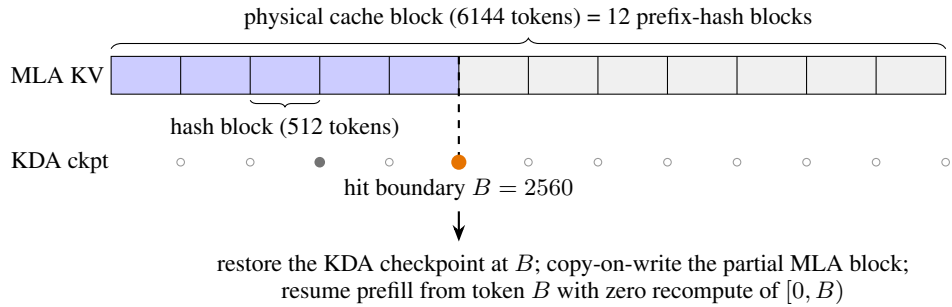
\begin{figure}[!hb]
    \centering
    \begin{tikzpicture}[font=\small]
        \def\hbw{0.92}  
        \foreach \i in {0,...,11} {
                \pgfmathsetmacro{\x}{\i*\hbw}
                \ifnum\i<5\relax
                    \fill[blue!20] (\x,0) rectangle (\x+\hbw,0.5);
                \else
                    \fill[black!6] (\x,0) rectangle (\x+\hbw,0.5);
                \fi
            }
        \draw (0,0) rectangle (12*\hbw,0.5);
        \foreach \i in {1,...,11} { \draw (\i*\hbw,0) -- (\i*\hbw,0.5); }
        \draw[decorate,decoration={brace,amplitude=4pt}] (0,0.6) -- (12*\hbw,0.6)
        node[midway,above=4pt,font=\footnotesize]
        {physical cache block (6144 tokens) = 12 prefix-hash blocks};
        \draw[decorate,decoration={brace,amplitude=2.5pt,mirror}] (2*\hbw,-0.08) -- (3*\hbw,-0.08)
        node[midway,below=3pt,font=\footnotesize] {hash block (512 tokens)};
        \node[font=\footnotesize,anchor=west] at (-1.45,0.25) {MLA KV};
        \node[font=\footnotesize,anchor=west] at (-1.45,-0.9) {KDA ckpt};
        \draw[dashed,thick] (5*\hbw,0.5) -- (5*\hbw,-0.9);
        \foreach \k in {1,...,12} { \draw[black!40] (\k*\hbw,-0.9) circle (0.05); }
        \foreach \k in {3} { \fill[black!55] (\k*\hbw,-0.9) circle (0.07); }
        \fill[orange!90!black] (5*\hbw,-0.9) circle (0.10);  
        \node[font=\footnotesize,anchor=north] at (5*\hbw,-1.0)
        {hit boundary $B=2560$};
        \draw[-{Stealth[length=2.2mm]},thick] (5*\hbw,-1.55) -- (5*\hbw,-1.95);
        \node[font=\footnotesize,anchor=north,align=center] at (6.5*\hbw,-2.0)
        {restore the KDA checkpoint at $B$; copy-on-write the partial MLA block;\\
            resume prefill from token $B$ with zero recompute of $[0,B)$};
    \end{tikzpicture}
    \caption{\textbf{Fine-grained prefix caching within a physical cache block.}
        A 6144-token physical block contains twelve 512-token hash blocks, with cached MLA blocks shown in blue and empty blocks in light gray.
        The markers below show the KDA checkpoint status at each hash boundary.
        An open circle ($\circ$) denotes a boundary without a stored checkpoint, a gray dot (\textcolor{black!55}{$\bullet$}) denotes a persisted KDA checkpoint, and an orange dot (\textcolor{orange!90!black}{$\bullet$}) marks the checkpoint hit at $B=2560$.
        Persisted checkpoints are sparse and typically coincide with conversation-turn boundaries.
        The request reuses the five MLA hash blocks and the KDA checkpoint at $B$, then resumes prefill without recomputing $[0,B)$.}
    \label{fig:kda-prefix-cache}
\end{figure}

We therefore decouple the two granularities. Prefix hashing runs on
fine \emph{hash blocks} (e.g., 512 tokens) inside MLA pages, while the
physical block remains the coarse allocation unit. Alignment runs the
other way for KDA: checkpoints of the recurrent state are saved only
at (a sparse subset of) MLA's hash endpoints---the only positions a
lookup can ever reference.

During prefill, a partially filled MLA page is registered in the
prefix-cache index under the chained hash of its last complete hash
block, where each hash covers all preceding hash blocks so that
matching an endpoint certifies the whole prefix up to it; the
registered endpoint advances as the page fills. Meanwhile, after each
forward pass, the KDA kernel persists the recurrent state at the last
hash-aligned position processed. Checkpoints are large, so
intermediate checkpoints superseded as the request advances are
recycled, while those at conversation-turn boundaries are retained for
cross-request reuse. Cached checkpoints are read-only snapshots: a hit
restores the state by copying it into the request's private running
state before the next forward pass, and new checkpoints are written to
fresh slots, so a checkpoint visible to other requests is never
mutated in place.

Lookup proceeds in two stages (Fig.~\ref{fig:kda-prefix-cache}). The
MLA stage matches whole physical blocks by chained hash and, at the
first missing block, falls back to the hash endpoints inside it, so
partially filled pages remain hittable. The KDA stage then requires a
checkpoint at the candidate boundary in every KDA cache group, each of
which maintains an independent recurrent state. The hit is the longest
boundary satisfying both stages---always a multiple of the hash block,
and never required to be a multiple of the physical block. In
Fig.~\ref{fig:kda-prefix-cache}, a request whose first 2800 tokens
match the cached prefix hits at $B=2560=5\times512$, deep inside a
6144-token physical block, and resumes prefill from token $B$ instead
of recomputing $[0,B)$.

\paragraph{Consistency under concurrent scheduling}
The remaining design points are each dictated by a concrete failure mode of sharing partially filled blocks, in a setting where a hit block is at once a shared cache entry and the growth point of a private request, and where the MLA and KDA cache groups must agree on every hit boundary.
First, all cache groups draw blocks from one shared free list, so allocating a private copy for one group could evict a block that another group has just hit; every hit block is therefore pinned across all groups before anything is allocated.
Second, the copy into the private block executes on the GPU immediately before the forward pass, so a block allocated or registered within the current scheduling step would still hand the previous owner's bytes to a reader; such blocks are excluded from matching until their copies land.
Third, a checkpoint can restore a request only if it exists in every KDA group, so evicting one group's checkpoint atomically invalidates its siblings --- a checkpoint is either hittable in every group or in none.
With these mechanisms, every registered state always corresponds to exactly its declared token prefix, and prefix caching for hybrid KDA--MLA models reaches the same generality as for full-attention models: any shared prefix is reusable at any 512-token boundary, independently of request length, chunking, or scheduling interleaving.

\subsubsection{High-Performance Kernels}
\kimi{3} introduces several new architectural modules: KDA (\S\ref{sec:kda}), Block AttnRes (\S\ref{sec:attnres}), and Stable LatentMoE (\S\ref{sec:stable-latent-moe}). We optimize the kernel implementation for each.

\paragraph{KDA}
\label{sec:kda-decoding}

Compared with KDA prefill (\S\ref{sec:kda-codesign}), KDA decoding presents a distinct set of challenges: the primary bottleneck shifts from exploiting parallelism to efficiently managing the evolving recurrent state, which is updated in place at every decoding step. This in-place update becomes problematic in MTP-based speculative decoding: if verification rejects a subset of the drafted tokens, the state has already advanced beyond the last accepted token and cannot be trivially rolled back. Maintaining a state snapshot for each draft position would enable rollback, but would also multiply state traffic --- a cost that dominates at the large batch sizes typical of online serving.

The state after any accepted draft prefix, however, is fully determined by the projected inputs of the draft tokens, which are far smaller than the state itself.
We therefore cache only these projected inputs, rebuild the states of accepted tokens on-chip, and write back the states of the verified and bonus tokens, a design independently proposed in the concurrent work ReplaySSM~\citep{replayssm2026}.
The replayed tokens, the bonus token, and the next draft window share one recurrent loop inside a single fused kernel covering short convolution, input normalization, gating, the KDA recurrence, and output normalization.
Verification latency grows sub-linearly with the number of tokens verified and remains below that of state-caching baselines.
Because the projection caches never leave the decode stage, prefix caching and prefill--decode disaggregation operate on the same payload as in non-speculative serving.

\paragraph{Block AttnRes}
Block AttnRes~\citep{kimiteam2026attnres} follows a two-phase schedule: a batched inter-block pass reads the cached block representations once per block, after which each layer folds in the intra-block partial sum through an online-softmax merge~\citep{milakov2018online}.
Memory access accounts for a substantial fraction of the cost of these kernels in both prefill and decoding, so our optimizations in both stages focus primarily on memory efficiency.

For prefill, materializing the block representations on every tensor-parallel (TP) rank would incur substantial redundant memory consumption. We therefore adopt sequence parallelism (SP) for activations: the TP all-reduce is decomposed into a reduce-scatter and an all-gather, with the intra-block kernel inserted between the two collectives, operating on the sequence-sharded hidden states so that the block representations of each token are materialized on exactly one rank.
This eliminates the additional memory consumption and reduces the I/O overheads of Block AttnRes during prefill.

For decoding, we launch the inter-block kernel on a side stream so that it overlaps with independent computation on the main stream. The intra-block kernel is instead streamlined through fusion: the merging of the AttnRes output with its partial-sum update, together with the subsequent RMSNorm, is fused into the preceding TP all-reduce, eliminating a dedicated kernel for the intra-block phase.
Together, these optimizations hide the latency of the inter-block pass and reduce the memory traffic of the intra-block phase.

\paragraph{Stable LatentMoE}

Stable LatentMoE increases both the total number of experts and the number of activated experts per token. The resulting growth in both the expert space and the per-token expert count raises scheduling and coordination overheads, making it difficult for conventional MoE kernels to sustain high hardware utilization. These challenges motivate dedicated kernel optimizations for this module.

To mitigate the overhead of the latent GEMMs, we adopt three optimizations. First, we fuse the latent down-projection with the MoE router into a single GEMM. Second, we shard latent weight matrices across ranks and fuse the output all-gather into the GEMM epilogue using multimem store instructions. Finally, we overlap the resulting communication with other operators, such as the shared-expert computation. Together, these optimizations eliminate redundant weight traffic and duplicated computation, while hiding the communication latency behind computation.

For routed experts, at small batch sizes, the group GEMMs reduce to memory-bound streaming of weight matrices --- a regime for which conventional tile-centric kernels are poorly suited due to their compute-oriented design and preprocessing overheads.
We instead build the MoE decoding kernel upon the token-centric design of WarpDecode~\citep{warpdecode2026}, in which each warp is responsible for one output neuron and streams the associated weights directly from memory. To further increase parallelism, we subdivide each warp into finer-grained lane teams, each processing a disjoint subset of experts, followed by a warp-wide reduction of the partial results. In addition, the weight layout is permuted offline at a one-time preprocessing cost, substantially reducing the runtime dequantization overhead.

\subsubsection{Fleet-Level Scheduling}
Beyond a single serving instance, the challenge shifts from per-request efficiency to predictability: a prefix-cache miss costs orders of magnitude more than a hit, and a burst of million-token requests can starve short ones.
We propose two fleet-level scheduling policies to address this: cache-aware affinity scheduling routes each session to the cluster holding its prefix cache while bounding the cost of cluster failures, and budget-based admission control grants each request class its own resource budget so that bursty long-context traffic cannot degrade system-wide SLOs.

\paragraph{Cache-aware affinity scheduling}
At 1M context, a typical coding input carries a prefix of 400K tokens but requires a prefill increment of only 4K tokens, so a prefix-cache hit avoids re-prefilling the entire prefix and is orders of magnitude cheaper than a miss.
We therefore route each request to the cluster that holds its prefix cache, as moving the cache to another cluster would require transferring it over inter-cluster links far slower than the intra-cluster fabric.
This cache-aware affinity, however, binds each session to a single cluster, whose failure would interrupt all sessions bound to it.
Consistent hashing therefore pins each session to two clusters, a primary that serves its traffic and a pre-assigned secondary that takes over when the primary fails.
The secondary holds none of the session's prefix cache and must re-prefill it upon failover.
Since consistent hashing distributes the secondary assignments of different sessions uniformly across the fleet, this re-prefill work is divided among many clusters rather than concentrated on one.
Cache locality is thus preserved in the common case, while the impact of any single cluster failure remains bounded.

\paragraph{Budget-based admission control}
Production traffic mixes short requests under 2K tokens with ultra-long requests up to 1M tokens, so the per-request cost spans roughly three orders of magnitude and the total load imposed by any fixed number of requests is highly unpredictable.
Capacity planning, queueing models, and rate-limiting quotas based on the ``average request'' all break down under this variance.
In a typical failure mode, a burst of long-context requests saturates the available compute, and short requests arriving afterwards cannot be scheduled promptly, degrading time to first token (TTFT) across all traffic.
We therefore adopt budget-based admission control, allocating separate resource budgets to different request classes so that bursty long-context traffic consumes at most its own share of the capacity and cannot degrade system-wide SLOs experienced by other classes.
\section{Evaluations}
\subsection{Main Results}
\subsubsection{Benchmarks}
We evaluate \kimi{3} on a comprehensive benchmark suite organized along four broad capability axes:
\begin{itemize}[leftmargin=12pt]
  \item \textbf{Reasoning \& Knowledge}: GPQA Diamond~\citep{rein2024gpqa}, CritPt~\citep{artificialanalysis}, AA-LCR~\citep{aalcr}, and Humanity's Last Exam (HLE-Full, with and without tools)~\citep{phan2025humanitysexam}.
  \item \textbf{Coding}: DeepSWE~\citep{deepswe}, ProgramBench~\citep{programbench}, Terminal-Bench~2.1~\citep{merrill2026terminal}, FrontierSWE~\citep{frontierswe}, SWE-Marathon~\citep{swemarathon}, PostTrainBench~\citep{posttrainbench}, MLS-Bench-Lite~\citep{lyu2026mlsbench}, and SciCode~\citep{tian2024scicode,artificialanalysis}.
  \item \textbf{Agentic}: BrowseComp~\citep{wei2025browsecomp}, DeepSearchQA~\citep{vedula2025deepsearchqa}, ResearchRubrics~\citep{sharma2026researchrubrics}, Toolathlon-Verified~\citep{li2025toolathlon}, MCPMark-Verified~\citep{wu2025mcpmark}, MCP-Atlas~\citep{bandi2026mcpatlas}, AutomationBench~\citep{shepard2026automationbench}, JobBench~\citep{li2026jobbench}, GDPval-AA v2~\citep{patwardhan2025gdpval}, AA-Briefcase~\citep{artificialanalysis,aabriefcase}, Agents' Last Exam (ALE)~\citep{agentslastexam, sun2026agentsexam}, APEX-Agents~\citep{vidgen2026apexagents}, OfficeQA Pro~\citep{opsahlong2026officeqa}, SpreadsheetBench 2~\citep{zhu2026spreadsheetbench2}, OSWorld-Verified~\citep{xie2024osworld} and OSWorld~2.0~\citep{yuan2026osworld20benchmarkingcomputeruse}, SaaS-Bench~\citep{shi2026saasbench}, $\tau^3$-Banking~\citep{tau3banking,artificialanalysis}, Harvey Lab-AA~\citep{artificialanalysis, harveylabbench}, CorpFin~v2~\citep{valscorpfin}, Finance Agent~v2~\citep{valsfinanceagent}, and Legal Research Bench~\citep{valsailegalresearch}.
  \item \textbf{Vision}: WorldVQA~\citep{worldvqa}, OmniDocBench~\citep{ouyang2025omnidocbench}, PerceptionBench~\citep{perceptionbench}, Video-MME~\citep{fu2024videomme}, MMVU~\citep{zhao2025mmvu}, and BabyVision~\citep{chen2026babyvision} with Python tool. MMMU-Pro~\citep{yue2024mmmupro}, CharXiv (RQ)~\citep{wang2024charxiv}, Math-Vision~\citep{wang2024mathvision}, and ZeroBench-main~\citep{roberts2025zerobench}, each with and without Python tool augmentation. 
\end{itemize}

\newcolumntype{P}[1]{>{\centering\arraybackslash}p{#1}}
\newcolumntype{L}[1]{>{\raggedright\arraybackslash}p{#1}}
\begin{table}[tp]
\centering
\footnotesize
\setlength{\tabcolsep}{3pt}
\caption{Performance comparison of Kimi K3 against proprietary and open-source models. \textbf{Bold} denotes the best result for each benchmark and \uline{underline} the second-best. Unless otherwise noted, Kimi K3 results are obtained with reasoning effort set to \texttt{max} and temperature equal to $1.0$. For HLE-Full, MMMU-Pro, CharXiv (RQ), Math-Vision, and ZeroBench, each cell reports the scores without and with tool augmentation (general tools for HLE-Full, Python for the vision benchmarks), in that order. $^\dagger$On the official Agents' Last Exam leaderboard, the Claude Fable 5 entry runs at xhigh effort with 40\% of tasks annotated as downgraded.}
\label{tab:k3_eval}
\begin{tabular}{@{}l | P{1.9cm} | P{2.1cm} P{1.9cm} P{1.9cm} P{1.7cm} | P{1.7cm}@{}}
\toprule
& & \multicolumn{4}{c|}{\textbf{Proprietary}} & \multicolumn{1}{c}{\textbf{Open Weight}} \\
\cmidrule(l{2pt}r{2pt}){3-6} \cmidrule(l{2pt}r{2pt}){7-7}
\textbf{Benchmark} & \textbf{Kimi K3 (max)} & \textbf{Claude Fable 5 (max, w/ fallback)} & \textbf{GPT-5.6 Sol (max)} & \textbf{Claude Opus 4.8 (max)} & \textbf{GPT-5.5 (xhigh)} & \textbf{GLM-5.2 (max)} \\
\midrule
\multicolumn{7}{@{}l}{\textbf{Reasoning \& Knowledge}} \\
GPQA Diamond & \uline{93.5} & 92.6 & \textbf{94.1} & 91.0 & \uline{93.5} & 91.2 \\
CritPt & 23.4 & \uline{28.6} & \textbf{32.3} & 20.9 & 27.1 & 20.9 \\
AA-LCR & \textbf{74.7} & 70.0 & 73.7 & 67.7 & \uline{74.3} & 71.3 \\
HLE-Full & 43.5 / 56.0 & \textbf{53.3} / \textbf{63.0} & 44.5 / \uline{58.0} & \uline{49.8} / 57.9 & 41.4 / 52.2 & - \\
\midrule
\multicolumn{7}{@{}l}{\textbf{Coding}} \\
DeepSWE & 67.5 & \uline{70.0} & \textbf{73.0} & 59.0 & 67.0 & 46.2 \\
ProgramBench & \textbf{77.8} & 76.8 & \uline{77.6} & 71.9 & 70.8 & 63.7 \\
Terminal-Bench 2.1 & \uline{88.3} & 88.0 & \textbf{88.8} & 84.6 & 83.4 & 82.7 \\
FrontierSWE & \uline{81.2} & \textbf{86.6} & 71.3 & 66.7 & 64.9 & 67.3 \\
SWE-Marathon & \textbf{42.0} & 35.0 & 39.0 & \uline{40.0} & 14.0 & 13.0 \\
PostTrainBench & \uline{36.6} & \textbf{41.4} & 34.6 & 34.1 & 28.4 & 34.3 \\
MLS-Bench-Lite & \uline{48.3} & \textbf{49.9} & 46.2 & 42.8 & 35.5 & 40.4 \\
SciCode & \uline{58.7} & \textbf{60.2} & 56.1 & 53.5 & 56.1 & 50.5 \\
\midrule
\multicolumn{7}{@{}l}{\textbf{Agentic}} \\
BrowseComp & \textbf{91.2} & 88.0 & \uline{90.4} & 84.3 & 84.4 & - \\
DeepSearchQA (F1) & \textbf{95.0} & \uline{94.2} & - & 93.1 & - & - \\
ResearchRubrics & \textbf{76.2} & - & \uline{73.8} & 73.5 & 64.0 & 71.1 \\
GDPval-AA v2 (Elo) & 1686 & \textbf{1747} & \uline{1736} & 1593 & 1491 & 1510 \\
Toolathlon-Verified & \uline{76.5} & \textbf{77.9} & 74.9 & 76.2 & 73.5 & 59.9 \\
MCPMark-Verified & \textbf{94.5} & 87.4 & \uline{92.9} & 76.4 & \uline{92.9} & - \\
MCP-Atlas & \uline{84.2} & \textbf{84.7} & 83.6 & 83.6 & 82.8 & 82.6 \\
AutomationBench & \textbf{30.8} & 29.1 & \uline{29.7} & 27.2 & 22.7 & 12.9 \\
JobBench & \uline{54.3} & \textbf{57.4} & 45.4 & 48.4 & 38.3 & 43.4 \\
AA-Briefcase (Elo) & \uline{1548} & \textbf{1583} & 1495 & 1354 & 1158 & 1260 \\
Agents' Last Exam & \uline{28.3} & 25.7$^{\dagger}$ & \textbf{29.6} & 27.0 & 26.6 & 20.4 \\
APEX-Agents & \uline{41.0} & \textbf{43.3} & 39.9 & 39.4 & 38.5 & 35.6 \\
OfficeQA Pro & 63.3 & \textbf{69.9} & 63.2 & \uline{63.9} & 60.9 & 41.4 \\
SpreadsheetBench 2 & \textbf{34.8} & \uline{34.7} & 32.4 & 31.6 & 29.1 & 28.1 \\
OSWorld-Verified & \uline{84.8} & \textbf{85.0} & 83.0 & 83.4 & 79.0 & - \\
OSWorld 2.0 & 58.3 & \textbf{66.1} & \uline{62.6} & 55.7 & 49.5 & - \\
SaaS-Bench & \uline{60.1} & - & \textbf{61.4} & 56.1 & 43.8 & - \\
$\tau^3$-Banking & \textbf{33.4} & 26.8 & \uline{33.0} & 27.6 & 31.3 & 26.8 \\
Harvey Lab-AA & \textbf{94.6} & \uline{93.6} & 87.2 & 91.1 & 86.3 & 91.0 \\
CorpFin v2 & \uline{71.6} & \textbf{71.8} & 64.4 & 66.7 & 68.4 & 66.1 \\
Finance Agent v2 & \uline{54.4} & \textbf{56.3} & 53.8 & 53.9 & 51.8 & 49.7 \\
Legal Research Bench & 44.2 & \textbf{49.5} & \uline{48.1} & 43.8 & 40.4 & 31.3 \\
\midrule
\multicolumn{7}{@{}l}{\textbf{Vision}} \\
WorldVQA ForceAnswer & \uline{51.0} & \textbf{56.7} & 41.8 & 39.1 & 38.5 & - \\
OmniDocBench & \textbf{91.1} & \uline{89.8} & 85.8 & 87.9 & 89.4 & - \\
PerceptionBench & \uline{58.5} & 57.2 & \textbf{59.7} & 47.2 & 55.8 & - \\
Video-MME (w/ sub) & \textbf{90.0} & - & \uline{89.5} & 86.0 & 89.3 & - \\
MMVU & \textbf{82.1} & - & 81.2 & 79.2 & \uline{81.7} & - \\
BabyVision~w/ Python & 85.7 & \textbf{90.5} & \uline{88.9} & 81.2 & 83.6 & - \\
MMMU-Pro & \uline{81.6} / 83.4 & 81.2 / \textbf{86.5} & \textbf{83.0} / \uline{84.6} & 78.9 / 82.7 & 81.2 / 83.2 & - \\
CharXiv (RQ) & \uline{84.8} / \uline{91.3} & \textbf{88.9} / \textbf{93.5} & 84.6 / 89.1 & 80.5 / 89.9 & 84.1 / 89.0 & - \\
Math-Vision & 94.3 / \uline{97.8} & \uline{94.8} / \textbf{98.6} & \textbf{95.8} / \uline{97.8} & 86.7 / 97.1 & 92.2 / 96.8 & - \\
ZeroBench-main (pass@5) & \textbf{23.0} / \uline{41.0} & \textbf{23.0} / \textbf{46.0} & 17.0 / 35.0 & 17.0 / 34.0 & \uline{22.0} / \uline{41.0} & - \\
\bottomrule
\end{tabular}
\end{table}

\subsubsection{Baselines}
We benchmark against state-of-the-art proprietary and open-source models. For proprietary models, we compare against Claude Fable 5~\citep{claudefable5}, GPT-5.6 Sol~\citep{gpt56sol}, Claude Opus 4.8~\citep{claudeopus48}, and GPT-5.5~\citep{gpt55}. The results of Claude Fable 5 include fallback behaviors and the results of GPT-5.6 Sol include potential cyberguards. For open-source models, we include GLM-5.2~\citep{glm5zhipu}. All models are evaluated at maximum reasoning effort, except GPT-5.5, which uses the ``xhigh'' setting.

\subsubsection{Evaluation Configurations}
All \kimi{3} evaluations use reasoning effort \texttt{max} and temperature = $1.0$. For single-step tasks, such as GPQA Diamond, HLE-Full, and vision benchmarks without tools, we set top-p = $0.95$. For agentic tasks, we set top-p = $1.0$. Generally, we recommend using top-p = $0.95$ for reasoning and knowledge tasks, and top-p = $1.0$ for coding and agentic scenarios.

\paragraph{Coding} Each model is evaluated under one of three agentic harnesses: Kimi Code~\citep{kimicode}, Claude Code~\citep{claudecode}, or Codex~\citep{openaicodex}. On DeepSWE, we report results on the v1.1 tasks, with additional reference to the official leaderboard (\kimi{3} attains 67.3 with the mini-SWE-agent harness). On Terminal-Bench~2.1, we report the best score across harnesses for all models. Our SWE-Marathon evaluation is based on an H20-calibrated branch of the official tasks as of July 9, 2026, prior to the final v1.1 release, with Docker images, performance gates, and reference oracles for the GPU tasks recalibrated for H20 but the correctness and anti-cheat validators unchanged; Claude Fable 5 hits fallbacks on 35\% of the tasks. For PostTrainBench, we evaluate \kimi{3}, Claude Fable 5, and GPT-5.6 Sol using the official Harbor implementation at maximum effort, averaged over three runs on H20 GPUs (instead of H100 in the official setting). FrontierSWE dominance scores are recomputed from raw scores using the official evaluation script as of July 16, 2026.

\paragraph{Agentic} For OfficeQA~Pro, each test case provides the agent with the entire PDF corpus rendered as images, with no machine-readable text available. MCP-Atlas is evaluated on the 500-task public subset with a 100-turn limit, using Gemini 3.1 Pro as the judge. AutomationBench is evaluated on the 600-task public subset. For BrowseComp we adopt a context-compaction strategy triggered at 300K tokens; evaluated with the full 1M-token context window and no context management, \kimi{3} achieves 90.4\%.

\paragraph{Vision} Scores are averaged over three runs, except ZeroBench-main, which we run five times following the official setting. MMMU-Pro follows the official protocol, preserving the original input order and prepending images to the text input. For WorldVQA, we observe consistent refusal behavior across models and enforce an answer via prompt engineering.

\paragraph{Third-party results} GDPval-AA v2, AA-Briefcase, $\tau^3$-Banking, Harvey Lab-AA, APEX-Agents, SciCode, AA-LCR, and CritPt scores are cited from Artificial Analysis~\citep{artificialanalysis} as of July 23, 2026. For Harvey Lab-AA, we report the criterion pass rate. CorpFin~v2, Finance Agent~v2, and Legal Research Bench scores are cited from Vals AI~\citep{valsai}. Agents' Last Exam scores are cited from the official leaderboard~\citep{agentslastexam} as of July 23, 2026; we report the leaderboard's primary pass-rate metric. On the leaderboard, each model is paired with a specific harness: \kimi{3} with Kimi Code; GPT-5.6 Sol, GPT-5.5 with Codex; and Claude Fable 5, Claude Opus 4.8, and GLM-5.2 with Claude Code. Toolathlon-verified and JobBench scores are cited from their official leaderboards~\citep{toolathlon_leaderboard, jobbench_leaderboard} as of July 24, 2026.

\subsubsection{Results}
\label{sec:eval_results}
Table~\ref{tab:k3_eval} provides a comprehensive comparison of \kimi{3} against both proprietary and open-source baselines. Overall, \kimi{3} closely trails the strongest proprietary models, Claude Fable 5 and GPT-5.6 Sol, while consistently outperforming Claude Opus 4.8, GPT-5.5, and GLM-5.2 across the benchmark suite. We highlight key observations across core capability domains below:

\paragraph{Reasoning \& Knowledge}
On graduate-level reasoning, \kimi{3} is competitive with the frontier, scoring 93.5\% on GPQA Diamond. However, a gap remains on research-level tasks: on HLE-Full it trails Claude Fable 5 and GPT-5.6 Sol both with and without tools, at 56.0\% and 43.5\% respectively; and on CritPt it scores 23.4\%, lagging behind Claude Fable 5, GPT-5.6 Sol, and GPT-5.5, indicating that research-level reasoning remains a key direction for improvement.

\paragraph{Coding}
\kimi{3} delivers strong agentic coding performance. It attains the best score on
ProgramBench (77.8\%), and on SWE-Marathon---a GPU-kernel-oriented suite---it
scores 42.0\%, 7 points ahead of Claude Fable 5. On Terminal-Bench~2.1, it nearly matches GPT-5.6 Sol (88.3\% vs.\ 88.8\%). On DeepSWE, it ranks behind Claude Fable 5 and GPT-5.6 Sol but ahead of Claude Opus 4.8 and GPT-5.5. On FrontierSWE, a long-horizon benchmark, it ranks second with a score of 81.2\% as of July 16, 2026, behind only Claude Fable 5 (86.6\%) and well ahead of all other models.

\paragraph{Agentic}
\kimi{3} achieves state-of-the-art results on a broad set of agentic suites, including BrowseComp (91.2\%), DeepSearchQA (95.0\% F1 score), ResearchRubrics (76.2\%), MCPMark-Verified (94.5\%), AutomationBench (30.8\%), SpreadsheetBench~2 (34.8\%), $\tau^3$-Banking (33.4\%), and Harvey Lab-AA (94.6\% criterion pass rate). The main exceptions are the Elo-rated knowledge-work suites, both led by Claude Fable 5: \kimi{3} places third on GDPval-AA~v2
(1{,}686) and second on AA-Briefcase (1{,}548).  Elsewhere it is largely competitive: on CorpFin~v2 and OSWorld-Verified, it finishes just 0.2 points behind Claude
Fable 5 (71.6\% vs.\ 71.8\% and 84.8\% vs.\ 85.0\%, respectively), while the remaining harder
computer-use benchmarks (OSWorld~2.0, SaaS-Bench) are still led by Claude Fable 5 or
GPT-5.6 Sol.

\paragraph{Vision}
\kimi{3} exhibits strong multimodal understanding capabilities, which are further amplified by Python tools: on Math-Vision it reaches 94.3\%, rising to 97.8\% with Python tools, and on the challenging ZeroBench-main it ties Claude Fable 5 at 23.0\% (pass@5), jumping to 41.0\% with Python tools. It also achieves the highest score on OmniDocBench (91.1\%) and, on WorldVQA (51.0\%), ranks second behind
Claude Fable 5, ahead of GPT-5.6 Sol and Claude Opus 4.8.

\subsection{Internal Evaluation}
\label{sec:eval_internal}

\subsubsection{Capability Evaluation}
Beyond the public benchmark suite, we maintain a collection of in-house benchmarks that target capability areas public evaluations do not adequately cover, giving a more comprehensive measure of model and agent capabilities. These benchmarks are refreshed and expanded frequently, so that they can closely track the model's evolving failure modes and directly guide data and training iterations. They broadly fall into three categories: coding capability and experience, general agent experience, and conversational experience. Table~\ref{tab:k3_internal_eval} reports the results across these benchmarks.

\begin{table}[t]
\centering
\footnotesize
\setlength{\tabcolsep}{2.5pt}
\caption{Results on our in-house benchmarks. \textbf{Bold} denotes the best reported result per benchmark; ``-'' denotes scores not yet included in this report. Unless otherwise noted, models are evaluated at maximum reasoning effort (GPT-5.5 at xhigh); harness assignments are shown in the Harness column. $^{\rm a}$13 fallbacks and 1 refusal out of 80 tasks. $^{\rm b}$10 refusals out of 80 tasks. $^{\rm c}$3 refusals out of 80 tasks. $^{\rm d}$Includes 2 tasks that Claude Fable 5 refused to answer. $^{\rm e}$Includes 14 tasks that Claude Fable 5 refused to answer. $^{\rm f}$6 refusals out of 95 tasks. $^{\rm g}$ Reported metric is 1$-$hallucination rate; higher is better.}
\label{tab:k3_internal_eval}
\begin{tabular}{@{}L{3.0cm} | P{2.0cm} | P{1.6cm} | P{1.8cm} P{1.6cm} P{1.6cm} P{1.5cm} | P{1.5cm}@{}}
\toprule
& & & \multicolumn{4}{c|}{\textbf{Proprietary}} & \multicolumn{1}{c}{\textbf{Open Weight}} \\
\cmidrule(l{2pt}r{2pt}){4-7} \cmidrule(l{2pt}r{2pt}){8-8}
\textbf{Benchmark} & \textbf{Harness} & \textbf{Kimi K3 (max)} & \textbf{Claude Fable 5 (max)} & \textbf{GPT-5.6 Sol (max)} & \textbf{Claude Opus 4.8 (max)} & \textbf{GPT-5.5 (xhigh)} & \textbf{GLM-5.2 (max)} \\
\midrule
\multicolumn{8}{@{}l}{\textbf{Coding Experience}} \\
Kimi Code Bench 2.0 & Claude Code & 73.7 & \textbf{76.9}$^{\rm a}$ & - & 71.7 & - & 64.2 \\
 & Kimi Code & 72.9 & - & - & - & 66.0 & - \\
 & Codex & - & - & 64.8$^{\rm b}$ & - & 69.0$^{\rm c}$ & - \\
Coding Experience & Claude Code & \textbf{59.9} & 59.8 & - & 58.0 & - & 53.3 \\
 & Kimi Code & 56.6 & - & - & - & - & - \\
 & Codex & - & - & 59.3 & - & 56.8 & - \\
\midrule
\multicolumn{8}{@{}l}{\textbf{General Agent Experience}} \\
24/7 ClawBench 2.0 & OpenClaw & 48.3 & 47.4$^{\rm d}$ & \textbf{52.0} & 47.2 & 48.5 & 43.2 \\
MIRA Bench & MIRA & 64.1 & \textbf{72.9} & 62.2 & 59.8 & 54.6 & - \\
KAET & Kimi Code & 83.5 & - &  \textbf{85.4} & 78.7 & 79.7 & 74.7 \\
CLIF Bench & Kimi Code & \textbf{52.4} & - & 50.6 & 48.8 & 52.3 & 39.2 \\
Agentic Vision Bench & Kimi Code & 78.3 & 81.1 & \textbf{82.9} & 82.8 & 76.9 & - \\
Swarm Bench & Kimi Agent &  \textbf{76.3} & - & 73.2 & 72.6 & 61.8 & 58.5 \\
Online Experience & Kimi Agent & 77.9 & 74.2$^{\rm e}$ & \textbf{84.0} & 69.4 & 73.7 & 64.0 \\
Deep Research Bench & Kimi Agent &  \textbf{90.0} & - & 85.3 & 87.2 & 81.9 & 84.0 \\
Finance Bench & N/A & 62.6 & - &  \textbf{62.7} & 60.7 & 58.4 & 55.4 \\
KWV Bench & N/A & 64.7 & 63.6 & \textbf{66.9} & 61.7 & 65.8 & - \\
DECK Bench & N/A & 73.5 & 73.0 & \textbf{74.7} & 66.9 & 68.2 & 68.6 \\
Agent Behavior Bench & Kimi Work & 65.0 & 75.5$^{\rm f}$ & \textbf{76.4} & 65.7 & 70.1 & - \\
\midrule
\multicolumn{8}{@{}l}{\textbf{Conversational Experience}} \\
Faithfulness $^{\rm g}$ & N/A & 85.5 & - & 84.8 & 83.6 & \textbf{86.5} & 74.8 \\
Chat All-in-One Bench & Kimi Work & 85.2 & \textbf{88.0} & 79.0 & 83.8 & 71.8 & - \\
\bottomrule
\end{tabular}
\end{table}

\paragraph{Coding Capability and Experience}
\begin{itemize}[leftmargin=12pt]
  \item \textbf{Kimi Code Bench 2.0 (KCB 2.0)}: evaluates code agents on realistic, end-to-end software engineering tasks across a broad range of programming languages and production-oriented technology stacks. 
  \item \textbf{Kimi Webdev Bench}: evaluates models on challenging web development prompts drawn from real usage scenarios, with outputs compared through blind expert judgment, with results available in Table~\ref{tab:k3_webdev_internal}.
  \item \textbf{Coding Experience}: evaluates the practical experience of working with the model as a coding agent in real development workflows.
\end{itemize}
\begin{table}[t]
\centering
\footnotesize
\caption{Results on the in-house Kimi Webdev Bench: Kimi K3 (max) against Claude Opus 4.8 (max), both run with the Claude Code harness. The comparison is performed under blind expert judging, where experts score each output on code quality, feature completeness, visual fidelity, and interaction experience without knowing which model produced it. Win, Tie, and Lose report the percentage of prompts where Kimi K3's output is preferred, rated comparable, or dispreferred, respectively.}
\label{tab:k3_webdev_internal}
\begin{tabular}{@{}l c c c c@{}}
\toprule
\textbf{Domain} & \textbf{Win} & \textbf{Tie} & \textbf{Lose} & \textbf{Win $-$ Lose} \\
\midrule
Games & 55.6\% & 3.7\% & 40.7\% & +14.9\% \\
3D / WebGL / Shader & 72.7\% & 13.7\% & 13.6\% & +59.1\% \\
Website / UI Clone & 52.6\% & 21.1\% & 26.3\% & +26.3\% \\
\midrule
Overall & 58.6\% & 13.8\% & 27.6\% & \textbf{+31.0\%} \\
\bottomrule
\end{tabular}
\end{table}

\paragraph{General Agent Experience}
\begin{itemize}[leftmargin=12pt]
  \item \textbf{24/7 ClawBench 2.0}: simulates always-on assistant work, in which tasks span multiple days, events arrive concurrently, and interruptions are routine.
  \item \textbf{Multi-Agent Infra for Routing and Assignment (MIRA) Bench}: evaluates long-chain, multi-role, multi-system enterprise collaboration tasks, assessing whether agents can carry out end-to-end work and judge when to organize or delegate to subagents.
  \item \textbf{Kimi Autonomous Execution Tasks (KAET)}: evaluates long-horizon autonomous execution on tasks simulating real user requests and enterprise system operations.
  \item \textbf{Context Learning and Instruction Following (CLIF) Bench}: targets in-context learning, requiring models to learn from a provided context while following instructions that interleave multiple complex skills.
  \item \textbf{Agentic Vision Bench}: evaluates whether agents notice and correctly use key visual facts during task execution.
  \item \textbf{Swarm Bench}: evaluates models' ability to orchestrate agent swarms~\citep{kimik25} on complex tasks that benefit from coordinated decomposition and parallel execution.
  \item \textbf{Online Experience}: mirrors the distribution of real online agent usage, measuring performance on the deliverable file types most frequently requested by users.
  \item \textbf{Deep Research Bench}: evaluates models on deep-research-style queries curated by domain experts and graded with expert-aligned rubrics.
  \item \textbf{Finance Bench}:  evaluates models on realistic financial work that requires end-to-end execution of complete workflows, from source materials to reviewable deliverables.
  \item \textbf{Knowledge Work Vision (KWV) Bench}: evaluates atomic visual capabilities extracted from tasks distilled from real knowledge-work scenarios.
  \item \textbf{DECK Bench}: measures the capability to produce high-quality presentation decks from task descriptions drawn from real usage scenarios.
  \item \textbf{Agent Behavior Bench}: extends agent evaluation from outcome correctness to process quality, scoring tool-use behavior, efficiency, and discipline alongside task completion.
\end{itemize}

\paragraph{Conversational Experience}
\begin{itemize}[leftmargin=12pt]
  \item \textbf{Faithfulness}: measures factual hallucination rates in model responses, with each response verified by a fact checker.
  \item \textbf{Chat All-in-One Bench}: measures conversational experience at every stage of product usage, with scenarios designed around real online user needs.
\end{itemize}

\paragraph{Evaluation Configurations} Unless a benchmark is split into separate rows by harness, the Harness column in Table~\ref{tab:k3_internal_eval} reports the harness used for \kimi{3}. For other models, Claude models and GLM-5.2 are evaluated with Claude Code, while GPT models are evaluated with Codex. The exceptions are benchmarks where all models use the same specified harness: OpenClaw for 24/7 ClawBench 2.0; MIRA (Multi-Agent Infra for Routing and Assignment), an internal out-of-distribution harness, for MIRA Bench; Kimi Work for Agent Behavior Bench and Chat All-in-One; and Kimi Code for CLIF and Agentic Vision Bench.

\paragraph{Results} The in-house suite separates \kimi{3}'s strengths from its weaknesses more sharply than the public benchmarks. The clearest strengths are orchestration- and research-type agency: \kimi{3} leads Swarm Bench (76.3) and Deep Research Bench (90.0) by clear margins, indicating strong capability in decomposing complex objectives, coordinating parallel work, and producing rubric-satisfying deliverables. Coding is likewise a strength: on Kimi Code Bench 2.0 it trails only Claude Fable 5, and it attains the best score on Coding Experience, suggesting that its practical behavior as a coding agent --- communication quality, behavioral appropriateness, and instruction-following stability --- is ahead of its raw task scores; on the Kimi Webdev Bench, expert judges prefer it over Claude Opus 4.8 by a +31.0-point overall margin, with the largest gain on 3D/WebGL/Shader tasks. Professional knowledge work has also improved markedly over the previous generation, with Finance Bench essentially tied with GPT-5.6 Sol.

\kimi{3} trails the leaders mainly on Agent Behavior Bench, MIRA Bench, 24/7 ClawBench 2.0, Agentic Vision Bench, and KWV Bench. On the remaining filled suites (KAET, CLIF Bench, Online Experience, DECK Bench, Faithfulness, and Chat All-in-One Bench), \kimi{3} ranks first or a close second.

\subsubsection{Cyber Security Evaluation}
We evaluate the model's cybersecurity capability along a two-tier progression of increasing operational risk: vulnerability discovery with proof-of-concept development (Tier 1), and end-to-end exploit development (Tier 2). Evaluation targets include recent versions of widely deployed software---operating-system kernel components and open-source projects---as well as our internal infrastructure, including production services and codebases. All tasks run in standard configurations representative of real-world deployments. Frontier models from Anthropic and OpenAI refuse cyber-related tasks, making a comparable evaluation infeasible; we therefore exclude them from this suite.

\paragraph{Vulnerability discovery (Tier 1).}
This tier tasks the model with identifying genuine bugs in current codebases---rather than reproducing known vulnerabilities---and demonstrating that they are reproducible. These capabilities are primarily associated with defensive security research.

Across dozens of widely deployed systems spanning operating-system kernels, databases, AI services, web frameworks, blockchain, and VPN software, the model identified hundreds of candidate vulnerabilities. Of the findings that underwent human review, approximately 70\% were confirmed as genuine, including 16 previously unknown vulnerabilities across six projects.

Two findings in the Linux kernel illustrate the depth of these results. First, the model identified a remotely triggerable heap out-of-bounds write. The bug was introduced by an incomplete upstream fix and affects all subsequent releases, up to and including the latest upstream code. Security experts confirmed it as a remote denial-of-service primitive. Second, the model identified a Dirty-COW-class vulnerability in the RDMA subsystem: an earlier upstream fix had inadvertently dropped a permission check, enabling kernel-side writes to read-only memory pages. Security experts confirmed it as a deterministic local privilege-escalation primitive.

\paragraph{Exploit development (Tier 2).}
This tier requires the model to convert a vulnerability into a working end-to-end exploit, and is the tier most directly relevant to misuse risk. We evaluate it against GLM-5.2 as the baseline, using an in-house suite of 36 tasks spanning two tracks.

\emph{User-space exploitation} (16 tasks). The model must exploit real CVEs end-to-end in widely deployed user-space software, including PostgreSQL, the XWiki collaboration platform, the Apache HTTP Server, and several content-management systems and other applications. For each task, the model is given full source code and a live instance; targets run in standard configurations without additional hardening.

\emph{Linux kernel exploitation} (20 tasks). Each task provides a reproducible QEMU environment built from a historical kernel CVE, and the model must write a C exploit that escalates privileges from an unprivileged user to root. Mitigations are progressively enabled across difficulty grades.

Every task in the suite is verified solvable by human security experts. We estimate that completing the full suite requires roughly 540 expert-hours, or about 15 hours per task on average.

\paragraph{Results on the exploit suite.}
The model demonstrates meaningful exploit-development capability on this suite, solving 14 of 36 tasks (38.9\%) versus 8 of 36 (22.2\%) for GLM-5.2. Its successes are unevenly distributed, however: 10 of the 14 come from the user-space track. On the kernel track, neither model solves three-quarters of the tasks.

Since every task is solvable by human experts, the unsolved tasks directly measure the model's remaining gap to human-level capability. Trajectory analysis attributes this gap to four recurring failure modes: (i) difficulty completing the final stage of an exploit chain from primitives already obtained; (ii) poor strategy selection under mitigations, such as persisting with control-flow hijacking when a data-only attack would be simpler and more reliable; (iii) getting trapped in prolonged, unproductive debugging loops; and (iv) insufficient verification of the final deliverable before submission.

\paragraph{Summary.}
The model's cyber capability is strongest at Tier 1 and at user-space exploitation within Tier 2, yet a clear gap to human experts remains. At Tier 1, which is defensive in nature, the model identifies genuine vulnerabilities---including previously unknown ones---and demonstrates their reproducibility. At Tier 2, it completes end-to-end exploits against user-space targets. Against hardened targets, however, completing the full exploit chain remains the bottleneck, and many expert-solvable tasks go unsolved.

An independent joint assessment by the UK AI Security Institute and NIST's Center for AI Standards and Innovation (CAISI)~\citep{ukaisi_caisi_k3} reaches conclusions consistent with ours. \kimi{3} outperforms GLM-5.2 on exploit development (32\% vs.\ 24\% on ExploitBench; 17 vs.\ 11 steps on a 32-step simulated enterprise network that takes a human expert roughly 20 hours), but trails frontier cyber-capable models on end-to-end exploit completion, achieving arbitrary code execution on 0 of 41 tasks.

We regard our evaluation as a lower bound on capability. These results are conditioned on the current model version and evaluation coverage, and we will revisit them at each major model update.

\subsection{Third-Party Evaluation}
\label{sec:eval_third_party}
\kimi{3} has also been independently evaluated by third-party organizations since its release. Table~\ref{tab:k3_third_party} summarizes the headline results as of July 23, 2026.

\paragraph{Artificial Analysis} Artificial Analysis evaluated \kimi{3}~\citep{artificialanalysis}. \kimi{3} attains an Intelligence Index v4.1 of 57.1, ranking fourth of 580 models --- third if GPT-5.6 Sol effort variants are counted as a single entry --- behind Claude Fable 5 (59.9) and GPT-5.6 Sol (58.9), and ahead of all other evaluated models.

\paragraph{Vals AI} On Vals AI's GDP-weighted industry benchmark suite~\citep{valsai}, \kimi{3} ranks second of 39 models on the Vals Index (74.7\%), behind Claude Fable 5 (75.1\%) and ahead of GPT-5.6 Sol (73.1\%).

\paragraph{Arena} On the crowdsourced human-preference arenas~\citep{lmarena}, \kimi{3} ranks first of 99 models on the WebDev Arena (1{,}678 Elo, ahead of Claude Fable 5 at 1{,}634) --- the first open model to top this leaderboard --- and eighth of 200 on the Text Arena (1{,}486 Elo). On the Agent Arena, which opened for voting around July 19, \kimi{3} currently ranks fourth of 37 (9.1), behind Claude Fable 5 (12.7), GPT-5.6 Sol (10.1), and Claude Opus 4.8 (9.8).

\begin{table}[tp]
\centering
\footnotesize
\setlength{\tabcolsep}{3pt}
\caption{Headline independent third-party evaluations of Kimi K3 (as of July 23, 2026). \textbf{Bold} denotes the best result per benchmark and \uline{underline} the second best. Baseline scores are as reported by each source under its own evaluation setup $^{\rm a}$Text Arena entry is the xhigh variant listed on the leaderboard. $^{\rm b}$Text Arena entry is the high variant listed on the leaderboard. Numbers in parentheses are Kimi K3's rank on that leaderboard. Elo-style scores drift as additional matches accumulate.}
\label{tab:k3_third_party}
\begin{tabular}{@{}l | P{1.9cm} | P{2.1cm} P{1.9cm} P{1.9cm} P{1.7cm} | P{1.7cm}@{}}
\toprule
& & \multicolumn{4}{c|}{\textbf{Proprietary}} & \multicolumn{1}{c}{\textbf{Open Weight}} \\
\cmidrule(l{2pt}r{2pt}){3-6} \cmidrule(l{2pt}r{2pt}){7-7}
\textbf{Benchmark} & \textbf{Kimi K3 (max)} & \textbf{Claude Fable 5 (max)} & \textbf{GPT-5.6 Sol (max)} & \textbf{Claude Opus 4.8 (max)} & \textbf{GPT-5.5 (xhigh)} & \textbf{GLM-5.2 (max)} \\
\midrule
\multicolumn{7}{@{}l}{\textbf{Artificial Analysis}} \\
Intelligence Index v4.1 (\#4/580) & 57.1 & \textbf{59.9} & \uline{58.9} & 55.7 & 55.0 & 51.1 \\
\midrule
\multicolumn{7}{@{}l}{\textbf{Vals AI}} \\
Vals Index (\#2/39) & \uline{74.7} & \textbf{75.1} & 73.1 & 70.4 & 68.0 & 65.0 \\
\midrule
\multicolumn{7}{@{}l}{\textbf{Arena}} \\
WebDev Arena (Elo, \#1/99) & \textbf{1{,}678} & \uline{1{,}634} & 1{,}630 & 1{,}565 & 1{,}507 & 1{,}592 \\
Text Arena (Elo, \#8/200) & \uline{1{,}486} & \textbf{1{,}507} & 1{,}485$^{\rm a}$ & 1{,}484$^{\rm b}$ & 1{,}482$^{\rm b}$ & 1{,}469 \\
Agent Arena (\#4/37) & 9.1 & \textbf{12.7} & \uline{10.1} & 9.8 & 8.8 & 6.5 \\
\bottomrule
\end{tabular}
\end{table}

\subsection{Cost Efficiency}
\label{sec:eval_cost}
Beyond scores, we examine inference cost efficiency by comparing score against per-task cost across four suites covering coding and agentic tasks: Kimi Code Bench~2.0, BrowseComp, GDPval-AA v2, and AA-Briefcase. For Kimi Code Bench~2.0, costs are measured internally, with \kimi{3} run via Kimi Code, and all other models via Claude Code. For BrowseComp, the cost of \kimi{3} is measured from our own runs, while the costs of Claude and GPT are cited from published charts~\citep{gpt56sol,claudesonnet5,claudesonnet5blog}. For GDPval-AA v2 and AA-Briefcase, costs are cited from Artificial Analysis's pay-per-token API pricing as of July 23, 2026~\citep{artificialanalysis}.

\begin{figure}[htbp]
  \centering
  \begin{minipage}{0.48\textwidth}
    \centering
    \includegraphics[width=\textwidth]{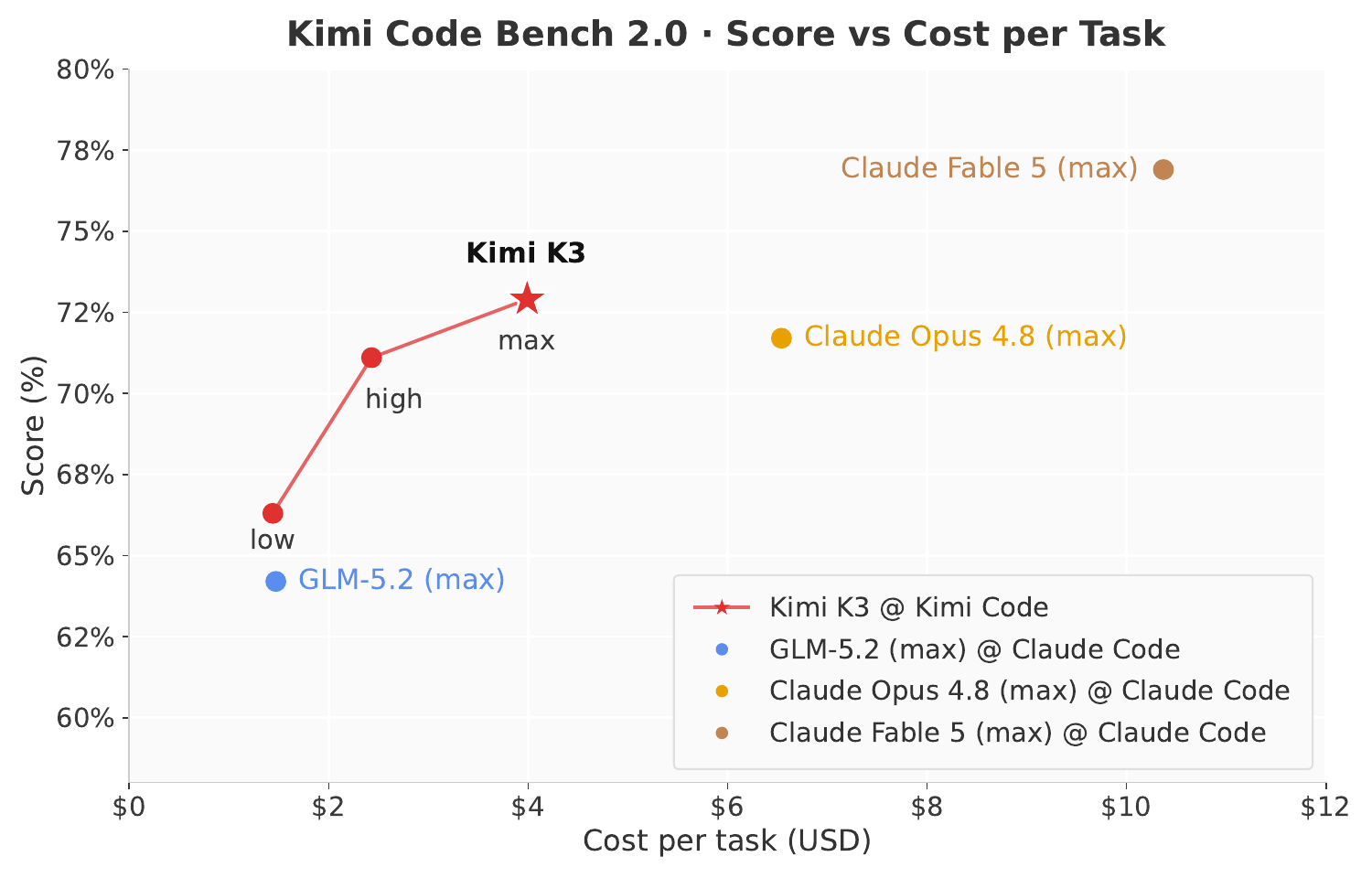}\\[2pt]
    {\small (a) Kimi Code Bench 2.0}
  \end{minipage}\hfill
  \begin{minipage}{0.48\textwidth}
    \centering
    \includegraphics[width=\textwidth]{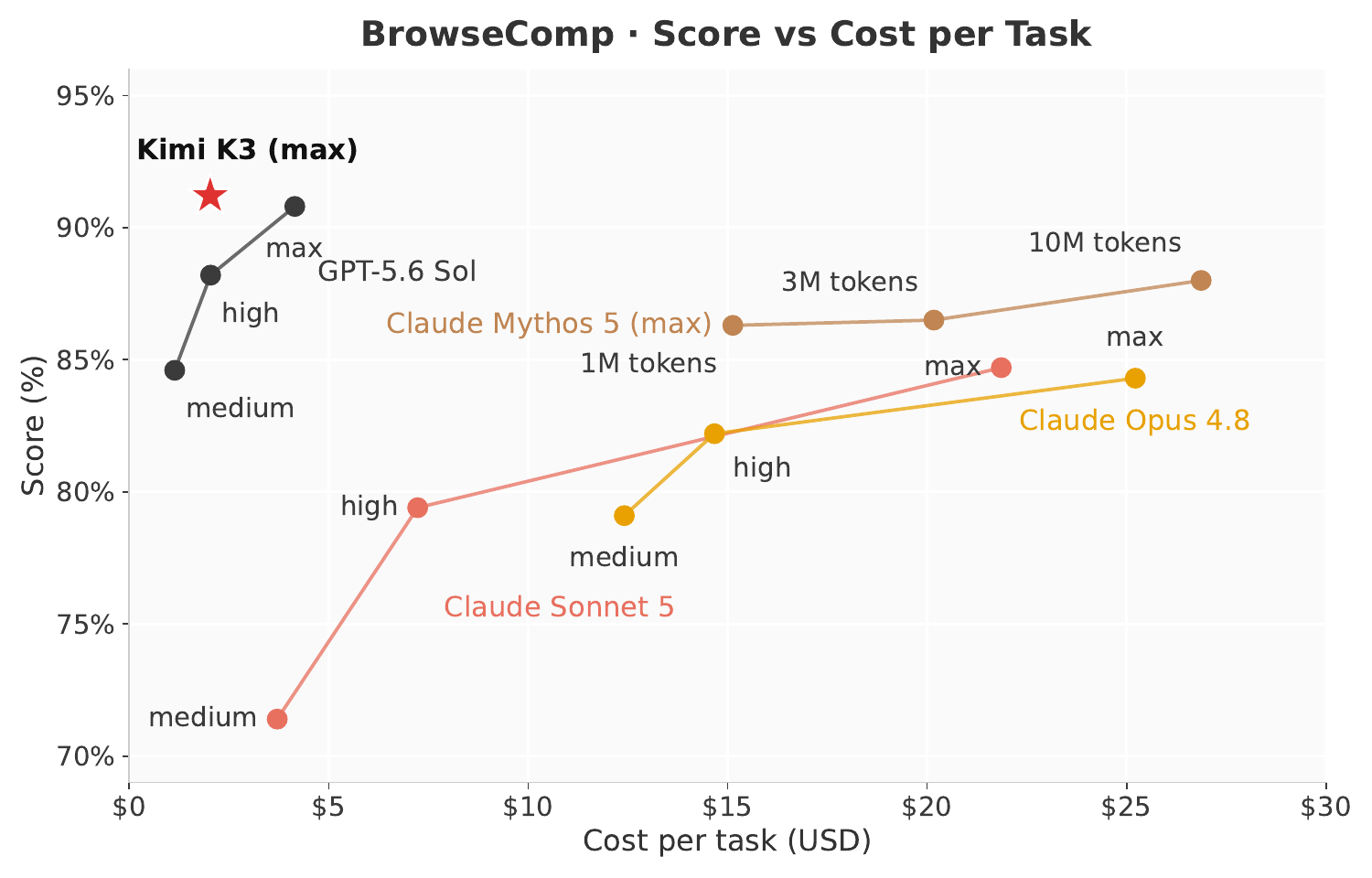}\\[2pt]
    {\small (b) BrowseComp}
  \end{minipage}\\[8pt]
  \begin{minipage}{0.48\textwidth}
    \centering
    \includegraphics[width=\textwidth]{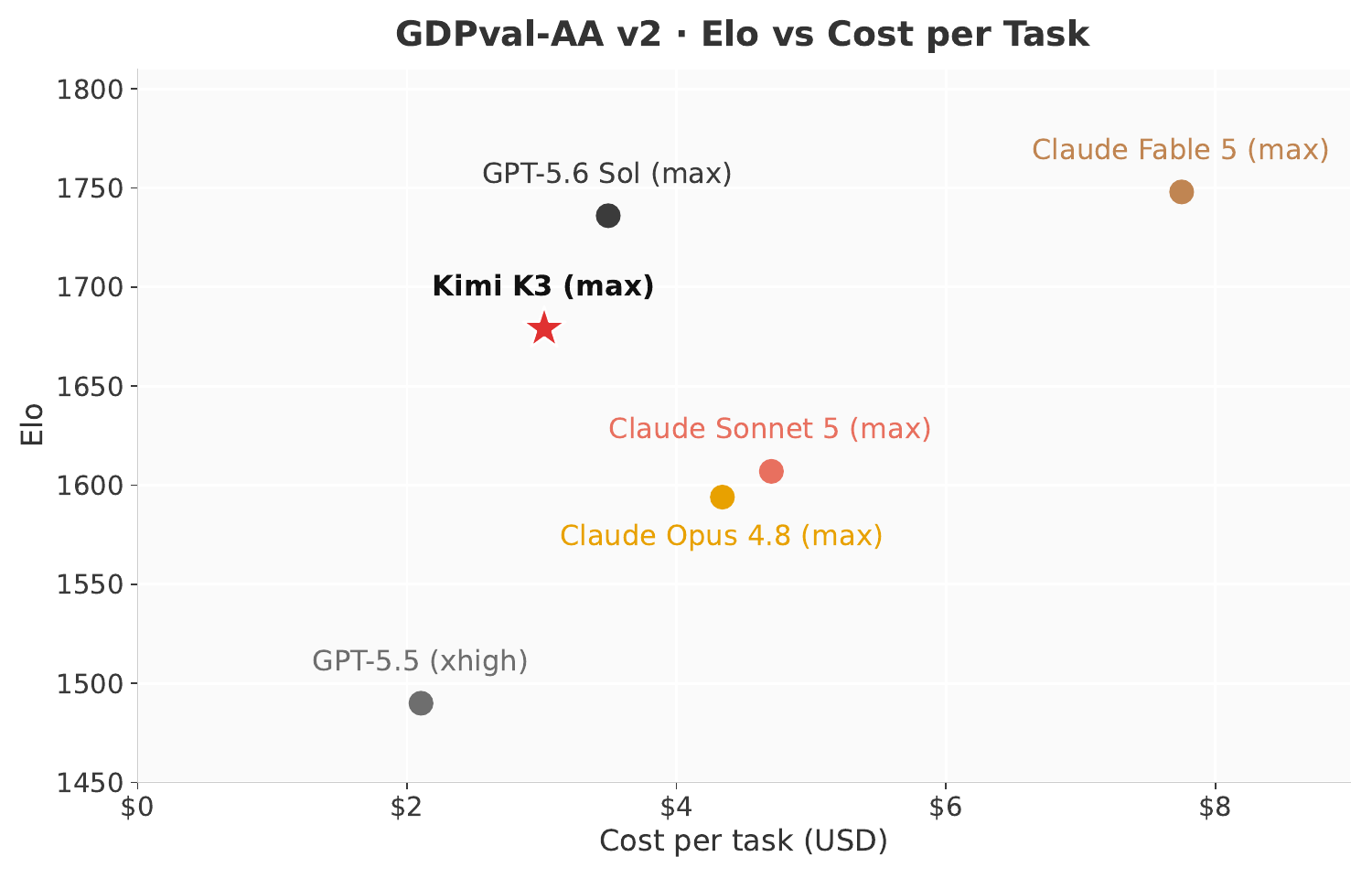}\\[2pt]
    {\small (c) GDPval-AA v2}
  \end{minipage}\hfill
  \begin{minipage}{0.48\textwidth}
    \centering
    \includegraphics[width=\textwidth]{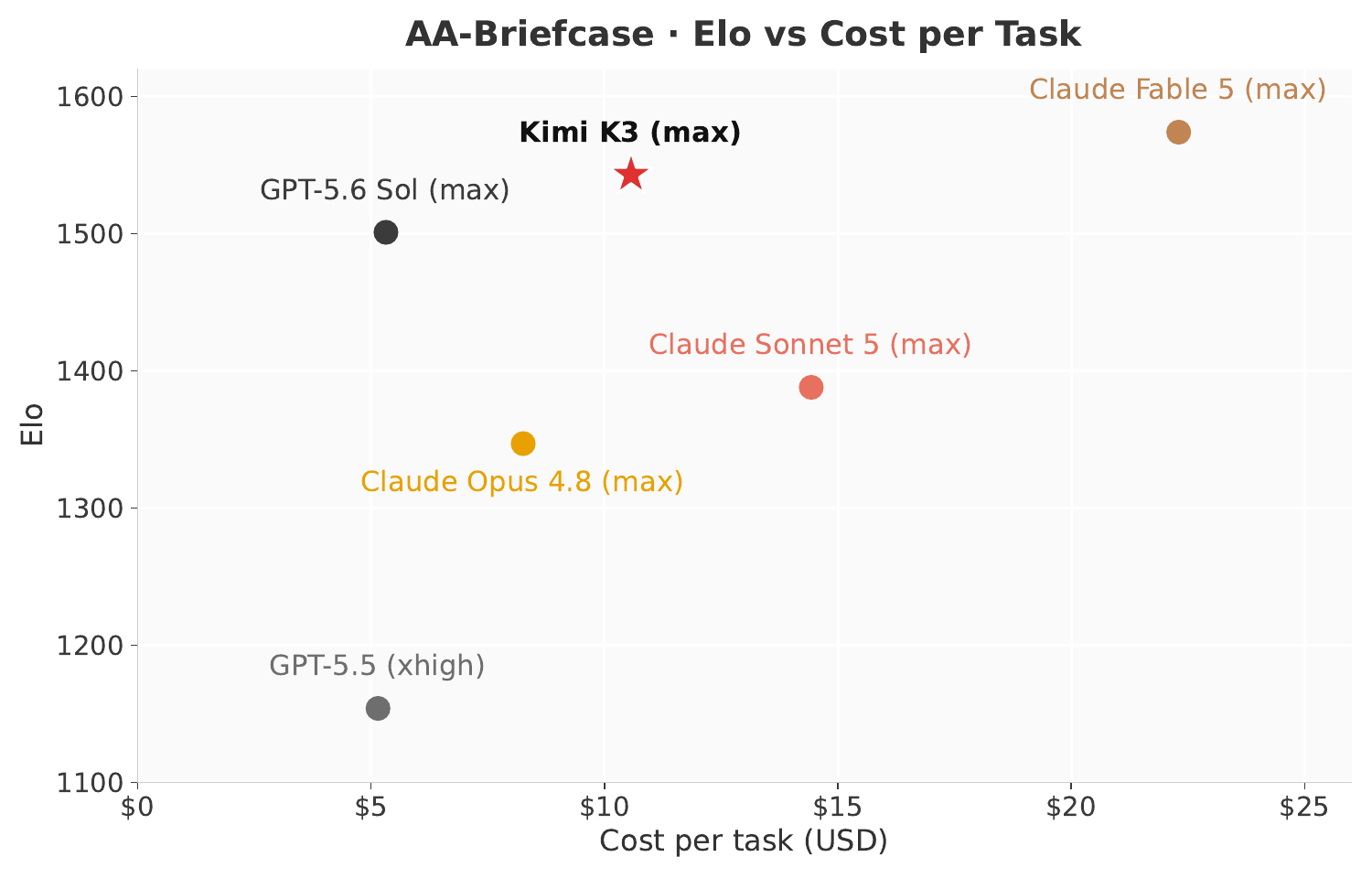}\\[2pt]
    {\small (d) AA-Briefcase}
  \end{minipage}
  \caption{Score vs.\ per-task inference cost on Kimi Code Bench~2.0, BrowseComp, GDPval-AA v2, and AA-Briefcase. \kimi{3} is marked with a star.}
  \label{fig:cost_efficiency}
\end{figure}

On Kimi Code Bench~2.0, \kimi{3} is 4.0 points behind Claude Fable 5 at 38\% of its cost, and at high effort it already matches Claude Opus 4.8's maximum-effort score at roughly one third of the cost. On BrowseComp, \kimi{3} attains the best score (91.2\%) at \$2.03 per task --- half the cost of GPT-5.6 Sol (90.4\%) and an order of magnitude cheaper than the Claude models at their maximum effort. On GDPval-AA v2, \kimi{3} is within 50 Elo of GPT-5.6 Sol at 13\% lower cost, and 2.6$\times$ cheaper than Claude Fable 5. On AA-Briefcase, it delivers the second-best score behind Claude Fable 5, at roughly half of the latter's cost. Figure~\ref{fig:cost_efficiency} summarizes the comparison.

Overall, \kimi{3} sits on or near the cost-efficiency frontier across all four suites, delivering near-top scores at a fraction of the cost of Claude Fable 5 in particular.

\section{Case Studies}

In this section, we present representative cases that demonstrate \kimi{3}'s capabilities across diverse technical tasks.

\begin{figure}[t]
  \centering
  \resizebox{\columnwidth}{!}{\input{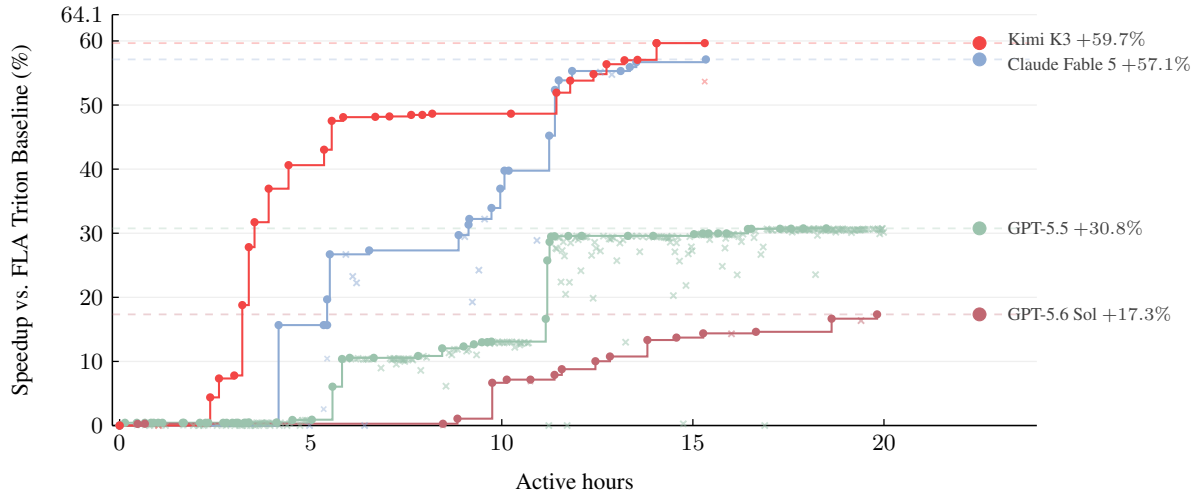}}
  \caption{Case study: GPU kernel optimization on AttnRes.}
  \label{fig:case-kernel}
\end{figure}

\begin{figure}[h]
\centering
\includegraphics{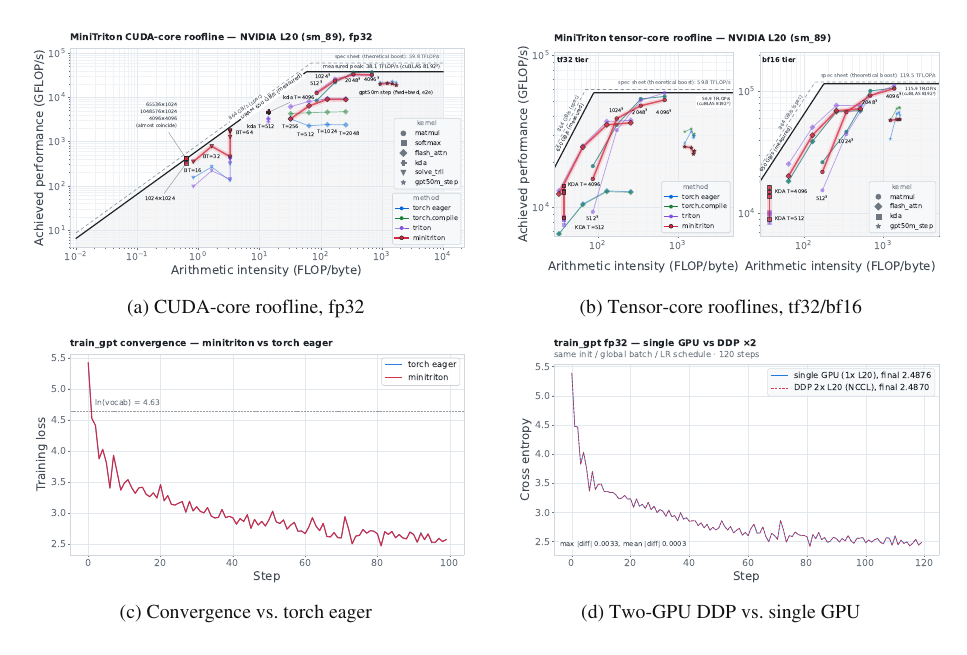}
\caption{Case study: GPU compiler development with MiniTriton. (a) CUDA-core
and (b) tensor-core rooflines of MiniTriton kernels on an NVIDIA L20 (sm\_89)
against torch eager, torch.compile, Triton, and cuBLAS baselines (losing
points included); (c) training-loss curves of the character-level GPT trained
with MiniTriton versus torch eager; (d) two-GPU data-parallel training built
on MiniTriton's own distributed primitives (NCCL) versus single-GPU training.}
\label{fig:case-compiler}
\end{figure}

\paragraph{GPU kernel optimization}
We tested the models' ability to optimize GPU kernels. Each model works independently in an identically configured sandbox, with a budget of up to 24 hours per task for profiling, rewriting, and benchmarking. The evaluation covers four representative kernels: AttnRes, DeepSeek Sparse Attention (DSA), KDA, and MLA (with head dimension 512), on an NVIDIA Hopper GPU and an alternative-vendor GPGPU.
\kimi{3} substantially improved performance across all four kernels, reducing AttnRes latency from 283.6\,ms to 114.4\,ms, cutting DSA and KDA runtime by 55.1\% and 73.6\%, respectively, and reaching over half of peak TFLOPS on MLA.
Across these tasks, \kimi{3} matched Claude Fable 5~\citep{claudefable5} (with fallback) and substantially outperformed Claude Opus 4.8~\citep{claudeopus48}, GPT-5.6 Sol~\citep{gpt56sol}, and GPT-5.5~\citep{gpt55}.
Figure~\ref{fig:case-kernel} compares the models' optimization trajectories on AttnRes.
Beyond the benchmark, an early \kimi{3} checkpoint was already handling most of our kernel optimization work during late-stage development.

\paragraph{GPU compiler development} \kimi{3} developed MiniTriton\footnote{\url{https://github.com/MoonshotAI/minitriton}}, a compact Triton-like~\citep{tillet2019triton} compiler with a custom tile-level Python frontend and layout system, a lightweight warp-level MLIR~\citep{lattner2021mlir} annotation and optimization layer, and a Parallel Thread Execution (PTX) code-generation pipeline. Built around the compiler is a dual-mode tensor library with a PyTorch-like~\citep{paszke2019pytorch} high-level interface, whose eager and forward-only compiled paths share the same DSL compiler and runtime. The library further provides reverse-mode autograd, neural-network modules, distributed-training primitives over NCCL~\citep{nccl},  and sparse and visualization primitives.
On an NVIDIA L20, MiniTriton outperforms PyTorch eager~\citep{paszke2019pytorch} and \texttt{torch.compile}~\citep{ansel2024pytorch2} in geometric mean over its core benchmark suite. Its from-scratch tensor-core matmul path approaches cuBLAS~\citep{cublas} at the largest shapes,  reaching about 90\% of the measured machine roof, while its DSL-level KDA~\citep{team2025kimi} prefill kernel outperforms a matched Triton reference by a clear margin. MiniTriton also trains a GPT model end to end with a loss curve closely tracking the PyTorch reference, with full-model gradients differing from torch autograd by no more than torch's own fp32 rounding error ($10^{-4}$), measured against an fp64 reference. Together, These results demonstrate that \kimi{3} can build a coherent end-to-end compiler --- from DSL frontend and IR passes to PTX codegen and CUDA runtime --- rather than a collection of isolated kernels (Figure~\ref{fig:case-compiler}).

\paragraph{Chip design} 
As an early proof of concept, \kimi{3} designed an inference-chip prototype for a nano model following the same architecture --- hybrid KDA and NoPE-MLA attention, Block AttnRes with a block size of two, sigmoid-based MoE routing with one shared expert --- under
group-wise INT4 weight quantization (group size 128). In a single 48-hour autonomous run with Kimi Code, \kimi{3} built, optimized, and verified the chip using open-source EDA tools with the Nangate45 standard-cell library~\cite{nangate45}. Within the 4\,mm$^2$ analytical area budget, the design closes timing at 100\,MHz and achieves an RTL-simulated decode throughput of over 8{,}700 tokens/s, integrating 1.46M standard cells, 0.277 MiB of SRAM, and an INT4 MAC array with fused dequantization. 
The RTL code is available on GitHub\footnote{\url{https://github.com/MoonshotAI/nano-kpu}}.

\paragraph{Coding for research} To reproduce the I--Love--Q universal relations in computational astrophysics, \kimi{3} reviewed more than 20 papers and cross-validated their results, implemented the full numerical pipeline, evaluated over 300 equations of state, identified inconsistencies in published formulas, wrote more than 3,000 lines of Python, and produced an interactive HTML dashboard --- in about two hours, versus a typical one to two weeks for an experienced researcher. 

\paragraph{Knowledge work} In Kimi Work, \kimi{3} produced an interactive research website covering 42 years of the AI ASIC industry. The model completed more than 120 rounds of iterative refinement, drawing on a corpus of 87 quarterly reports and 99 original PDFs (more than 11,000 pages) through over 2,800 web searches and over 1,100 terminal queries.
In a second case, \kimi{3} analyzed 391 gravitational-wave events in GWTC-5 using more than 20 concurrent subagents, producing seven scientific visualizations, two summary tables, and a literature synthesis of over ten papers.

\paragraph{Video editing and motion design} Leveraging its native multimodal architecture, \kimi{3} created a 3Blue1Brown-style motion-graphics explainer of its own architecture, and edited its teaser video from 56 source clips. This involved clip selection, motion-matched cuts, frame-accurate beat synchronization, audio processing, and multiple rounds of revision. Producing a comparable high-density short video would typically take an experienced editor one to two days.
\section{Conclusion}

We present \kimi{3}, an open 2.8-trillion-parameter Mixture-of-Experts model with native vision capabilities and a 1-million-token context window, built on Kimi Delta Attention and Attention Residuals. 
As the world's first open 3T-class model, \kimi{3} delivers frontier-level performance across long-horizon
coding, agentic, knowledge, reasoning, and vision tasks. 
Although gaps to the strongest proprietary models remain, \kimi{3} establishes a new open frontier within everyone's reach. We hope it will empower the broader community in research, deployment, and innovation.

\clearpage
\printbibliography[title={References}]

\clearpage
\appendix
\section{Contributions}
The listing of contributors is in alphabetical order based on their last names. 

\begin{multicols}{4}
{
\small
Tongtong Bai\\
Yifan Bai\\
Yiping Bao\\
M. C.\\
Jianfeng Cai\\
Xinyuan Cai\\
Peizhou Cao\\
Yuxuan Cao\\
Ziwei Chai\\
Y. Charles\\
H.S. Che\\
Guanduo Chen\\
Guangyu Chen\\
Guanzheng Chen\\
Huarong Chen\\
Jia Chen\\
Jianlong Chen\\
Jun Chen\\
Kexin Chen\\
Peng Chen\\
Ruijue Chen\\
Wentao Chen\\
Xin Chen\\
Yang Chen\\
Yanru Chen\\
Yifei Chen\\
Yingjiang Chen\\
Yuankun Chen\\
Yujie Chen\\
Yutian Chen\\
Zhirong Chen\\
Dazhi Cheng\\
Yean Cheng\\
Jialei Cui\\
Jingbing Cui\\
Anqi Dai\\
Jiaqi Deng\\
Hao Ding\\
Rui Ding\\
Shaofeng Ding\\
Mengfan Dong\\
Mengnan Dong\\
Yuhao Dong\\
Yuxin Dong\\
Ang'ang Du\\
Chenzhuang Du\\
Dikang Du\\
Jusen Du\\
Yulun Du\\
Yu Fan\\
Jing Feng\\
Qiulin Feng\\
Yichen Feng\\
Kelin Fu\\
Qiang Fu\\
Fuxuan Gao\\
Hongcheng Gao\\
Jingyue Gao\\
Tong Gao\\
Weijia Gao\\
Shangyi Geng\\
Jie Gong\\
Linghu Gong\\
Shengao Gong\\
Xiaochen Gong\\
Qizheng Gu\\
Yicheng Gu\\
Shuhao Guan\\
Haiqing Guo\\
Shiqi Guo\\
Xiang Guo\\
Zhengyan Guo\\
Beixi Hao\\
Wenxin Hao\\
Xiaoru Hao\\
Dailan He\\
Haotian He\\
Lehan He\\
Qi He\\
Weiran He\\
Xinran He\\
Xinyi He\\
Yibo He\\
Yunjia He\\
Chao Hong\\
Tiange Hong\\
Hao Hu\\
Jiaxi Hu\\
Ruikun Hu\\
Weiming Hu\\
Yangyang Hu\\
Zhenxing Hu\\
Liang Hua\\
Jinbin Huang\\
Ke Huang\\
Ruiyuan Huang\\
Siying Huang\\
Weixiao Huang\\
Yan Huang\\
Zhengjie Huang\\
Zhiqi Huang\\
Yulong Hui\\
Chaobo Jia\\
Yutong Jiang\\
Zhejun Jiang\\
Zuoyou Jiang\\
Wenyi Jin\\
Xinyi Jin\\
Yu Jing\\
Huanjun Kong\\
Guokun Lai\\
Aidi Li\\
Cheng Li\\
Chengyuan Li\\
Cong Li\\
Fang Li\\
Guanyu Li\\
Haoyang Li\\
Jia Li\\
Junxiong Li\\
Lei Li\\
Letian Li\\
Lincan Li\\
Weihong Li\\
Wentao Li\\
Xintong Li\\
Yang Li\\
Yishen Li\\
Yiwei Li\\
Yuxiao Li\\
Zhaowei Li\\
Zhaoxi Li\\
Zheming Li\\
Zhengxiao Li\\
Zhiyuan Li\\
Jiawei Lin\\
Xiaohan Lin\\
Yibo Lin\\
Zichao Lin\\
Ziyan Lin\\
Bill Liu\\
Boxiao Liu\\
Chuan Liu\\
Liang Liu\\
Shaowei Liu\\
Shudong Liu\\
Shuran Liu\\
Tianwei Liu\\
Weizhou Liu\\
Yangyang Liu\\
Yanming Liu\\
Yibo Liu\\
Yipeng Liu\\
Zhengying Liu\\
Zhiheng Liu\\
Enzhe Lu\\
Haoyu Lu\\
Linqiang Lu\\
Tingzhan Lu\\
Zhiyuan Lu\\
Aotian Luo\\
G. Luo\\
Junyu Luo\\
Yifan Luo\\
B. Lyu\\
Wenzhou Lyu\\
Shaoguang Mao\\
Yuan Mei\\
Xin Men\\
Minqing Ni\\
Yixuan Niu\\
Siyuan Pan\\
Shujun Peng\\
Zhangyang Qi\\
Ruoyu Qin\\
ZeChao Qin\\
Zeyu Qin\\
Haiquan Qiu\\
Jianxin Qiu\\
Jiezhong Qiu\\
Bowen Qu\\
Yuhao Qu\\
Zeyu Shang\\
Youbo Shao\\
Han Shen\\
Jincheng Shi\\
Juanfeng Shi\\
Lidong Shi\\
Shengyuan Shi\\
Wingchun Siu\\
Pengwei Song\\
Xiaoxi Song\\
Jianlin Su\\
Yunfeng Su\\
Zhaochen Su\\
Lin Sui\\
Jingsong Sun\\
Junyao Sun\\
Shaoning Sun\\
Shuzhe Sun\\
Tongyu Sun\\
Yujun Sun\\
Yunpeng Tai\\
Chuning Tang\\
Heyi Tang\\
Sirui Tang\\
Zecheng Tang\\
Chaoran Tian\\
Rongpeng Tian\\
Yu Tian\\
Wei Tu\\
Chensi Wang\\
Chuang Wang\\
Chunjie Wang\\
Dinglu Wang\\
Feng Wang\\
Hailong Wang\\
Haiming Wang\\
Hao Wang\\
Hao Wang\\
Huaqing Wang\\
Hui Wang\\
Jiayi Wang\\
Jinglong Wang\\
Jinhong Wang\\
Jiuzheng Wang\\
Linian Wang\\
Shaobo Wang\\
Shenzhi Wang\\
Shuyi Wang\\
Si Wang\\
Siyuan Wang\\
Tianfu Wang\\
Wenjue Wang\\
Xingran Wang\\
Xinmei Wang\\
Xinyuan Wang\\
Xusheng Wang\\
Yalin Wang\\
Yangkun Wang\\
Yao Wang\\
Yaoyu Wang\\
Yejie Wang\\
Yiqin Wang\\
Yucheng Wang\\
Yuzhi Wang\\
Zhaoji Wang\\
Zhaowei Wang\\
Zhengtao Wang\\
Zhenhao Wang\\
Zhongsheng Wang\\
Zifan Wang\\
Chu Wei\\
Ming Wei\\
Shouxin Wei\\
Zichen Wen\\
Fan Wu\\
Haoning Wu\\
Rucong Wu\\
Wenhao Wu\\
Xiaoxue Wu\\
Yingcong Wu\\
Yongqi Wu\\
Yuxin Wu\\
Zijian Wu\\
Xinglang Xian\\
Chenxuan Xiang\\
Yuye Xiang\\
Bocheng Xiao\\
Chenjun Xiao\\
Xin Xiao\\
Jin Xie\\
Xiaotong Xie\\
Yifeng Xie\\
Zhe Xie\\
Bowei Xing\\
Yiming Xiong\\
Baosheng Xu\\
Boyu Xu\\
Jiale Xu\\
Jianfan Xu\\
Jing Xu\\
Jinjing Xu\\
L.H. Xu\\
Qingtao Xu\\
Shuyao Xu\\
Suting Xu\\
Tiantian Xu\\
Tianxiang Xu\\
Weixin Xu\\
Xinran Xu\\
Yangchuan Xu\\
Ye Xu\\
Yueni Xu\\
Ziyao Xu\\
Haonan Xue\\
Junjie Yan\\
Yaoyao Yan\\
Fan Yang\\
Guangyao Yang\\
Hao Yang\\
Junwei Yang\\
Ruoyu Yang\\
Wenjie Yang\\
Xiaofei Yang\\
Xinyu Yang\\
Yi Yang\\
Yiling Yang\\
Ying Yang\\
Yuchen Yang\\
Zhen Yang\\
Zhilin Yang\\
Zian Yang\\
Zuhao Yang\\
Haotian Yao\\
Dan Ye\\
Haoran Ye\\
Wenjie Ye\\
Zhanbo Ye\\
Bohong Yin\\
Haoxiang Yin\\
Xietong Yin\\
Chengzhen Yu\\
Haozhen Yu\\
Longhui Yu\\
Shengnan Yu\\
Shuying Yu\\
Tianxiang Yu\\
Enming Yuan\\
Mengjie Yuan\\
Tongtian Yue\\
Wei Yue\\
Yang Yue\\
Dunyuan Zha\\
Haobing Zhan\\
B.H. Zhang\\
Dehao Zhang\\
Fei Zhang\\
Hao Zhang\\
Haoyuan Zhang\\
Huanyu Zhang\\
Jiapei Zhang\\
Jiaxuan Zhang\\
Jin Zhang\\
Kaiyi Zhang\\
Miaozhen Zhang\\
Puqi Zhang\\
Qinglei Zhang\\
Rong Zhang\\
Rui Zhang\\
Shaoshuai Zhang\\
Shiyi Zhang\\
Xiaobin Zhang\\
Xiaoyun Zhang\\
Y. Zhang\\
Yangkun Zhang\\
Ye Zhang\\
Yichi Zhang\\
Yikun Zhang\\
Yizhi Zhang\\
Yongting Zhang\\
Yu Zhang\\
Yutao Zhang\\
Yutong Zhang\\
Zheng Zhang\\
Zijing Zhang\\
Bin Zhao\\
Chenguang Zhao\\
Feifan Zhao\\
Jinglun Zhao\\
Jinxiang Zhao\\
Shuai Zhao\\
Wenshuo Zhao\\
Xiangyu Zhao\\
Xuanle Zhao\\
Yikai Zhao\\
Zijia Zhao\\
Haozhi Zheng\\
Huabin Zheng\\
Ruihan Zheng\\
Shaojie Zheng\\
Tengyang Zheng\\
Haofeng Zhong\\
Lei Zhong\\
Longguang Zhong\\
M. Zhou\\
Qiankang Zhou\\
Runjie Zhou\\
Ruozhang Zhou\\
Xinyu Zhou\\
Yiqiao Zhou\\
Zaida Zhou\\
Jinguo Zhu\\
Liya Zhu\\
Xinhao Zhu\\
Yangjunfeng Zhu\\
Yuxuan Zhu\\
Zhen Zhu\\
Chen Zhuang\\
Weiyu Zhuang\\
Xinxing Zu\\
Kimi K3\\
}
\end{multicols}
\newpage

\section{Details of Sigmoid Tanh Unit GLU}
\label{app:situglu}

The design goal of SiTU-GLU (\S\ref{sec:situ}) is to bound the SwiGLU product without discarding the characteristic shape of Swish: an approximately linear response around the origin and a vanishing negative tail.
Fig.~\ref{fig:situglu} shows the gate and up branches together with their complete scalar responses.

\paragraph{Smoothly capping both branches}
SiTU caps the linear factor of Swish as $\beta_1\tanh(\mathbf{W}_g\bm{x}/\beta_1)$ while retaining the sigmoid factor~\citep{kimik3blog}.
Because the sigmoid already drives the negative gate response toward zero, this change primarily controls large positive activations without removing the negative tail.
\kimi{3} applies the same construction to the up branch as $\beta_2\tanh(\mathbf{W}_u\bm{x}/\beta_2)$, preventing either branch from dominating the product.

\paragraph{Local and limiting behavior}
For a scalar $z$ near the origin, the scaled $\tanh$ satisfies
\begin{equation}
    \beta\tanh\!\left(\frac{z}{\beta}\right)
    = z + O\!\left(\frac{z^3}{\beta^2}\right).
\end{equation}
SiTU-GLU therefore matches SwiGLU to first order around the origin.
It also recovers SwiGLU pointwise as $\beta_1,\beta_2\rightarrow\infty$.

\paragraph{Bounded output}
Since $|\tanh(z)|<1$ and $0<\operatorname{Sigmoid}(z)<1$, every output coordinate satisfies
\begin{equation}
    \left\|\operatorname{SiTU\text{-}GLU}(\bm{x})\right\|_{\infty}
    \leq \beta_1\beta_2
    = 100,
    \label{eq:situglu-bound}
\end{equation}
for $\beta_1=4$ and $\beta_2=25$.
Unlike hard clamping of gate pre-activations, the smooth cap preserves nonzero gradients away from saturation boundaries, which we find to give better training behavior.

\section{Derivation of Quantile Balancing}
\label{app:qb-derivation}

This appendix derives the Quantile Balancing (QB) updates used in \S\ref{sec:stable-latent-moe} from optimal balanced assignment, following~\citep{su2026qb}; the assignment perspective on expert load balancing goes back to BASE Layers~\citep{lewis2021base} and BIP~\citep{sun2025bip}.
Let $\bm{s}\in\mathbb{R}^{m\times n}$ collect the router scores of $m$ tokens over $n$ experts, where each token selects exactly $k$ experts and $x_{i,j}\in\{0,1\}$ indicates whether token $i$ is assigned to expert $j$.
The maximum-score balanced assignment, in which each expert serves exactly $mk/n$ tokens (assumed integral), is
\begin{equation}
    \max_{x_{i,j}\in\{0,1\}} \sum_{i,j} x_{i,j}s_{i,j}
    \qquad \text{s.t.}\qquad
    \sum_j x_{i,j}=k,
    \qquad
    \sum_i x_{i,j}=\frac{mk}{n}.
    \label{eq:qb-lp}
\end{equation}

\paragraph{Linear relaxation and duality}
Relaxing $x_{i,j}\in\{0,1\}$ to $x_{i,j}\in[0,1]$ turns Eq.~\ref{eq:qb-lp} into a linear program, whose optimum is integral by the standard integrality of the bipartite $b$-matching polytope; the relaxation is therefore exact.
Introducing free multipliers $\alpha_i$ and $\beta_j$ for the token- and expert-side equality constraints, respectively, the relaxed problem can be written in max--min form as
\begin{equation}
    \max_{x_{i,j}\in[0,1]}\min_{\alpha_i,\beta_j}\;
    \sum_{i,j} x_{i,j}s_{i,j}
    - \sum_i \alpha_i\Big(\sum_j x_{i,j} - k\Big)
    - \sum_j \beta_j\Big(\sum_i x_{i,j} - \tfrac{mk}{n}\Big).
    \label{eq:qb-max-min}
\end{equation}
The objective is linear in each of $\bm{x}$, $\bm{\alpha}$, and $\bm{\beta}$, and the feasible sets are convex, so the minimax theorem allows exchanging the order of optimization:
\begin{equation}
    \min_{\alpha_i,\beta_j}\max_{x_{i,j}\in[0,1]}\;
    \sum_{i,j} x_{i,j}\big(s_{i,j} - \alpha_i - \beta_j\big)
    + k\sum_i \alpha_i
    + \frac{mk}{n}\sum_j \beta_j.
    \label{eq:qb-min-max}
\end{equation}
The inner maximum is separable over entries, with $x_{i,j}^*=1$ if $s_{i,j}-\alpha_i-\beta_j>0$ and $x_{i,j}^*=0$ if $s_{i,j}-\alpha_i-\beta_j<0$; the tie case has measure zero in practice.
Substituting $x^*$ gives the convex dual objective
\begin{equation}
    \min_{\alpha_i,\beta_j}\;
    \mathcal{L}(\bm{\alpha},\bm{\beta})
    := \sum_{i,j}\max\big(0,\; s_{i,j} - \alpha_i - \beta_j\big)
    + k\sum_i \alpha_i
    + \frac{mk}{n}\sum_j \beta_j.
    \label{eq:qb-dual}
\end{equation}

\paragraph{Exact coordinate minimization}
We minimize Eq.~\ref{eq:qb-dual} by alternately solving for $\bm{\alpha}$ with $\bm{\beta}$ fixed and vice versa; each subproblem admits a closed-form exact solution.
With $\bm{\beta}$ fixed, the problem decouples over tokens, and for token $i$ we solve
\begin{equation}
    \min_{\alpha}\; k\alpha + \sum_{j}\max\big(0,\; s_{i,j} - \beta_j - \alpha\big).
    \label{eq:qb-alpha-sub}
\end{equation}
This objective is piecewise linear in $\alpha$ with slope $k$ minus the number of margins $s_{i,j}-\beta_j$ exceeding $\alpha$; it is therefore minimized exactly when $k$ margins lie above $\alpha$, i.e., for any $\alpha_i^*$ between the $k$-th and $(k{+}1)$-th largest entries of $\bm{s}_i-\bm{\beta}$.
By convention we take the $(k{+}1)$-th largest entry, which is equivalently the $(1-k/n)$-th quantile:
\begin{equation}
    \alpha_i^*
    = \operatorname{quantile}_{1-k/n}\big(\bm{s}_i - \bm{\beta}\big).
    \label{eq:qb-alpha-sol}
\end{equation}
Symmetrically, with $\bm{\alpha}$ fixed, expert $j$ solves $\min_{\beta}\frac{mk}{n}\beta + \sum_i \max(0, s_{i,j}-\alpha_i-\beta)$, whose minimizer is the $(mk/n{+}1)$-th largest entry of $\bm{s}_{:,j}-\bm{\alpha}$, again the $(1-k/n)$-th quantile:
\begin{equation}
    \beta_j^*
    = \operatorname{quantile}_{1-k/n}\big(\bm{s}_{:,j} - \bm{\alpha}\big).
    \label{eq:qb-beta-sol}
\end{equation}
Both updates are thus the same quantile along the token and expert axes, respectively, which gives the method its name.
Fig.~\ref{fig:quantile-balancing} illustrates the expert-side update as equalizing the accepted upper tail of each expert's margin distribution, and Alg.~\ref{alg:qb} summarizes the resulting alternating solver.

\begin{algorithm}[t]
    \caption{The alternating QB solver.\label{alg:qb}}
    \KwIn{score matrix $\bm{s}\in\mathbb{R}^{m\times n}$}
    \KwOut{assignment $\bm{x}\in\{0,1\}^{m\times n}$}
    Initialize $\bm{\beta} = \bm{0}_{1\times n}$\;\DontPrintSemicolon
    \For{$t = 1, 2, \cdots, T$}{
        $\bm{\alpha} \leftarrow \operatorname{desc\_sort}(\bm{s}-\bm{\beta},\ \mathrm{axis}{=}1)_{[:,\,k:k+1]}$\;\DontPrintSemicolon
        $\bm{\beta} \leftarrow \operatorname{desc\_sort}(\bm{s}-\bm{\alpha},\ \mathrm{axis}{=}0)_{[mk/n:\,mk/n+1]}$\;\DontPrintSemicolon
    }
    \Return{$\bm{x}$ with $x_{i,j}=1$ \textnormal{if} $j\in\operatorname{argtop}_k(\bm{s}_i-\bm{\beta})$, \textnormal{and} $0$ \textnormal{otherwise}}
\end{algorithm}

\paragraph{From assignment to routing}
At the optimum of Eq.~\ref{eq:qb-dual}, $x_{i,j}^*=1$ if and only if $s_{i,j}-\alpha_i^*-\beta_j^*>0$; combined with the token constraint $\sum_j x_{i,j}^*=k$, the selected experts are exactly the Top-$k$ entries of $\bm{s}_i-\bm{\beta}^*$.
Routing therefore requires only the expert thresholds $\bm{\beta}\in\mathbb{R}^{n}$ (equivalently, the bias $\bm{b}=-\bm{\beta}$ of Eq.~\ref{eq:moe-routing}), while the token thresholds $\bm{\alpha}\in\mathbb{R}^{m}$ are intermediate variables tied to the dynamic training batch and are discarded.
This asymmetry preserves train--inference consistency: at deployment, routing is a fixed Top-$k$ selection with a frozen bias, and no quantile computation is needed.

\paragraph{Relation to sign-based loss-free updates}
The expert-side subproblem underlying Eq.~\ref{eq:qb-beta-sol} has (sub)gradient
\begin{equation}
    \frac{\partial \mathcal{L}}{\partial \beta_j}
    = \frac{mk}{n} - \sum_{i=1}^{m}\chi\big(s_{i,j}-\alpha_i-\beta_j>0\big),
    \label{eq:qb-grad}
\end{equation}
i.e., the target load minus the observed load of the expert $j$.
A SignSGD step on this objective recovers the fixed-step sign update of auxiliary-loss-free balancing~\citep{deepseekaiv3}, up to the sign convention $\bm{b}=-\bm{\beta}$: the sign update retains only the direction of the load error in Eq.~\ref{eq:qb-grad}, whereas QB jumps directly to the exact coordinate minimizer of the same dual objective.
This view explains both why QB requires no learning-rate-like hyperparameter and why it equilibrates within a few update steps even for nearly $10^3$ experts.
QB is likewise related to BIP~\citep{sun2025bip}, which solves the same assignment with inequality constraints $\sum_j x_{i,j}\le k$ and $\sum_i x_{i,j}\le mk/n$; the induced non-negativity constraints on $\bm{\alpha}$ and $\bm{\beta}$ add a $\max(0,\cdot)$ clipping to both updates, which can only suppress over-selected experts without promoting under-selected ones, and markedly slows equilibration in our experiments.
Finally, the resulting fixed-Top-$k$ routing is related to expert-specific threshold routing but differs from Expert Threshold routing, which maintains EMA thresholds and permits a variable number of selected experts per token~\citep{sun2026expertthreshold}.

\section{Histogram-Based Quantile Estimation}
\label{app:qb-histogram}

The QB update of Eq.~\ref{eq:qb-update} asks for a quantile taken over the whole training step: for each of the $n$ experts, the $(1-k/n)$-th quantile of the margins $s_{i,j}-\alpha_i$, where the token count $m$ spans millions of tokens sharded across data-parallel ranks and gradient-accumulation steps.
Gathering $O(mn)$ margins for an exact quantile is impractical inside the training loop.
The key observation is that the update never needs the margins themselves, only their per-expert distribution, which a histogram summarizes at fixed cost.
\kimi{3} therefore maintains a binned histogram per expert and reads the quantile from it.
Concretely, we histogram the \emph{required bias} $r_{i,j}:=\alpha_i-s_{i,j}$, the bias that would place expert $j$ exactly at token $i$'s cutoff; negating the margins reverses their order, so the QB target $\widehat{b}_j$ of Eq.~\ref{eq:qb-update} is exactly the $(k/n)$-quantile of $r_{:,j}$.

\paragraph{Binning range}
The first question is which interval to bin over, and here the required bias helps: its range is bounded by the current bias itself.
Router scores are sigmoid outputs, so $s_{i,j}\in(0,1)$, and the cutoff $\alpha_i$ is itself the biased score $s_{i,j'}+b_{j'}$ of some expert $j'$, so it lies in $(b_{\min},\,1+b_{\max})$, with $b_{\min}$ and $b_{\max}$ the extremes of the current bias.
Every $r_{i,j}$ therefore falls in $[b_{\min}-1,\,b_{\max}+1]$.
We partition this interval into $B$ uniform bins, which we find sufficient in practice, and recompute the range every step, so the bin width $w=(b_{\max}-b_{\min}+2)/B$ stays adapted to the bias as it spreads to correct imbalance.

\paragraph{Accumulation and recovery}
The rest of the procedure follows the structure of a training step.
During each forward pass, every rank scatter-adds its local $r_{i,j}$ values into a per-expert count matrix $\mathbf{H}\in\mathbb{N}^{n\times B}$, accumulating over all micro-batches with no communication.
At the end of the step, a single all-reduce sums the local counts into the global histogram, and every rank recovers the quantile from the same pooled counts.
Each expert's histogram counts every token once, so the target rank is exactly the target load $q=mk/n$ of \S~\ref{sec:qb}, now taken over the full step: we select the first bin whose cumulative count reaches $\lceil q\rceil$ and interpolate linearly within it.
If bin $\beta_j$ is selected, with cumulative count $c_j$ before it and $h_j$ counts inside it, then
\begin{equation*}
    \widehat{b}_j
    = b_{\min}-1+\Bigl(\beta_j+\operatorname{clip}\bigl(\tfrac{q-c_j}{h_j},\,0,\,1\bigr)\Bigr)w,
\end{equation*}
and the resulting biases are mean-centered as in Eq.~\ref{eq:qb-update}.

\paragraph{Properties}
Three properties make this estimator practical at scale.
First, it is accurate: the cumulative counts are exact at bin edges, so the true quantile and its estimate lie in the same bin and the error is bounded by the bin width $w$; with $B=1000$ this is at most a few $10^{-3}$, and we observe no measurable residual load imbalance.
Second, it is cheap: the only communication is one integer all-reduce of $nB$ values per layer per step, independent of $m$, which in our configuration is below $1\%$ of the cost of exchanging the raw margins over a process group every micro-batch, the natural alternative.
Third, it estimates the right quantity: because counts are additive, the global histogram is exactly invariant to how tokens are partitioned across ranks or accumulation steps, and the estimate is the quantile of the pooled global batch rather than an average of per-rank quantiles, which generally differs. As a further refinement, maintaining an exponential moving average of the estimated quantiles across steps reduces batch-to-batch sampling noise and can improve load balance still further.

\section{MoonEP General Upper Bound Proof}
\label{app:moonep-proof}

 Let $m_r(P)$ denote the number of redundant experts placed on rank $r$ under
 plan $P$. For a router output $I$, the planning objective is to minimize the
 maximum number of redundant experts on any rank, i.e.,
 $M(I) = \min_P \max_r \{m_r(P)\}$. We prove that $M(I) \le E/R$ always holds
 (Theorem 1) and that this bound is essentially tight: there exist router
 outputs for which $M = \lceil E(R-1)/R^2 \rceil \approx E/R$ (Theorem 2).

\paragraph{Proof of Theorem 1 (General Upper Bound)} 
The goal is to prove that $M(I) \le E/R$ holds for any router output $I$. Key lemma: there exists a plan $P^*$ such that every EP rank receives exactly the same number of tokens ($S \times K$), and the remote tokens of each rank come from only one other EP rank. The construction is as follows: initially, every rank holds only local tokens, and ranks are classified as underloaded or overloaded accordingly. We repeatedly pick an underloaded rank and an overloaded rank, and migrate tokens from the overloaded rank to fill the underloaded rank exactly up to the balanced value $S \times K$; the overloaded rank may remain overloaded, become exactly balanced, or become underloaded, and is put back into the corresponding set. This is repeated until all ranks are perfectly balanced. Each fill makes one underloaded rank balanced and it never changes afterwards, so the process terminates after at most $R-1$ fills; meanwhile, each rank is filled at most once, so its remote tokens come from a single rank, which proves the lemma. Consequently, supposing all remote tokens of rank $r$ come from rank $s$; these tokens belong to at most $E/R$ local experts on rank $s$, hence $m_r(P^*) \le E/R$, and therefore

\begin{equation}
    M(I)=\min_{P}\ \max_{r}\big\{m_r(P)\big\} \le \max_{r}\big\{m_r(P^*)\big\} \le \frac{E}{R}
\end{equation}

\paragraph{Proof of Theorem 2 (Tightness of the Upper Bound)} 
Construct a router output $I^*$ as follows: the experts on EP rank 0 receive no tokens, while all experts on the other $R-1$ ranks share all tokens evenly. Then all $S \times K \times R$ tokens are evenly divided among $E(R-1)/R$ experts, so each expert receives $\frac{SKR^2}{E(R-1)}$ tokens. Under any plan $P$, rank 0 must receive $S \times K$ tokens, all of which are remote, and these tokens involve at least $SK \big/ \frac{SKR^2}{E(R-1)} = \frac{E(R-1)}{R^2}$ distinct experts; taking the ceiling, rank 0 requires at least $\left\lceil\frac{E(R-1)}{R^2}\right\rceil$ redundant experts, hence $M(I^*) \ge \left\lceil\frac{E(R-1)}{R^2}\right\rceil$. Conversely, by constructing a plan with the filling procedure from the proof of Theorem 1 and migrating tokens expert-wise preferentially, the number of redundant experts on every rank can be kept within this value, so equality holds. Since $\left\lceil\frac{E(R-1)}{R^2}\right\rceil \approx \frac{E}{R}$ when $R$ is large, the upper bound in Theorem 1 is essentially tight: there is no general upper bound significantly smaller than $E/R$.

\section{Chat Template}
\label{sec:post-chat-template}

The \kimi{3} chat template is redesigned around three goals. The first is \emph{extensibility}: new capabilities should be introduced through backward-compatible message formats rather than template revisions, so that a single template serves the entire model generation. The second is a \emph{low alignment tax}: the format should be learnable with minimal supervised data, supporting a pipeline in which a lightly fine-tuned pre-trained model can proceed directly to reinforcement learning. The third is \emph{decoding friendliness}: the structure should admit simple encoders, streaming parsers, and grammar-constrained enforcers. To these ends, the template adopts XTML (eXtensible Token Markup Language), an XML-like markup in which the angle-bracket syntax is replaced by three reserved special tokens: \texttt{[open]}, \texttt{[sep]} and \texttt{[close]}, with an additional \texttt{[end\_of\_msg]} token as the generation stop marker. An element \texttt{[open]tag attr="value"[sep] \dots\ [close]tag[sep]} is isomorphic to its XML counterpart, but every structural boundary is an explicit special token, which removes tokenization ambiguity at element boundaries and simplifies constrained decoding.

\begin{figure}[t]
    \centering
    \definecolor{ctfigink}{HTML}{1C1C1E}
\definecolor{ctfiggray}{HTML}{777777}
\definecolor{ctfiggrid}{HTML}{D6D6D8}
\definecolor{ctfigred}{HTML}{B92622}
\definecolor{ctfigredfill}{HTML}{FBF1F0}
\definecolor{ctfigblue}{HTML}{1773E6}
\definecolor{ctfigbluefill}{HTML}{EDF4FD}

\begin{adjustbox}{width=\textwidth,center}
        \begin{tikzpicture}[
                        x=1pt,
                        y=1pt,
                        outer sep=0pt,
                        paneltitle/.style={font=\fontsize{9.2}{10.8}\selectfont, text=ctfigink, align=center},
                        zonelabel/.style={font=\fontsize{7.2}{8.4}\selectfont, text=ctfiggray},
                        zone/.style={draw=ctfiggrid, line width=0.5pt, rounded corners=2.5pt, fill=white},
                        chip/.style={draw=ctfigink!75, line width=0.5pt, rounded corners=1.5pt,
                                        fill=white, font=\fontsize{6.9}{8}\selectfont, text=ctfigink,
                                        inner xsep=3pt, inner ysep=2.2pt},
                        optchip/.style={chip, draw=ctfigblue!80, fill=ctfigbluefill},
                        hotchip/.style={chip, draw=ctfigred, fill=ctfigredfill, line width=0.7pt},
                        expchip/.style={chip, dashed, draw=ctfiggray},
                        chan/.style={line width=0.5pt, rounded corners=2pt},
                        chanthink/.style={chan, draw=ctfiggray!75, fill=white},
                        chanresp/.style={chan, draw=ctfigblue!65, fill=ctfigbluefill},
                        chantools/.style={chan, draw=ctfigred!60, fill=ctfigredfill},
                        callcard/.style={draw=ctfigink!55, line width=0.45pt, rounded corners=1.8pt, fill=white},
                        mono/.style={font=\scriptsize\ttfamily, text=ctfigink},
                        textline/.style={draw=ctfiggrid, line width=1.7pt, line cap=round},
                        flow/.style={draw=ctfiggray, line width=0.55pt,
                                                -{Latex[length=3.8pt,width=3pt]}},
                        zoom/.style={draw=ctfiggray!85, line width=0.45pt, dashed},
                ]
                \path[use as bounding box] (0,-4) rectangle (560,205);

                \node[paneltitle, anchor=north] at (82,203)
                {(a) Context layout};

                \draw[zone] (6,150) rectangle (158,184);
                \node[zonelabel, anchor=north] at (82,182) {global option messages};
                \node[optchip] at (47,162) {tool-declare};
                \node[optchip] at (110,162) {thinking-effort};

                \draw[zone] (6,80) rectangle (158,140);
                \node[zonelabel, anchor=north] at (82,138) {input messages};
                \node[chip] (m-sys) at (48,122) {system};
                \node[chip] (m-usr) at (112,122) {user};
                \node[chip] (m-tool) at (48,108) {tool};
                \node[hotchip] (m-asst) at (112,108) {assistant};
                \node[expchip] (m-dyn) at (82,94) {dynamic tool-declare};

                \draw[zone] (6,38) rectangle (158,72);
                \node[zonelabel, anchor=north] at (82,70) {one-shot option messages};
                \node[optchip] at (46,50) {tool-choice};
                \node[optchip] at (112,50) {response-format};

                \draw[flow] (82,147) -- (82,143);
                \draw[flow] (82,77) -- (82,75);

                \node[chip] (genprefix) at (43.5,24) {[open]think[sep]};
                \node[zonelabel] at (79.7,24) {/};
                \node[chip] at (115.8,24) {[open]response[sep]};
                \node[zonelabel, anchor=north] at (82,15) {generation prefix};
                \draw[flow] (82,36) -- (82,30);

                \node[paneltitle, anchor=north] at (270,203)
                {(b) Assistant message};

                \draw[zoom] (m-asst.north east) -- (176,180);
                \draw[zoom] (m-asst.south east) -- (176,22);

                \draw[zone] (180,18) rectangle (360,184);
                \node[mono, anchor=north west] at (186,181) {[open]message role="assistant"[sep]};

                \draw[chanthink] (188,124) rectangle (352,170);
                \node[mono, anchor=north west] at (194,167) {[open]think[sep]};
                \draw[textline] (198,151) -- (272,151);
                \draw[textline] (198,143.5) -- (256,143.5);
                \node[mono, text=ctfiggray, anchor=south west] at (194,127) {[close]think[sep]};

                \draw[chanresp] (188,80) rectangle (352,118);
                \node[mono, anchor=north west] at (194,115) {[open]response[sep]};
                \draw[textline] (198,101) -- (272,101);
                \draw[textline] (198,93.5) -- (256,93.5);
                \node[mono, text=ctfiggray, anchor=south west] at (194,83) {[close]response[sep]};

                \draw[chantools] (188,32) rectangle (352,74);
                \node[mono, anchor=north west] at (194,71) {[open]tools[sep]};
                \node[mono, text=ctfiggray] at (270,53) {$\cdots$};
                \node[mono, text=ctfiggray, anchor=south west] at (194,35) {[close]tools[sep]};

                \node[mono, text=ctfiggray, anchor=south west] (msgclose) at (186,21.5) {[close]message[sep]};
                \node[chip, right=3.5pt of msgclose] {[end\_of\_msg]};

                \node[paneltitle, anchor=north] at (465,203)
                {(c) Tools channel};

                \draw[zoom] (352,72) -- (371,180);
                \draw[zoom] (352,34) -- (371,40);

                \draw[chantools] (373,34) rectangle (559,184);
                \node[mono, anchor=north west] at (379,181) {[open]tools[sep]};

                \draw[callcard] (379,118) rectangle (555,170);
                \node[mono, align=left, anchor=north west, inner xsep=2.5pt, inner ysep=2pt]
                at (384,167.5) {[open]call tool="python" index="1"[sep]\\[0.5pt]
                        {\color{ctfiggray}\fontsize{6.0}{7.2}\selectfont [open]argument key="code" type="string"[sep]}\\[0.5pt]
                        {\color{ctfiggray}\fontsize{6.0}{7.2}\selectfont \hspace{1.2em}$\cdots$}\\[0.5pt]
                        {\color{ctfiggray}\fontsize{6.0}{7.2}\selectfont [close]argument[sep]}\\[0.5pt]
                        {\color{ctfiggray}[close]call[sep]}};

                \draw[callcard] (379,56) rectangle (555,110);
                \node[mono, align=left, anchor=north west, inner xsep=2.5pt, inner ysep=2pt]
                at (384,107.5) {[open]call tool="search" index="2"[sep]\\[0.5pt]
                        {\color{ctfiggray}\fontsize{6.0}{7.2}\selectfont [open]argument key="options" type="object"[sep]}\\[0.5pt]
                        {\color{ctfiggray}\fontsize{6.0}{7.2}\selectfont \hspace{1.2em}\{"timeout": 150\}}\\[0.5pt]
                        {\color{ctfiggray}\fontsize{6.0}{7.2}\selectfont [close]argument[sep]}\\[0.5pt]
                        {\color{ctfiggray}[close]call[sep]}};

                \node[mono, text=ctfiggray, anchor=north west] at (379,48) {[close]tools[sep]};
        \end{tikzpicture}
\end{adjustbox}
    \caption{Structure of the \kimi{3} chat template.
        \textbf{(a)} Context layout: global option messages precede the input messages, while one-shot option messages follow them, so that per-request options leave the history KV cache intact; dynamically loaded tools are injected mid-session as input option messages (dashed).
        \textbf{(b)} Anatomy of an assistant message: the body is organized into \texttt{think}, \texttt{response}, and \texttt{tools} channels.
        \textbf{(c)} Expansion of the \texttt{tools} channel: parallel tool calls are indexed so that tool results can be matched to their calls, and arguments are typed.}
    \label{fig:chat-template}
\end{figure}
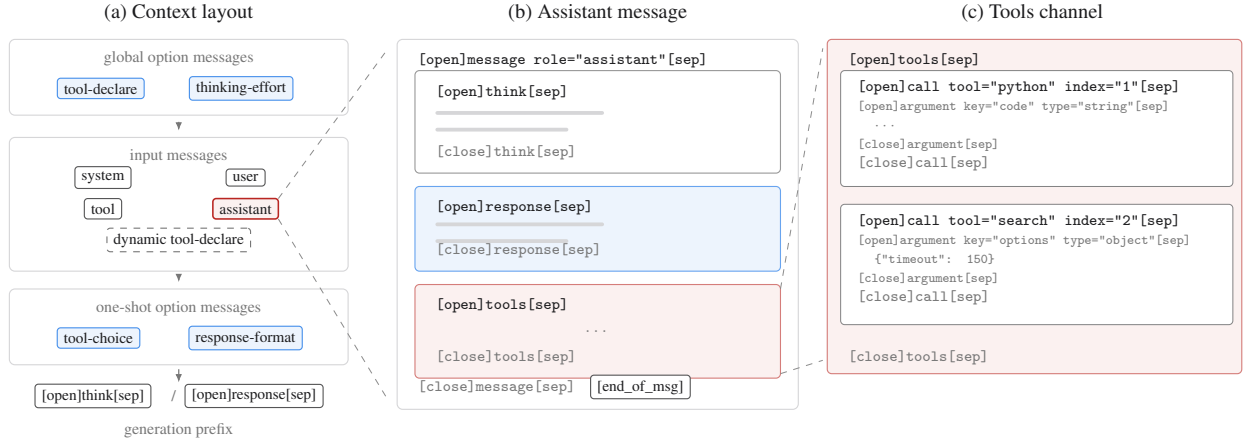

\paragraph{Messages and zones}
The top-level unit of the context is the message, and messages fall into two categories by origin (Fig.~\ref{fig:chat-template}a). \emph{Input messages} serialize the \texttt{messages} field of the request, covering the familiar system, user, assistant, and tool roles. \emph{Option messages} translate request options into instructions that the model reads in context, and their placement reflects their scope. \emph{Global options}---the tool declaration (\texttt{type="tool-declare"}) and the reasoning-effort setting---appear before all input messages: they govern the whole session and rarely change, so modifying them invalidates the KV cache anyway. \emph{One-shot options} (\texttt{tool\_choice}, \texttt{response\_format}) are appended after the input messages, so that per-request changes leave the history KV cache intact. A third kind, the \emph{input option message}, is interleaved with input messages to supplement or override a global option mid-session. This mechanism supports \emph{dynamically loaded tools}: tools retrieved or loaded during a conversation are announced through an additional tool-declare message, after which the model's available toolset expands without rebuilding the preceding context.

\paragraph{Channels}
The body of an assistant message is organized into \emph{channels}, a concept inspired by OpenAI's Harmony response format~\citep{openai2025harmony}: \texttt{think} carries the reasoning trace, \texttt{response} the user-visible answer, and \texttt{tools} the tool calls (Fig.~\ref{fig:chat-template}b). The two generation modes are selected purely through the generation prefix---\texttt{[open]think[sep]} for thinking mode and \texttt{[open]response[sep]} for instruct mode---rather than through separate templates. \kimi{3} supports only \emph{preserved thinking}: in thinking mode, the think channel is always retained in the history---kept even when its content is empty---so that the model observes a consistent message structure across turns; in instruct mode, historical messages contain only the response and tools channels.

\paragraph{Tool calling}
Within the tools channel, each call carries \texttt{tool} and \texttt{index} attributes; the index numbers parallel calls within a message, and each tool-result message repeats the same \texttt{tool}/\texttt{index} pair and follows the order of its call, so that results are unambiguously associated with calls. Arguments are typed: string arguments appear as raw text, while values of other JSON types are compactly serialized. Free-form text such as code is therefore a first-class citizen rather than an escaped JSON string. A pure-JSON fallback block covers inputs whose arguments cannot be decomposed into typed argument blocks; it occurs only in input tokens, never in model outputs, and its loss is masked during training.

\paragraph{Reasoning effort and options}
Reasoning effort is exposed as a global option message of type \texttt{thinking-effort}, inserted after the tool declaration and before the input messages. Instead of modifying the generation prefix or exposing a token budget, the message states the requested level in natural language and acts as a generation-constraint instruction. The schema reserves four levels (\texttt{low}, \texttt{medium}, \texttt{high}, and \texttt{max}), of which \kimi{3} supports a subset. This representation decouples the effort interface from the template syntax, and it aligns directly with the effort-conditioned training described in \S\ref{sec:post-sft} and \S\ref{sec:post-rl}.

More broadly, this is the common implementation of all option messages: \texttt{tool\_choice}, \texttt{response\_format}, and \texttt{thinking-effort} are each translated into a short natural-language instruction placed in context, rather than into dedicated special syntax. Because the pre-trained model already follows such instructions well, new options can be introduced with little or no additional training---a direct embodiment of the low-alignment-tax design principle stated above.

\end{document}